\documentclass[journal,compsoc,twoside]{IEEEtran}

\ifCLASSOPTIONcompsoc
  \usepackage[nocompress]{cite}
\else
  \usepackage{cite}
\fi

\usepackage{graphicx}
\usepackage{amsmath}
\usepackage{mathrsfs}
\usepackage{array}

\makeatletter		

\newcommand{\Rmnum}[1]{\expandafter\@slowromancap\romannumeral #1@}
\makeatother
\usepackage{ragged2e}
\usepackage{algorithm}
\usepackage{algorithmic}
\usepackage{booktabs}
\usepackage{makecell}
\usepackage{amssymb}
\usepackage{bm}		
\usepackage[table]{xcolor}
\usepackage[colorlinks=False, linkcolor=black, citecolor=black]{hyperref} 
\hypersetup{hidelinks}
\usepackage{balance}
\usepackage{threeparttable}
\usepackage{caption}
\usepackage{subcaption}
\usepackage{colortbl}
\usepackage{nicematrix}
\usepackage{tabularx}
\usepackage{multirow}
\usepackage{dsfont}
\usepackage{enumitem}
\usepackage{utfsym}

\usepackage[american]{babel}

\newcolumntype{L}[1]{>{\raggedright\arraybackslash}p{#1}}
\newcolumntype{C}[1]{>{\centering\arraybackslash}p{#1}}
\newcolumntype{R}[1]{>{\raggedleft\arraybackslash}p{#1}}

\usepackage{amsthm}

\def\sx{NVDS$^{\textbf{+}}$}
\def\llarge{NVDS$^{\textbf{+}}_{\textbf{Large}}$}
\def\ssmall{NVDS$^{\textbf{+}}_{\textbf{Small}}$}
\def\FLOW{Flow-Guided Consistency Fusion}

\def\flow{flow-guided consistency fusion}

\def\SbN{Stabilization Network}

\def\sbn{stabilization network}
\def\data{VDW}
\def\sota{state-of-the-art}
\def\reffig{Fig.}
\def\reftab{Table}
\def\refequ{Eq.}
\def\refsec{Sec.}
\def\city{CityScapes}
\DeclareMathSymbol{@}{\mathord}{letters}{"3B}

\begin{document}
%
\title{NVDS$^{\textbf{+}}$: Towards Efficient and Versatile \\ Neural Stabilizer for Video Depth Estimation}

\author{Yiran ~Wang, 
	Min~Shi, 
        Jiaqi~Li, 
        Chaoyi~Hong, 
        Zihao~Huang, 
        Juewen Peng, \\
	Zhiguo~Cao,~\IEEEmembership{Member,~IEEE},
        Jianming~Zhang, 
        Ke~Xian, 
	and~Guosheng~Lin,~\IEEEmembership{Member,~IEEE}
\thanks{This work is supported by the National Natural Science Foundation of China under Grant 62406120. YW, MS, JL, CH, ZH, and ZC are with School of AIA, Huazhong University of Science and Technology (e-mail: \{wangyiran,min\_shi,lijiaqi\_mail,cyhong,zihaohuang,zgcao\}@hust.edu.cn).}


\thanks{KX is the corresponding author, with School of AIA and School of EIC, Huazhong University of Science and Technology (e-mail: kxian@hust.edu.cn).}
\thanks{JZ is with Adobe Research (e-mail: jianmzha@adobe.com).}

\thanks{JP and GL are with College of Computing and Data Science, Nanyang Technological University (e-mail: \{juewen.peng,gslin\}@ntu.edu.sg)}
}

\IEEEtitleabstractindextext{%

\begin{abstract}
\justifying
    Video depth estimation aims to infer temporally consistent depth.
    %
    One approach is to finetune a single-image model on each video with geometry constraints, which proves inefficient and lacks robustness.
    %
    An alternative is learning to enforce consistency from data, which requires well-designed models and sufficient video depth data. 
    To address both challenges, we introduce \sx{} that stabilizes inconsistent depth estimated by various single-image models in a plug-and-play manner. We also elaborate a large-scale Video Depth in the Wild (VDW) dataset, which contains 14,203 videos with over two million frames, making it the largest natural-scene video depth dataset.
    Additionally, a bidirectional inference strategy is designed to improve consistency by adaptively fusing forward and backward predictions.
    We instantiate a model family ranging from small to large scales for different applications.
    The method is evaluated on VDW dataset and three public benchmarks.
    To further prove the versatility, we extend \sx{} to video semantic segmentation and several downstream applications like bokeh rendering, novel view synthesis, and 3D reconstruction.
    Experimental results show that our method achieves
    significant improvements in consistency, accuracy, and efficiency. Our work serves as a solid baseline and data foundation for learning-based video depth estimation. Code and dataset are available at: \url{https://github.com/RaymondWang987/NVDS}

\end{abstract}

}

\maketitle

\IEEEdisplaynontitleabstractindextext

%
\IEEEpeerreviewmaketitle

\IEEEraisesectionheading{\section{Introduction}\label{intro}}

\IEEEPARstart{M}{onocular}
video depth estimation serves as a prerequisite for a variety of video applications, such as bokeh rendering~\cite{bokehme,videobokeh}, 2D-to-3D video conversion~\cite{n1}, and novel view synthesis~\cite{diudiu1,diudiu2,nsff}. 
%
An ideal video depth model should exhibit both spatial accuracy and temporal consistency.
%
Although recent developments in single-image depth models~\cite{dpt,MiDaS,zhuzhu1,newcrfs,ljq,k1,k3} and datasets~\cite{mega,kexian2020,kexian2018,jrtz} have notably improved the spatial accuracy, how to obtain temporal consistency, i.e., removing flickers in the predicted depth sequences, remains an unresolved question.
The prevailing video depth approaches~\cite{CVD,rcvd,dycvd} require test-time training. 
During inference, these methods finetune a single-image depth model on each specific testing video with geometry constraints and pose estimation, which are faced with two primary issues: limited robustness and heavy computation overhead. 
Due to heavy reliance on camera poses, \textit{e.g.}, 
CVD~\cite{CVD} shows erroneous predictions and robust-CVD~\cite{rcvd} produces artifacts for many videos with inaccurate pose estimation~\cite{colmapsfm,rcvd}. Moreover, test-time training is a time-consuming process. CVD~\cite{CVD} takes $40$ minutes for $244$ frames on four NVIDIA Tesla M40 GPUs.


\begin{figure}[!t]
\centering
\includegraphics[width=0.431\textwidth,trim=25 5 10 5,clip]{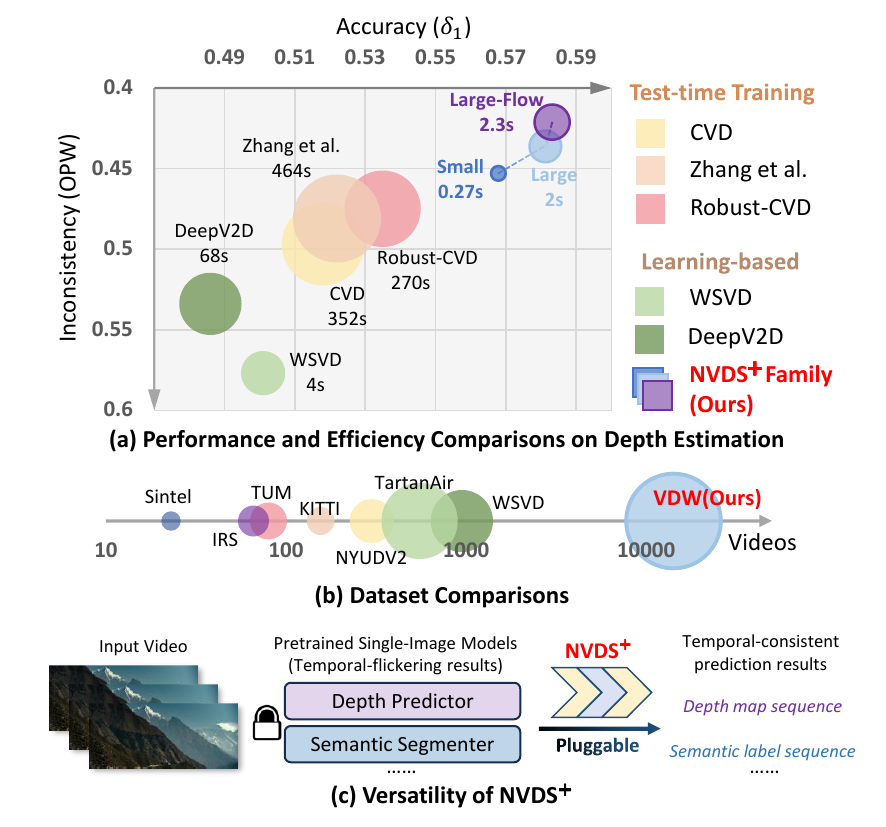}
\vspace{-5pt}
\caption{\textbf{(a) Performance and efficiency comparisons on depth estimation}. We provide \sx{} model family from \llarge{} for the best performance to \ssmall{} for real-time processing. Smaller circles mean faster speed. We also propose the \flow{} strategy (Large-Flow) to further enhance the consistency.
\sx{} outperforms prior arts by large margins. \textbf{ (b) Dataset comparisons}. 
Larger circles mean larger amounts of frames. We present \data{} dataset, the largest video depth dataset in the wild with diverse scenes. \textbf{ (c) Versatility of \sx{}.} We naturally extend the pluggable \sx{} to the video semantic segmentation task to prove the generality of our framework.} 
\label{fig:fig1}
\end{figure}

This motivates us to develop a learning-based model that learns to enforce consistency from video depth data. However, akin to all the deep-learning models, learning-based paradigm necessitates proper model design and sufficient training data.
In our scenario, where the scale and diversity of video depth data are limited, prior learning-based methods~\cite{deepv2d, fmnet, ST-CLSTM, MM21} exhibit inferior performance compared to test-time-training-based ones. Furthermore, these methods can not benefit from well-trained single-image depth predictors~\cite{dpt,MiDaS,newcrfs,MiDaSV31}. Both the model design and the availability of data persist as crucial challenges.
%


To address the two aforementioned challenges, based on our preliminary conference paper NVDS~\cite{nvds}, we propose a flexible learning-based framework termed \sx{}, which can be directly applied to different single-image depth models. \sx{} comprises a depth predictor and a \sbn{}. The depth predictor can be any off-the-shelf single-image depth model. Different from previous learning-based methods~\cite{MM21,fmnet,ST-CLSTM,deepv2d} that function as stand-alone models, \sx{} is a plug-and-play refiner for different depth predictors. Specifically, the \sbn{} processes initial flickering depth estimated by depth predictors and outputs temporally consistent results. 
Therefore, \sx{} can benefit from the depth models without extra effort. As for the design of \sbn{}, inspired by the attention~\cite{transformer} mechanism in other video tasks~\cite{stt,ctrans,vivit,cffm}, we adopt a cross-attention module in our framework. Each frame attends relevant information from adjacent frames for consistency. 


Apart from the pluggable \sbn{}, we also propose a training-free bidirectional inference strategy to further enlarge temporal receptive field and improve consistency, where outputs are obtained by adaptively fusing the forward and backward predictions. Specifically, we devise a \flow{} strategy to generate the fusion weights. Since frames or pixels with larger motions should have lower relevance with the final target depth, we will give smaller weights to the pixels with large motion amplitudes. Optical flow~\cite{gmflow,gmflowv2} is adopted to measure motion amplitudes of relevant frames and pixels. Bidirectional depth results are adaptively fused according to motions and relevance maps between reference and target frames. As shown in \reffig{}~\ref{fig:fig1}(a), our model with \flow{} (Large-Flow) achieves even better consistency. 

To balance efficiency and performance for different applications, we provide a model family of \sx{}, ranging from small to large scales. To achieve the best performance, we implement our \llarge{} working with different top-performing depth predictors~\cite{dpt,MiDaS,newcrfs}. On the other hand, the \ssmall{} model is built in pursuit of real-time processing, cooperating with varied lightweight depth predictors~\cite{MiDaS,MiDaSV31}. As shown in \reffig{}~\ref{fig:fig1}(a),  all our implementations outperform previous approaches in terms of consistency, accuracy, and efficiency significantly.

Moreover, we collect a large-scale natural-scene video depth dataset, Video Depth in the Wild (\data{}), to support the training of robust learning-based models. 
Current video depth datasets are mostly closed-domain~\cite{kitti,tum,nyu,scannet,irs}. 
A few in-the-wild datasets~\cite{sintel,tata,wsvd} are still limited in quantity, diversity, and quality, \textit{e.g.}, Sintel~\cite{sintel} only contains $23$ animated videos. 
In contrast, our \data{} dataset contains $14@203$ stereo videos of over $200$ hours and $2.23M$ frames from four different data sources, including movies, animations, documentaries, and web videos. We adopt a rigorous data annotation pipeline to obtain high-quality disparity ground truth for these data.
As shown in \reffig{}~\ref{fig:fig1}(b), to the best of our knowledge, \data{} is the largest in-the-wild video depth dataset with diverse scenes.

We conduct evaluations on \data{} and three public benchmarks: Sintel~\cite{sintel}, NYUDV2~\cite{nyu}, and KITTI~\cite{kitti}. 
Our method 
achieves \sota{} in both accuracy and consistency.  We also fit several different depth predictors~\cite{MiDaS,dpt,MiDaSV31,newcrfs} into our framework, which demonstrates that the \sx{} can stabilize the flickering results from different depth predictors without any extra effort.
Besides, our \ssmall{} only has $5$M parameters and achieves a real-time processing throughput of over $30$ fps.

As two exemplar tasks in dense prediction~\cite{dpt}, depth estimation and semantic segmentation are both important for downstream applications like autonomous driving and virtual reality. To demonstrate the versatility of \sx{} for video dense prediction, we extend our framework to video semantic segmentation~\cite{etc,cffm,mrcfa,ifr,SSLTM,tmanet}, as shown in \reffig{}~\ref{fig:fig1}(c). Similar to video depth estimation, video semantic segmentation aims to predict both accurate and consistent semantic labels for video frames. Naturally, we can use different single-image semantic segmentation models~\cite{segformer,oneformer} as semantic segmenters to output initial predictions. Then, through the plug-and-play paradigm, the neural stabilizer can stabilize the flickering results to improve the consistency. Our \sx{} attains \sota{} performance on \city{}~\cite{cityscapes} dataset and outperforms stand-alone video semantic segmentation models~\cite{SSLTM,mrcfa}, further demonstrating the generality of the proposed \sx{} framework. Our main contributions are summarized as follows:

\begin{itemize}[leftmargin=*]

    \item We propose a plug-and-play and bidirectional learning-based framework \sx{}, 
    which can be directly adapted to different depth predictors to remove flickers.
    
    \item We propose \data{} dataset, currently the largest video depth dataset in the wild with the most diverse scenes.

    \item The \flow{} strategy is proposed to further enhance the consistency, which can adaptively fuse the bidirectional results according to motion amplitudes between relevant frames and pixels.

    \item We implement a comprehensive model family, from \llarge{} model for the best performance to the \ssmall{} model for real-time applications.

    \item The \sx{} framework is naturally extended to the video semantic segmentation task and also achieves \sota{} performance in both accuracy and consistency.

\end{itemize}

Note that this paper is an extension of a conference version~\cite{nvds}.
Compared with the conference version~\cite{nvds}, we upgrade NVDS~\cite{nvds} to \sx{} with sufficient explorations in the following aspects: i) We provide a model family of \sx{} for performance-efficiency trade-off;
ii) We further explore the bidirectional inference strategy, proposing the \flow{} to adaptively fuse bidirectional results; iii) We prove the versatility of our framework for video dense prediction by extending \sx{} to video semantic segmentation;
iv) More comprehensive evaluations including results on KITTI~\cite{kitti} and \city{}~\cite{cityscapes} are conducted to prove our superiority; v) We showcase capabilities of \sx{} on downstream applications, including bokeh rendering, 3D video conversion, space-time view synthesis, and point cloud reconstruction. Please refer to our project page, demo video, and supplement for more details.

\section{Related Work}

\noindent\textbf{Consistent Video Depth Estimation.} In addition to predicting spatial-accurate depth, the core task of consistent video depth estimation is to achieve temporal consistency, \textit{i.e.}, removing the flickering effects between consecutive frames. Current video depth estimation approaches can be categorized into test-time training ones and learning-based ones.
Test-time-training-based methods train an off-the-shelf single-image depth estimation model on testing videos during inference with geometry~\cite{CVD,rcvd,dycvd} and pose~\cite{colmapmvs,colmapsfm,rcvd} constraints. The test-time training can be time-consuming. 
For example, as illustrated by CVD~\cite{CVD}, their method takes $40$ minutes on $4$ NVIDIA Tesla M40 GPUs to process a video of $244$ frames. Besides, these approaches are not robust on in-the-wild videos as they heavily rely on camera poses, which are not reliable for natural scenes.
In contrast, the learning-based approaches train models on video depth datasets by spatial and temporal supervision. ST-CLSTM~\cite{ST-CLSTM} adopts long short-term memory (LSTM) to model temporal relations. FMNet~\cite{fmnet} restores the depth of masked frames by the unmasked ones with convolutional self-attention~\cite{ctrans}. TC-Depth~\cite{tcdepth} improves consistency by pixel-wise similarities and self-supervision with camera pose. Cao \textit{et al.}~\cite{MM21} adopt a spatial-temporal propagation
network trained by knowledge distillation~\cite{kd0,kd1}. 
Khan \textit{et al.}~\cite{khan2023temporally} and CODD~\cite{wacv23} seek online depth estimation of stereo videos through point-based fusion and temporal depth aggregation respectively. ViTA~\cite{vita} utilizes a transformer adaptor with temporal embeddings in attention blocks. MAMo~\cite{mamo} proposes the memory update and memory attention mechanism to predict more accurate depth with temporal information. 
However, those methods cannot refine the results from different single-image depth models for consistency in a plug-and-use style. 
Their performance on consistency and accuracy is also limited. For example, as shown by FMNet~\cite{fmnet}, ST-CLSTM~\cite{stt} only exploits sub-sequences of several frames and produces flickers in the outputs. 
In this paper, we propose the \sx{} framework, which can be directly adapted to off-the-shelf depth models by the pluggable paradigm without extra training. 

\begin{figure*}[!t]
\centering
\includegraphics[width=0.97\textwidth,trim=0 30 0 30,clip]{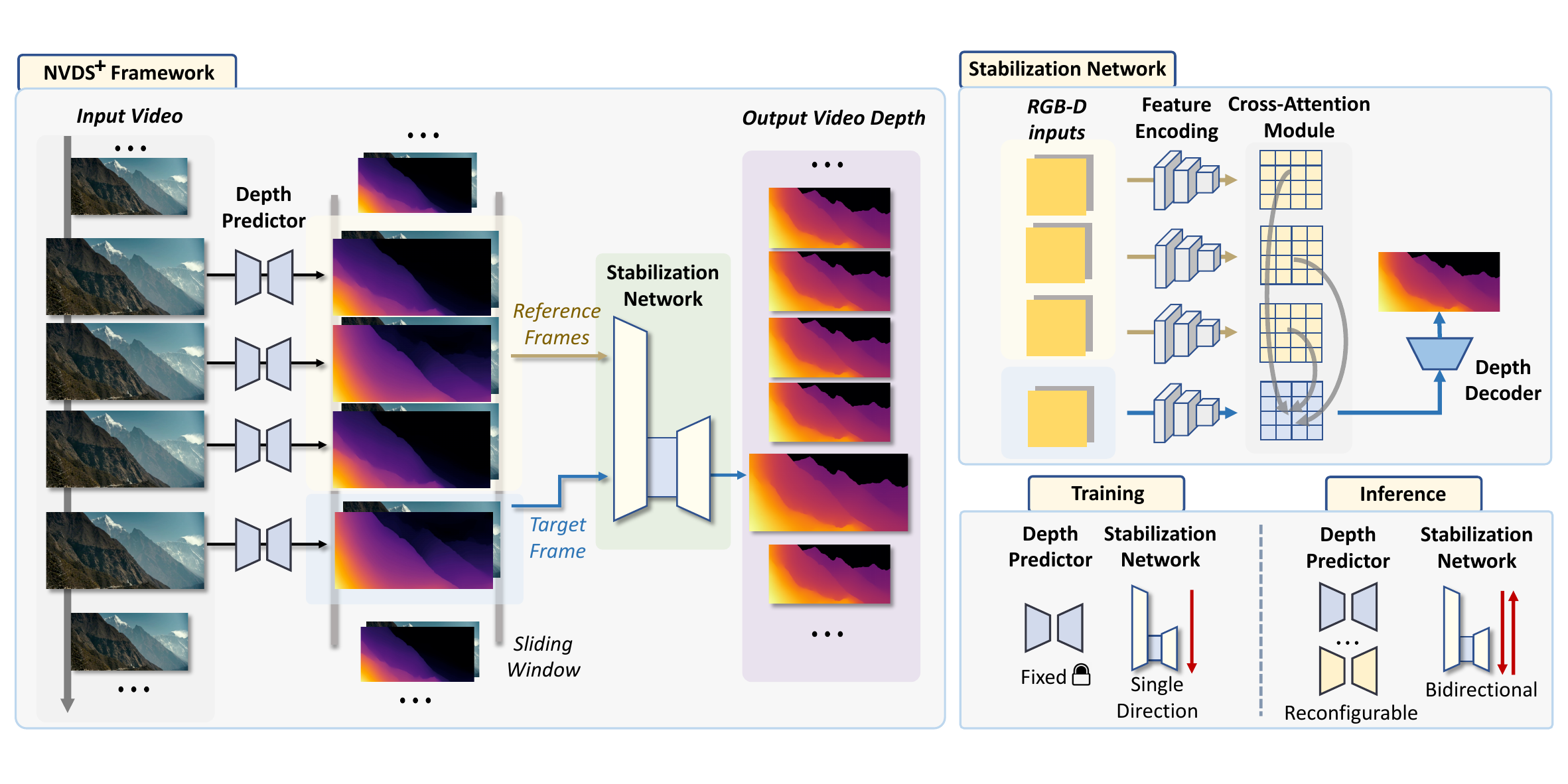}
\vspace{-3pt}
\caption{
\textbf{Overview of the \sx{} framework.}
Our framework consists of a depth predictor and a \sbn{}. The depth predictor can be any single-image depth model which produces initial flickering depth maps. Then, the \sbn{} refines the flickering depth maps into temporally consistent ones. The \sbn{} functions with a sliding window. The frame to be predicted fetches information from adjacent frames for stabilization.
During inference, our \sx{} framework can be directly adapted to any off-the-shelf depth predictors in a plug-and-play manner. We also devise bidirectional inference with \flow{} to further improve the consistency.}
\label{fig:pipeline}
\vspace{-7pt}
\end{figure*}

\noindent\textbf{Video Depth Datasets.} 
According to the scenes of samples, existing video depth datasets can be categorized into closed-domain datasets and natural-scene datasets. Closed-domain datasets only contain samples in certain scenes, \textit{e.g.,} indoor scenes~\cite{nyu,scannet,irs}, office scenes~\cite{tum}, and autonomous driving~\cite{kitti}. To enhance the diversity of samples, natural-scene datasets are proposed, which use computer-rendered videos~\cite{sintel,tata} or crawl stereoscopic videos from YouTube~\cite{wsvd}. However, the scene diversity and scale of these datasets are still very limited for training robust video depth estimation models that can predict consistent depth in the wild. For instance, WSVD~\cite{wsvd}, which shares a few similar data annotation steps with the proposed \data{} dataset, only contains $533$ YouTube videos with varied quality and insufficient diversity. Sintel~\cite{sintel} only contains $23$ animated videos. To better train and benchmark video depth models, we propose our \data{} dataset with $14@203$ videos from $4$ different data sources. To the best of our knowledge, our \data{} dataset is currently the largest video depth dataset in the wild with the most diverse scenes.

\noindent\textbf{Lightweight Depth Models.} Some lightweight single-image depth models are proposed for real-time applications. Wang~\textit{et al.}~\cite{zhuzhu1} leverage a large teacher model to improve the accuracy of the student network by distillation~\cite{kd1,kd2}. MiDaS~\cite{MiDaS} mixes multiple training data in diverse domains to enhance the generality of their MiDaS-Small~\cite{MiDaS} model. Birkl~\textit{et al.}~\cite{MiDaSV31} utilize the lightweight backbone SwinV2-Tiny~\cite{swin2} to achieve balanced speed and accuracy. Due to the limited model capacity, these lightweight single-image models also suffer from inaccurate and inconsistent predictions on video data. The key challenge is to stabilize these lightweight models and maintain the high efficiency, which is essential for real-time depth-based video applications. However, the prevailing lightweight video depth model ST-CLSTM~\cite{ST-CLSTM} is independent and still produces obvious flickers. For real-time consistent video depth estimation, we implement our \ssmall{}, which can stabilize different lightweight depth predictors in a pluggable manner and achieve real-time throughput of over $30$ fps. To satisfy different applications, we also provide a comprehensive \sx{} model family from small to large scales. 

\noindent\textbf{Video Semantic Segmentation.} Video semantic segmentation and depth estimation are two principal tasks in video dense prediction~\cite{dpt}. These two tasks jointly support applications like virtual
reality and autonomous driving. Similar to video depth estimation~\cite{fmnet,CVD,rcvd,dycvd,vita,ST-CLSTM}, temporal consistency is of vital importance for video semantic segmentation~\cite{cffm,etc,mrcfa,SSLTM}. Various approaches have been proposed to predict accurate and consistent segmentation maps. ETC~\cite{etc} utilizes knowledge distillation~\cite{kd0,kd1,kd2} and optical flow~\cite{flownet2} to supervise the consistency of segmentation maps. TMANet~\cite{tmanet} adopts self-attention~\cite{transformer} to build inter-frame correlations. MRCFA~\cite{mrcfa} mines the temporal relations by multi-scale affinities across adjacent frames. CFFM~\cite{cffm} leverages the feature assembling and cross-frame mining modules to build video contextual relations. IFR~\cite{ifr} utilizes the unlabeled frames to reconstruct the features of labeled frames within a video. SSLTM~\cite{SSLTM} simultaneously uses adjacent and distant frames to construct short- and long-term correlations. However, these prior arts are still stand-alone models, which can not enforce temporal consistency on different single-image semantic segmentation models. To this end, we naturally extend \sx{} to video semantic segmentation and prove the versatility of our framework. Single-image segmentation models are considered as initial semantic segmenters. \sx{} can stabilize different semantic segmenters~\cite{segformer,oneformer} in the effective plug-and-play paradigm.

\noindent\textbf{Blind Video Consistency and Deflickering.} Blind video consistency~\cite{blind1,blind2,blind3} constructs general approaches for extending image processing algorithms to videos with consistency, including colorization, enhancement, style transfer, and intrinsic decomposition. Bonneel~\textit{et al.} propose a gradient-domain technique to achieve blindness and generality of different tasks. Lai~\textit{et al.}~\cite{blind2} adopt a deep recurrent network~\cite{lstm} for temporal consistency. Lei~\textit{et al.}~\cite{blind3} leverage the deep video prior and reweighted training strategy to address the inconsistency. All-In-One-Deflicker~\cite{deflicker1} further advances the task by removing additional guidance of manual annotations and extra consistent videos. Discussing these methods could be illuminating for video depth estimation.

\section{\sx{} Framework}

As shown in \reffig{}~\ref{fig:pipeline}, the \sx{} framework consists of a depth predictor and a \sbn{}. The depth predictor predicts the initial flickering depth for each frame. The \sbn{} converts the depth maps into temporally consistent ones. Our \sx{} framework can coordinate with any off-the-shelf single-image depth models as depth predictors. We also devise a bidirectional inference strategy with \flow{} to further enlarge the temporal receptive field and enhance consistency. 

\subsection{\SbN{}}
The \sbn{} takes RGB frames along with initial depth maps as inputs. A backbone~\cite{segformer} encodes the input sequences into depth-aware features. The next step is to build inter-frame correlations. We use a cross-attention module to refine the depth-aware features with temporal information from relevant frames.
Finally, the refined features are fed into a decoder which restores depth maps with temporal consistency.

\noindent \textbf{Depth-aware Feature Encoding.}
Our \sbn{} works with a sliding window: each frame refers to a few previous frames, serving as reference frames, to stabilize the depth. We denote the frame to be predicted as the target frame. Each sliding window consists of four frames.

Due to the varied scale and shift of different depth predictors, the initial depth maps within a sliding window $\mathbf{F}=\left\{ F_1,F_2,F_3,F_4 \right\} $ should be normalized into $F_i^{norm}$:
\begin{equation}
    F_i^{norm}=\frac{F_i-\min \left( \mathbf{F} \right)}{\max \left( \mathbf{F} \right) -\min \left( \mathbf{F} \right)}\,, i\in \left\{ 1,2,3,4 \right\} .
    \label{eq:norm}
\end{equation}

Then, the normalized depth maps are concatenated with the RGB frames to form a RGB-D sequence. We use a transformer backbone~\cite{segformer} to encode the RGB-D sequence into depth-aware feature maps.

\noindent \textbf{Cross-attention Module.}
With the depth-aware features, the subsequent phase entails the establishment of inter-frame correlations. We leverage a cross-attention module to build temporal and spatial dependencies across pertinent video frames.
Specifically, in the cross-attention module, the target frame selectively attends the relevant features in the reference frames to facilitate depth stabilization. Pixels in the target frame feature maps serve as the query in the cross-attention operation~\cite{transformer}, while the keys and values are generated from the reference frames.

Computational cost can become prohibitively high when employing cross-attention for each position in depth-aware features. Hence, we utilize a patch merging strategy~\cite{VIT} to down-sample the target feature map. Besides, we also restrict the cross-attention into a local window, whereby each token in the target features can only attend a local window in the reference frames. 
Let $T$ denote the depth-aware feature of the target frame, while $R_1, R_2$ and $R_3$ represent the features for the three reference frames. $T$ is partitioned into $7\times 7$ patches with no overlaps; each patch is merged into one token $\mathbf{t}\in \mathbb{R} ^{c}$ , where $c$ is the dimension. For each $\mathbf{t}$, we conduct a local window pooling on $R_1, R_2,$ and $R_3$ and stack the pooling results into $R_p \in\mathbb{R}^{ c\times 3}$. Then, the cross-attention is computed as:
\begin{equation}
\mathbf{t}^{\prime}=\mathrm{softmax} \frac{W_q\mathbf{t}\left( W_kR_p \right) ^T}{\sqrt{c}}W_vR_p\,,
\end{equation}
where $W_q$, $W_k$, and $W_v$ are learnable linear projections. The cross-attention layer is incorporated into a standard transformer block~\cite{transformer} with residual connection and multi-layer perceptron (MLP). We denote the resulting target feature map refined by the cross-attention module as $T_{tem}$. 

Ultimately, a depth decoder with feature fusion modules~\cite{FFM1,FFM2} integrates the depth-aware feature of the target frame ($T$) with the cross-attention refined feature $T_{tem}$, and predicts the consistent depth map for target frame.

\subsection{Training the \SbN{}}
In the training phase, only the \sbn{} is optimized. The depth predictor is the freezed pre-trained DPT-Large~\cite{dpt}. For the \sbn{}, we apply spatial and temporal loss that supervises the depth accuracy and temporal consistency respectively. The training loss can be formulated by:
\begin{equation}
    \mathcal{L} = \sum_{n=2}^{N} \left[\mathcal{L}_s(n-1)+\mathcal{L}_s(n)+\lambda\mathcal{L}_t(n,n-1)\right]\,,
    \label{eq:lossall}
\end{equation}
where $\mathcal{L}_s(n-1)$ and $\mathcal{L}_s(n)$ denote the spatial loss of frame $n-1$ and $n$. $N$ represents the video frame number. $\mathcal{L}_t(n,n-1)$ denotes the temporal loss between frame $n-1$ and $n$.

We adopt the widely-used affinity invariant loss and gradient matching loss~\cite{MiDaS,dpt,mega} as the spatial loss $\mathcal{L}_s$. As for the temporal loss, we adopt the optical flow based warping loss~\cite{MM21,fmnet} to supervise temporal consistency:
\begin{equation}
 \begin{gathered}
    \mathcal{L}_t(n,n-1) = \frac{1}{M}\sum_{j=1}^M O^{(j)}_{n \Rightarrow n-1}|D^{(j)}_{n} - \hat{D}^{(j)}_{n-1}|\,,
\end{gathered}   
\end{equation}
where $|\cdot|$ represents the absolute value function. $\hat{D}_{n-1}$ is the predicted depth $D_{n-1}$ warped by the optical flow $FL_{n \Rightarrow n-1}$.  In our implementation, we adopt GMFlow~\cite{gmflow} for optical flow. $O_{n \Rightarrow n-1}$ is the mask calculated as~\cite{MM21,fmnet}. $M$ denotes pixel numbers. See supplement for more details on loss functions.

\subsection{Bidirectional Inference}
\label{sec:biinfer}
Expanding the temporal receptive range can be beneficial for consistency, \textit{e.g.}, adding more former or latter reference frames. However, directly training the \sbn{} with bidirectional reference frames will introduce large training burdens. To remedy this, we only train the \sbn{} with the former three reference frames. To further enlarge the temporal receptive field and enhance consistency, we introduce a bidirectional inference strategy.

Unlike the training phase, during inference, both the former and latter frames will be used as the reference frames. An additional sliding window is added, where the reference frames are the subsequent three frames of the target.
Let us define the stabilizing process as a function $\mathcal{S}(V_t, \mathbf{V}_r)$, where $V_t$ and $\mathbf{V}_r$ denotes the target RGB-D frame and the reference frames set. The stabilizing function $\mathcal{S}$ does not involve warping-based alignment, as it could lead to incorrect edges and artifacts.
When denoting the RGB-D sequence as $\{V_j |\,j\in{1,2,\cdots, N}\}$, $N$ represents the frame number of a certain video, using this additional sliding window for stabilization can be formulated as:
\begin{equation}
    D^{post}_n = \mathcal{S}(V_{n}, \{V_{n+1},V_{n+2},V_{n+3}\})\,,
\end{equation}
where $V_n$ denotes the target frame. Likewise, using the original sliding window for stabilization can be denoted by:
\begin{equation}
    D^{pre}_n = \mathcal{S}(V_{n}, \{V_{n-1},V_{n-2},V_{n-3}\})\,.
\end{equation}
We ensemble the bidirectional results for a larger temporal receptive field as:
\begin{equation}
    D^{bi}_n = \frac{(D^{pre}_n+D^{post}_n)}{2}\,.
    \label{eq:bisimple}
\end{equation}
$D^{bi}_n$ denotes the depth prediction of the $n^{th}$ frame (target frame). The bidirectional inference can further improve the temporal consistency as demonstrated in~\reftab{}~\ref{tab:sxsx}.

Note that, the cross-attention module is shared by the two sliding windows for inference. Besides, the initial depth maps and depth-aware features are pre-computed. Hence, the bidirectional inference only increases the inference time by $30\%$ compared with single-direction inference and brings no extra computation for the training process. 

The assumption of bidirectional inference is the availability of a few future reference frames, which is primarily applicable to existing videos or streaming videos with some subsequent frames. The strategy brings additional latency, which is not desirable in online applications. Thus, it is deemed optional to further enhance consistency during test time, since our stabilization network can already produce highly consistent results only with forward predictions. To improve our applicability in real-time applications, we develop the lightweight \ssmall{} model with end-to-end real-time throughput of over $30$ fps. 


\begin{figure}[!t]
\begin{center}
   \includegraphics[width=0.47\textwidth,trim=3 0 3 0,clip]{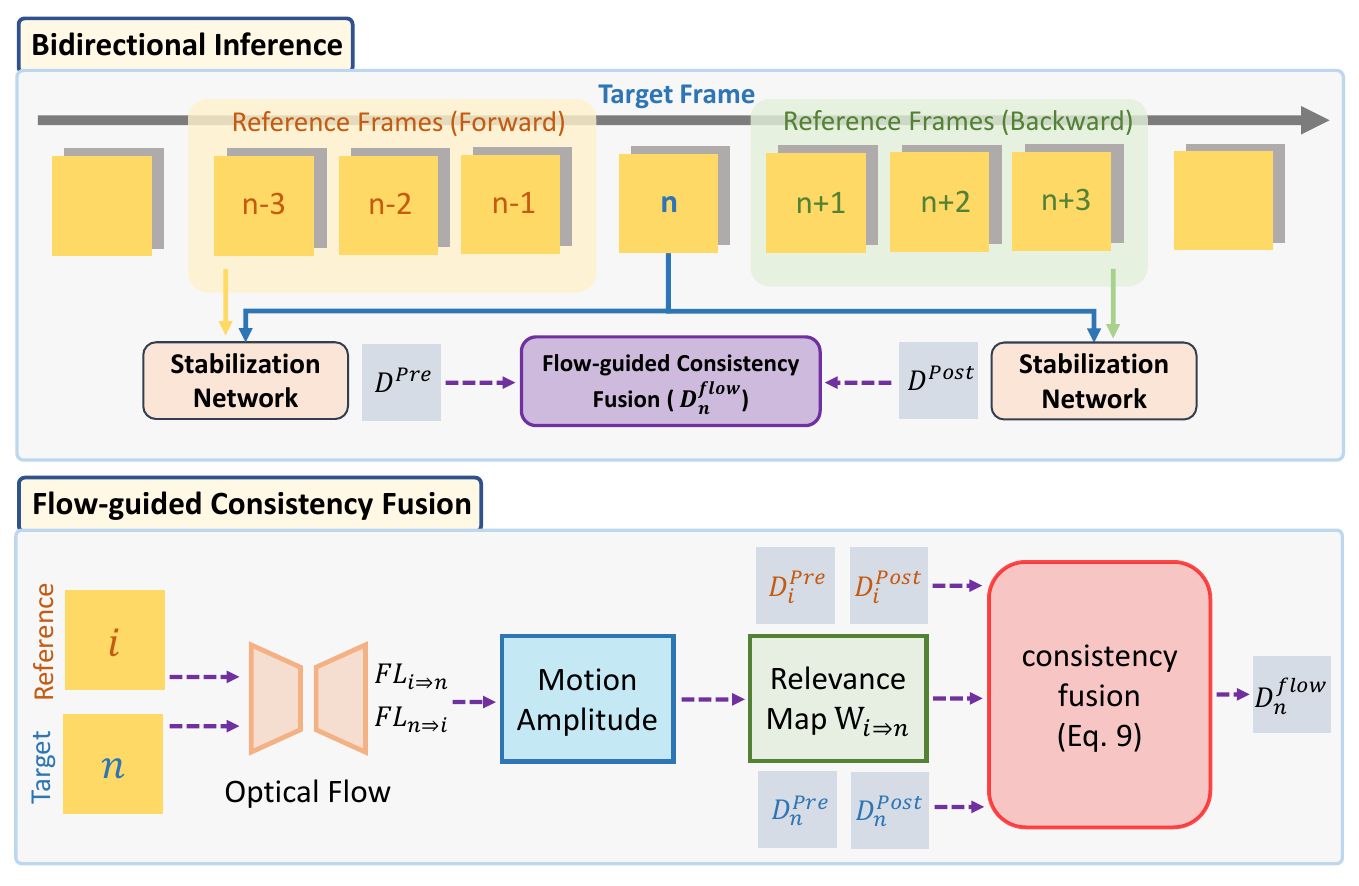}
\end{center}
\vspace{-8pt}
   \caption{
   \textbf{Bidirectional inference strategy with \flow{}}.  Frames or pixels displaying significant motions are considered less relevant to the final target depth. We adaptively fuse the bidirectional depth outcomes of the reference and target frames according to the motion amplitudes and relevance maps derived from optical flow~\cite{gmflow,flownet2}. This technique can further extend the temporal receptive range and improve the consistency.}
\label{fig:biflowqkv1}
\vspace{-8pt}
\end{figure}

\begin{figure*}[!t]
\begin{center}
   \includegraphics[width=0.97\textwidth,trim=0 0 0 0,clip]{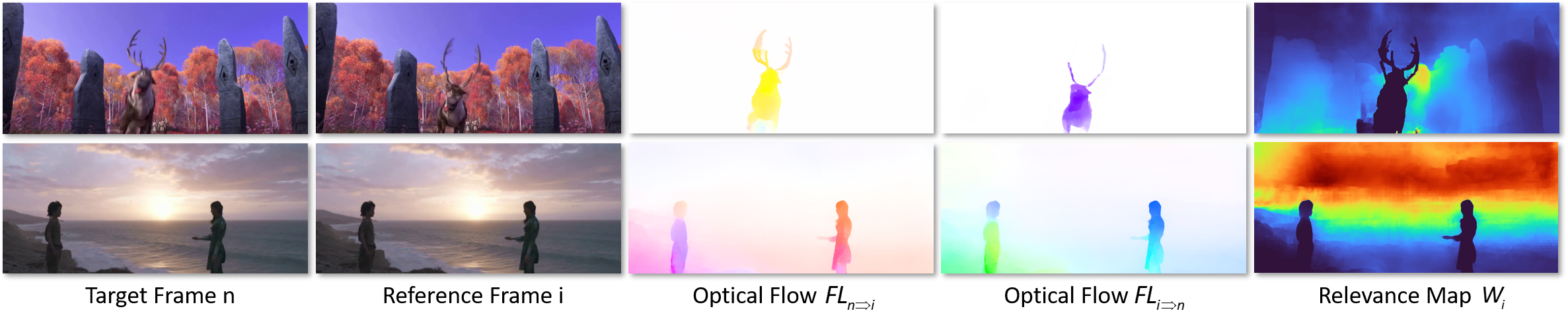}
\end{center}
\vspace{-12pt}
   \caption{
   \textbf{Visualizations of flow-guided fusion.} We visualize the intermediate results of \flow{}. For the relevance maps $W_i$, brighter colors indicate higher values, while darker colors indicate lower fusion weights.}
\label{fig:flowreal}
\vspace{-12pt}
\end{figure*}

\subsection{\FLOW{}}
\label{sec:flownew}

The bidirectional inference strategy can enhance
the temporal consistency based on the forward and backward predictions. The direct averaging as \refequ{}~\ref{eq:bisimple} is actually a straightforward yet effective way to merge $D^{pre}_n$ and $D^{post}_n$ using fixed weights, which expands the temporal receptive range without introducing heavy computational costs. However, compared with the direct averaging, using adaptive weights to fuse the bidirectional outcomes of reference and target frames can be more reasonable, since frames or pixels that exhibit larger motions relative to the target frame should be deemed less relevant for the final target depth.


To this end, we advance the bidirectional inference with a \flow{} strategy, which enables the adaptive fusion of bidirectional depth results from reference and target frames. As shown in \reffig{}~\ref{fig:biflowqkv1}, we utilize optical flow~\cite{gmflow,flownet2} to assess the pixel-wise motion amplitudes and relevance maps between reference and target frames. The bidirectional optical flow $FL_{i \Rightarrow n}$ and $FL_{n \Rightarrow i}$ are computed between the reference frame $i\in\{n\pm 3,n\pm 2,n\pm 1\}$ and the target frame $n$. Motion amplitude between reference and target frames can be estimated by the magnitude (\textit{i.e.}, Frobenius norm) of optical flow~\cite{gmflow,flownet2}. Pixels with larger motions in the reference frames tend to have lower relevance to the corresponding pixels in the target frame. To quantify this, we calculate the pixel-wise adaptive relevance map $W_{i}$ as follows:

\begin{equation}
    W_{i} = \exp{}\left[-\alpha\cdot(||FL_{i \Rightarrow n}||_F + ||FL_{n \Rightarrow i}||_F)\right]\,,
    \label{eq:biflow}
\end{equation}
where the coefficient $\alpha$ serves to restrict the weights assigned to pixels exhibiting significant motions. A larger value of $\alpha$ results in reduced weights for those pixels with more pronounced motions in the reference frames.

The final depth results of target frame $n$ with \flow{} can be articulated by:

\begin{equation}
    D_n^{flow} = \beta \cdot D^{bi}_n + (1-\beta) \cdot \sum_{i}(W_{i}\cdot(D_i^{pre}+D_i^{post}))\,,
    \label{eq:bimix}
\end{equation}
in which $\beta$ and $1-\beta$ represent the weights of the bidirectional averaging as \refequ{}~\ref{eq:bisimple} and the enhanced \flow{}, respectively. The second term of flow-guided consistency fusion solely incorporates information from the reference frames. Thus, we integrate the above two terms and further achieve an improvement in the temporal consistency beyond that of simple averaging, as demonstrated in \reftab{}~\ref{tab:sxsx}.

For better understanding, we visualize some intermediate results in \reffig{}~\ref{fig:flowreal}. For the first sample, a deer runs quickly across the scene. The relevance map accurately identifies the positions of the moving deer in both frames (\textit{e.g.}, the overlapping antlers) and produces low fusion weights in the moving areas. For the second sample, walking people are also assigned low relevance values. These regions exhibit significant motion, leading to pixel misalignment and low depth relevance. Thus, our strategy tends not to fuse the unreliable depth of reference frames, but preserves the original bidirectional depth $D_n^{bi}$ of the target frame. In this way, our method can improve the consistency without introducing errors in the presence of motion.

\subsection{\sx{} Real-time Model and Model Family}

The application paradigms of our \sx{} can be categorized into two distinct aspects. On the one hand, in pursuit of the best performance on both the spatial accuracy and temporal consistency, \llarge{} utilizes various top-performing depth models~\cite{MiDaS,dpt,newcrfs} as the single-image depth predictors during inference. On the other hand, for real-time processing and applications, \ssmall{} employs different lightweight depth predictors~\cite{MiDaSV31,MiDaS}.

To develop the \ssmall{} model, we use a lightweight attention-based backbone~\cite{segformer} to encode the depth-aware features. Compared with \llarge{}, the attention layers and the token embedding dimensions of \ssmall{} are reduced. Additionally, we have implemented model pruning~\cite{jianzhi} on \ssmall{} to further boost its efficiency. As detailed in \refsec{}~\ref{sec:spd}, \ssmall{} achieves a real-time processing speed of over $30$ fps, enforcing temporal consistency and significantly surpassing previous lightweight depth models~\cite{ST-CLSTM,MiDaS,MiDaSV31} in performance. Please refer to \refsec{}~\ref{sec:imd} for more implementation details of the large and small models.


\subsection{Extension to Video Semantic Segmentation}
Analogous to video depth estimation~\cite{CVD,rcvd,fmnet}, the temporal consistency is equally crucial for video semantic segmentation~\cite{etc,cffm,mrcfa,ifr,tmanet,SSLTM}. It is a logical progression for us to extend our \sx{} framework to the video semantic segmentation task.

We firstly delineate the inputs and outputs. With RGB frames as inputs, video semantic segmentation models~\cite{etc,cffm,mrcfa,ifr,tmanet,SSLTM} are trained to predict the per-pixel probability $\mathcal{P}$ with $C$ channels for each frame, in which $C$ refers to the number of semantic classes. The one-channel semantic label predictions $\mathcal{Q}$ are then derived by performing an $Argmax$ operation on the channel dimension of $\mathcal{P}$.

In our case, we employ various pre-existing single-image semantic segmentation models~\cite{segformer,oneformer} as the initial semantic segmenter. Our objective is to enhance the temporal consistency from the initial flickering results in a plug-and-play manner. Within a sliding window, the RGB frames are concatenated with the one-channel label predictions $\mathcal{Q}$ from semantic segmenters, serving as the input for our \sbn{}. We omit the normalization operation in \refequ{}~\ref{eq:norm}, as the label predictions $\mathcal{Q}$ from different segmenters share a uniform data range and format. The output of our \sbn{} is adjusted to $C$ channels as the probability predictions. We apply a common semantic decoder~\cite{segformer,cffm} for output and the cross entropy (CE) loss for supervision. Moreover, the bidirectional inference and the \flow{} are conducted on the probability predictions, considering that merging multiple integral semantic labels would be unsuitable.

Instead of the more fine-grained label probability distributions $\mathcal{P}$, \sx{} works with the one-channel label predictions $\mathcal{Q}$ for two main reasons. Firstly, the channel number $C$ of $\mathcal{P}$ equals the number of semantic classes, which can be large in practice~\cite{cityscapes,ade20k}. Using $\mathcal{P}$ as the multi-frame input significantly increases computational costs during training and inference. Besides, the channel number $C$ differs across datasets and scenarios. The varying input channels could restrict the uniformity and applicability from the perspective of model design. Therefore, \sx{} stabilizes the one-channel $\mathcal{Q}$ predicted by initial segmenters~\cite{segformer,oneformer}.

With similar settings of the sliding windows, the cross-attention module, and the inference protocol as in video depth estimation, our \sx{} can effectively stabilize different semantic segmenter~\cite{oneformer,segformer} in the pluggable paradigm. We attain \sota{} performance for video semantic segmentation on \city{} dataset~\cite{cityscapes}. With minor modifications, the efficacy of our \sx{} in both the depth estimation and semantic segmentation underscores the adaptability and versatility of our framework. 

\subsection{Implementation Details}
\label{sec:imd}
\noindent \textbf{Depth and Disparity.} For all our implementations of video depth, the inputs and outputs of \sx{} are disparity maps, \textit{i.e.}, the inverse of depth. The models are also supervised by the disparity ground truth from the VDW dataset. We illustrate the reasons for using disparity in the supplement.

\noindent \textbf{Model Architecture.} The disparity maps from DPT-Large~\cite{dpt} are used as the input of large and small models during training. For each target frame, we use three reference frames with inter-frame intervals $l=1$. \textbf{\llarge{}} adopts DPT-Large~\cite{dpt}, MiDaS-v2.1-Large~\cite{MiDaS}, and NeWCRFs~\cite{newcrfs} as depth predictors during inference. MiDaS-v2.1-Large~\cite{MiDaS} is the same depth model as test-time-training-based methods~\cite{CVD,rcvd,dycvd} for fair comparisons. Mit-b5~\cite{segformer} is adopted as the backbone to encode depth-aware features. The token embedding dimension $c$ of \llarge{} is $256$. \textbf{\ssmall{}} adopts lightweight depth predictors DPT-Swin2-Tiny~\cite{MiDaSV31} and MiDaS-v2.1-Small~\cite{MiDaS}. It utilizes Mit-b0~\cite{segformer} as the backbone and the token embedding dimension $c=128$. The small model only adopts the forward prediction for online and real-time applications.

\noindent \textbf{Training Recipe.} All frames are resized so that the shorter side equals $384$, and then randomly cropped to $384\times 384$ for training. In each epoch, we randomly sample $72@000$ input sequences. Note that the sampled frames in each epoch do not overlap. We use Adam optimizer to train the model for $30$ epochs with a batchsize of $9$. The initial learning rate is $6\times 10^{-5}$ and decreases by $1\times10^{-5}$ for every five epochs. When finetuning our model on NYUDV2~\cite{nyu} and KITTI~\cite{kitti} datasets, we use a learning rate of $1\times10^{-5}$ for only one epoch. In all experiments, the temporal loss weight $\lambda$ in \refequ{}~\ref{eq:lossall} is set to $0.2$. The coefficient $\alpha$ in \refequ{}~\ref{eq:biflow} is $10$. The weight $\beta$ of \flow{} in \refequ{}~\ref{eq:bimix} is $0.5$. 

\noindent \textbf{Video Semantic Segmentation.} The backbone and model parameters significantly influence the performance of semantic segmentation. Thus, we follow the experimental settings of the previous \sota{} video semantic segmentation method SSLTM~\cite{SSLTM} for fair comparisons with similar amounts of parameters. Identical to SSLTM~\cite{SSLTM}, we implement our model and compare different methods using ResNet-50~\cite{resnet} as the backbone. For the training of our model, only the SegFormer-B1~\cite{segformer} is used as the initial semantic segmenter. During inference, different single-image semantic segmenters are used for the plug-and-play paradigm, including SegFormer-B1, SegFormer-B3~\cite{segformer}, and OneFormer~\cite{oneformer}. Other settings such as reference frames and inter-frame intervals are the same as video depth. 

The ResNet-50~\cite{resnet} backbone is initialized with ImageNet-pretrained~\cite{imagenet} weight. Other parts are initialized randomly. We follow the standard training recipe as prior arts~\cite{SSLTM,cffm,mrcfa,etc} on \city{} dataset~\cite{cityscapes}. Specifically, frames are cropped to $512\times 1024$ for training and resized to the same resolution during inference. We adopt the commonly-applied single-scale test~\cite{SSLTM,cffm,mrcfa} for simplicity. The Adam optimizer is used to train the model for $30$ epochs with a batchsize of $4$. The learning rate, the temporal loss weight $\lambda$ in \refequ{}~\ref{eq:lossall}, the coefficient $\alpha$ in \refequ{}~\ref{eq:biflow}, and the weight $\beta$ of bidirectional inference in \refequ{}~\ref{eq:bimix} are all identical to video depth estimation. 

\begin{table}
\caption{\textbf{Comparisons of video depth datasets.}  The 3D Movies dataset of MiDaS~\cite{MiDaS} is not released
and only contains 75k images but not videos. TartanAir~\cite{tata} only has some limited dynamic scenes (\textit{e.g.}, fish in the ocean sequence). Most videos in TartanAir~\cite{tata} lack major dynamic objects (\textit{e.g.}, pedestrians). For example, models
trained on TartanAir cannot predict satisfactory results
in scenes with moving people as such scenes are rare. Our \data{} dataset shows advantages in diversity and quantity. }
\vspace{-10pt}
\label{tab:dacp}
\begin{center}
\addtolength{\tabcolsep}{-4.2pt}
\resizebox{\columnwidth}{!}{
\begin{tabular}{llccccccc}
\toprule
 Type &  Dataset & Videos & Frames($k$) & Indoor & Outdoor & Dynamic & Resolution \\
\midrule
\multirow{4}{*}{Closed} & NYUDV2~\cite{nyu} & $464$ & $407$ & $\usym{2714}$ & $\usym{2717}$ & $\usym{2717}$ & $640\times480$ \\
& KITTI~\cite{kitti} & $156$ & $94$ & $\usym{2717}$ & $\usym{2714}$ & $\usym{2714}$ & $1224\times370$ \\
\multirow{2}{*}{Domains} & TUM~\cite{tum} & $80$ & $128$ & $\usym{2714}$ & $\usym{2717}$ & $\usym{2714}$ & $640\times 480$ \\
& IRS~\cite{irs} & $76$ & $103$ & $\usym{2714}$ & $\usym{2717}$ & $\usym{2717}$ & $960\times 540$ \\
& ScanNet~\cite{scannet} & $1@513$ & $2@500$ & $\usym{2714}$ & $\usym{2717}$ & $\usym{2717}$ & $640\times 480$ \\
\midrule
\multirow{1}{*}{CG} & Sintel~\cite{sintel} & $23$ & $1$ & $\usym{2714}$ & $\usym{2714}$ & $\usym{2714}$ & $1024\times 436$ \\
\multirow{1}{*}{Rendered} & TartanAir~\cite{tata} & $1@037$ & $1@000$ & $\usym{2714}$ & $\usym{2714}$ & $\usym{2717}$ & $640\times 480$ \\
\midrule
\multirow{2}{*}{Natural} & MiDaS~\cite{MiDaS} & $\usym{2717}$ & $75$ & $\usym{2714}$ & $\usym{2714}$ & $\usym{2714}$ & $1880\times 800$ \\
 \multirow{2}{*}{Scenes} &WSVD~\cite{wsvd} & $553$ & $1@500$ & $\usym{2714}$ & $\usym{2714}$ & $\usym{2714}$ & $\sim 720p$ \\
& Ours & $14@203$ & $2@237$ & $\usym{2714}$ & $\usym{2714}$ & $\usym{2714}$ & $1880\times800$ \\
\bottomrule
\end{tabular}
}
\end{center}
\vspace{-12pt}
\end{table}

\section{\data{} Dataset}
As mentioned in \refsec{}~\ref{intro}, current video depth datasets are limited in both diversity and volume. To compensate for the data shortage and boost the performance of learning-based video depth models, we elaborate a large-scale natural-scene dataset, Video Depth in the Wild (VDW). To our best knowledge, our \data{} dataset is currently the largest video depth dataset with the most diverse video scenes.

\begin{figure}[!t]
\begin{center}
   \includegraphics[width=0.47\textwidth,trim=10 0 10 0,clip]{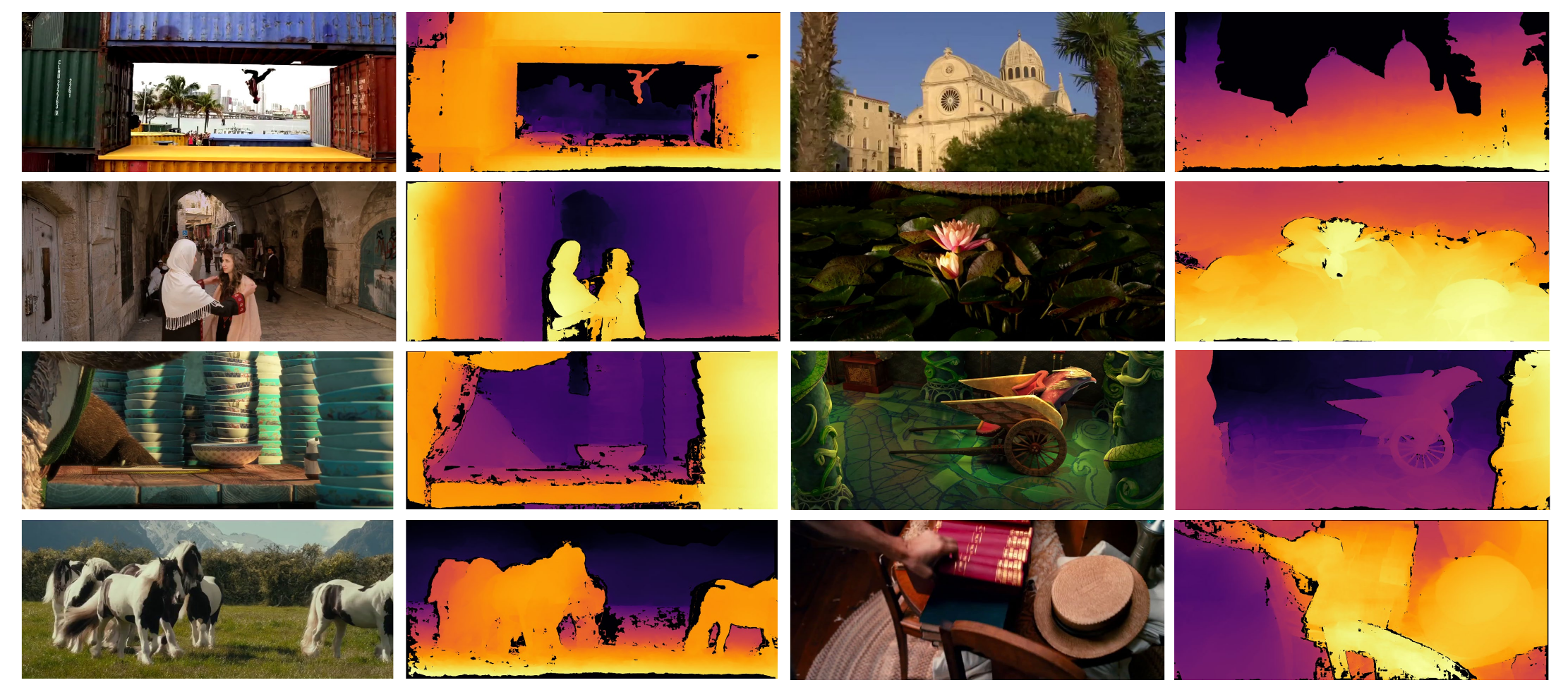}
\end{center}
\vspace{-7pt}
   \caption{
   \textbf{Examples of our \data{} dataset.}
   Four rows are from web videos, documentaries, animations, and movies, respectively. Sky regions and invalid pixels are masked out.}
\label{fig:datagt}

\end{figure}

\noindent\textbf{Dataset Construction.} We collect stereo videos from four data sources: movies, animations, documentaries, and web videos. A total of $60$ movies, animations, and documentaries in Blu-ray format are collected. We also crawl $739$ web stereo videos from YouTube with the keywords such as ``stereoscopic'' and ``stereo''. To balance the realism and diversity, only $24$ movies, animations, and documentaries are retained. For instance, ``Seven Wonders of the Solar System'' is removed as it contains many virtual scenes. The disparity ground truth is generated with two main steps: sky segmentation and optical flow estimation. A model ensemble method is adopted to remove errors and noises in the sky masks, which can improve the quality of the ground truth and the performance of the trained models, especially on sky regions as shown in \reffig{}~\ref{fig:qpqp}. A \sota{} optical flow model GMFlow~\cite{gmflow} is used to generate the disparity ground truth. Finally, a rigorous data cleaning procedure is conducted to filter the videos that are not qualified for our dataset. \reffig{}~\ref{fig:datagt} shows some examples of our \data{} dataset. 

\begin{table*}[!t]
    \caption{
\textbf{Comparisons with the state-of-the-art approaches.} We test the total time of processing eight $640\times480$ frames on one NVIDIA RTX A6000 GPU. For our \sx{}, we report the performance of our large model with complete designs, \textit{i.e.}, \llarge{} with \flow{}. The best performance is in boldface. Second best is underlined.}
\vspace{-7pt}
\label{tab:bigone}

    \begin{center}
    \addtolength{\tabcolsep}{-3pt}
    \resizebox{\textwidth}{!}{
    \begin{tabular}{llcccccccccccccccc}
    \toprule
    \multirow{2}{*}{Type} & \multirow{2}{*}{Method} & \multirow{2}{*}{Time($s$)}  &
    \multicolumn{3}{c}{\data{}} &
    \multicolumn{5}{c}{Sintel} &
    \multicolumn{3}{c}{NYUDV2} &
    \multicolumn{4}{c}{KITTI} \\
    \cmidrule{4-6} \cmidrule{8-10} \cmidrule{12-14} \cmidrule{16-18}
        & & & $\delta_1\uparrow$ & $Rel\downarrow$ &
        $OPW\downarrow$ & & 
        $\delta_1\uparrow$ & $Rel\downarrow$ & $OPW\downarrow$ & & 
        $\delta_1\uparrow$ & $Rel\downarrow$ & $OPW\downarrow$ & & 
        $\delta_1\uparrow$ & $Rel\downarrow$ & $OPW\downarrow$ \\
    \midrule
    \multirow{1}{*}{Single} & MiDaS-v2.1-Large~\cite{MiDaS}    &$0.76$&$0.651$& $0.288$ &$0.676$ &&$0.485$& $0.410$ & $0.843$ && $0.910$& $0.095$ &$0.862$ && $0.940$ & $0.088$ & $0.602$ \\
    \multirow{1}{*}{Image} & DPT-Large~\cite{dpt}        &$0.97$&$\underline{0.730}$&$\underline{0.215}$ &$0.470$ &&$\textbf{0.597}$&$\underline{0.339}$ & $0.612$&& $0.928$&$0.084$ &$0.811$ && $0.964$ & $0.069$ & $0.585$  \\
    \midrule
    \multirow{2}{*}{Test-time} & CVD~\cite{CVD}    &$352.58$&$-$& $-$ &$-$ &&$0.518$& $0.406$ & $0.497$ && $-$& $-$ &$-$ && $0.878$& $0.114$ &$0.374$  \\
    \multirow{2}{*}{Training} & Robust-CVD~\cite{rcvd}        &$270.28$ &$0.676$&$0.261$ &$0.279$ &&$0.521$&$0.422$ & $0.475$&& $0.886$&$0.103$ &$0.394$ &&$0.901$& $0.097$ &$0.338$ \\
    & Zhang \textit{et al.}~\cite{dycvd} &$464.83$&$-$& $-$&$-$ &&$0.522$& $0.342$& $0.481$&& $-$& $-$&$-$ &&$-$& $-$ &$-$ \\
    \midrule
     & ST-CLSTM~\cite{ST-CLSTM}    &$0.58$&$0.477$& $0.521$ &$0.448$ &&$0.351$& $0.517$ & $0.585$ && $0.833$& $0.131$ &$0.645$ && $0.890$& $0.101$ &$0.413$ \\
    \multirow{6}{*}{Learning} & Cao \textit{et al.}~\cite{MM21}        &$-$&$-$&$-$ &$-$ &&$-$&$-$ & $-$&& $0.835$&$0.131$ &$-$ && $0.872$&$0.109$ &$-$ \\
    \multirow{6}{*}{Based} & FMNet~\cite{fmnet}        &$3.87$ &$0.472$&$0.514$ &$0.402$ &&$0.357$&$0.513$ & $0.521$&& $0.832$&$0.134$ &$0.387$ && $0.886$&$0.099$ &$0.375$ \\
    & DeepV2D~\cite{deepv2d}&$68.71$ &$0.546$& $0.528$&$0.427$ &&$0.486$& $0.526$& $0.534$&& $0.924$& $0.082$&$0.402$ && $0.972$& $0.051$&$0.428$\\
    & WSVD~\cite{wsvd} &$4.25$&$0.637$& $0.314$&$0.462$ &&$0.501$& $0.439$& $0.577$&& $0.768$& $0.164$&$0.683$ && $0.812$& $0.156$&$0.497$\\
    & Li \textit{et al.}~\cite{icmr23} &$-$&$-$& $-$&$-$&&$0.475$& $0.389$& $-$&& $-$& $-$&$-$ && $-$& $-$&$-$ \\
    & ViTA~\cite{vita} &$1.38$&$0.689$& $0.243$&$0.252$ &&$0.554$& $0.376$& $0.492$&& $0.922$& $0.092$&$0.385$ && $0.912$& $0.095$&$0.316$ \\
    & MAMo~\cite{mamo} &$-$&$-$& $-$&$-$ &&$-$& $-$& $-$&& $0.919$& $0.094$&$-$ &&$\underline{0.977}$& $0.049$& $-$\\
    & Ours-Large(MiDaS-v2.1-Large) &$1.92$&$0.701$& $0.239$&$\underline{0.148}$ &&$0.532$& $0.372$& $\underline{0.447}$&& $\underline{0.941}$& $\underline{0.076}$&$\underline{0.347}$&& $0.970$& $\underline{0.049}$&$\underline{0.258}$ \\
    & Ours-Large(DPT-Large) &$2.31$&$\textbf{0.742}$& $\textbf{0.208}$&$\textbf{0.129}$ &&$\underline{0.591}$& $\textbf{0.335}$& $\textbf{0.403}$&& $\textbf{0.950}$& $\textbf{0.072}$&$\textbf{0.339}$
    && $\textbf{0.982}$& $\textbf{0.046}$&$\textbf{0.233}$ \\
    \bottomrule
    \end{tabular}
    }
\end{center}
\vspace{-10pt}
\end{table*}

\begin{figure}[!t]
\begin{center}
   \includegraphics[width=0.49\textwidth,trim=10 10 10 10,clip]{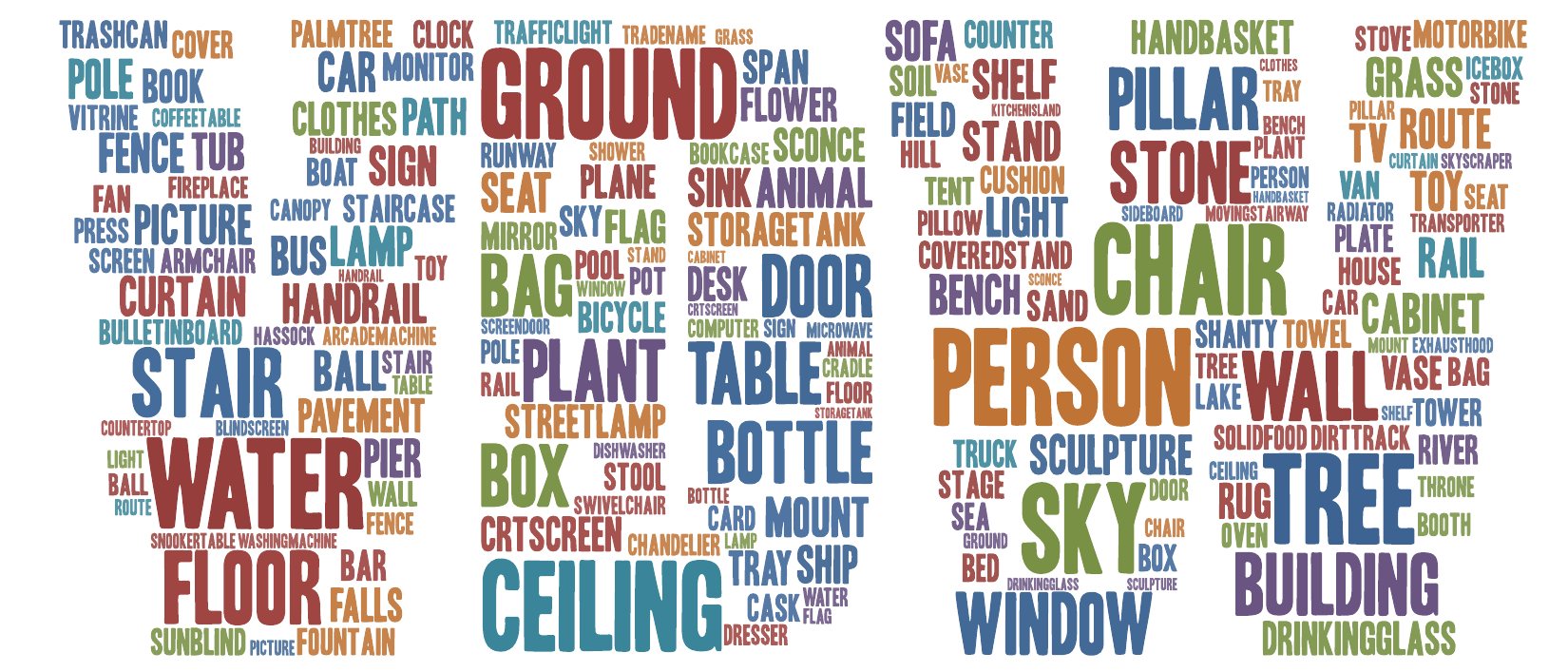}
\end{center}
\vspace{-6pt}
   \caption{\textbf{Objects presented in our \data{} dataset.} We conduct semantic segmentation with Mask2Former~\cite{mask2former} trained on ADE20K~\cite{ade20k}. Refer to our supplementary document for more detailed construction process and data statistics.}
\label{fig:wccc}
\vspace{-10pt}
\end{figure}

\noindent\textbf{Dataset Statistics.}
\label{sec:ds}
\data{} dataset contains $14@203$ videos with a total of $2@237@320$ frames. The total data collection and processing time takes over six months and about $4@000$ man-hours. To verify the diversity of scenes and entities in our dataset, we conduct semantic segmentation by Mask2Former~\cite{mask2former} trained on ADE20K~\cite{ade20k}. All the $150$ categories are covered in our dataset, and each category can be found in at least $50$ videos. \reffig{}~\ref{fig:wccc} shows the word cloud of the 150 categories.
We randomly choose $90$ videos with $12@622$ frames as the test set. The testing videos adopt different data sources from the training data, \textit{i.e.}, different movies, web videos, or animations.
\data{} not only alleviates the data shortage for learning-based approaches, but also serves as a comprehensive benchmark for video depth. 

\noindent\textbf{Comparisons with Other Datasets.} 
\label{sec:dcom}
As shown in \reftab{}~\ref{tab:dacp}, the proposed \data{} dataset has significantly larger numbers of video scenes. Compared with the closed-domain datasets~\cite{nyu,kitti,scannet,tum,irs}, the videos of \data{} are not restricted to a certain scene, which is more helpful to train a robust video depth model. For the natural-scene datasets, our dataset has more than ten times the number of videos as the previous largest dataset WSVD~\cite{wsvd}. Although WSVD~\cite{wsvd} has $1.5M$ frames, the scenes (video numbers) are limited. MiDaS~\cite{MiDaS} also proposes their 3D Movies dataset with in-the-wild images and disparity. Compared with the 3D Movies dataset of MiDaS~\cite{MiDaS}, VDW differs in two main aspects: (1) accessibility; and (2) dataset scale and format. 
MiDaS~\cite{MiDaS} does not provide related metadata (\textit{e.g.}, timestamps) and data generation scripts. In contrast, we have released the comprehensive VDW Dataset Toolkit~\cite{vdwtool} and our metadata, allowing researchers to reproduce VDW or generate new datasets. 
Besides, their 3D Movies dataset~\cite{MiDaS} only contains $75k$ images, while VDW contains 14,203 videos with $2.237M$ frames.
It is also worth noticing that our \data{} dataset has higher resolution and a rigorous data annotation and cleaning pipeline.  We only collect videos over $1080p$ and crop all our videos to $1880\times800$ to remove black bars and subtitles. See supplement for more statistics and details.

\section{Experiments}

To validate the effectiveness and generality of our \sx{} framework, we carry out experiments on the tasks of video depth estimation and video semantic segmentation, which are two key tasks in dense prediction~\cite{dpt}. For video depth estimation, we evaluate \sx{} on four different datasets, which contain videos for real-world and synthetic, static and dynamic, indoor and outdoor. We also demonstrate that our \flow{} can further enhance the temporal consistency by adaptively merging the bidirectional disparity results. In order to prove our efficacy in real-time applications, we compare our \ssmall{} model with other lightweight single-image depth predictors or video depth models in terms of performance and efficiency. For video semantic segmentation, we conduct evaluations on \city{}~\cite{cityscapes} dataset following prior arts~\cite{SSLTM,cffm,mrcfa}. \sx{} achieves \sota{} performance for both tasks, which proves the versatility of our framework.

\subsection{Datasets and Evaluation Protocol}
\label{sec:dataset}
In this section, we illustrate the datasets and evaluation protocols for video depth estimation. Please refer to \refsec{}~\ref{sec:yyfgsy} for the experimental details of video semantic segmentation.
\noindent \textbf{\data{} Dataset}. We use the proposed \data{} as the training data for its diversity and quantity on natural scenes. We also evaluate the previous video depth approaches on the test split of \data{}, serving as a new video depth benchmark.

\noindent \textbf{Sintel Dataset}. Following~\cite{rcvd,dycvd}, we use the final version of Sintel~\cite{sintel} to demonstrate the generalization ability of our \sx{}. We conduct zero-shot evaluations on Sintel~\cite{sintel}. All learning-based methods are not finetuned on Sintel dataset.
%

\noindent \textbf{DAVIS Dataset}. DAVIS~\cite{davis} is a natural-scene dataset for video object segmentation. we also test our \sx{} on the challenging videos from DAVIS~\cite{davis} for qualitative comparisons. Please refer to our demo video for video results.

\noindent \textbf{NYUDV2 Dataset}. Except for natural scenes, a closed-domain NYUDV2~\cite{nyu} is adopted for evaluation. It contains $464$ videos of indoor scenes. We pretrain the \sbn{} on \data{} and finetune the model on NYUDV2~\cite{nyu}. We follow the same train/test split as Eigen~\textit{et al.}~\cite{silog,ST-CLSTM,fmnet,MM21} with $249$ videos for training and $654$ samples from the rest $215$ videos for testing.

\noindent \textbf{KITTI Dataset}. Similar to NYUDV2~\cite{nyu} dataset, we also conduct the pretraining and finetuning protocol on KITTI~\cite{kitti}, which is another major closed-domain video depth dataset. KITTI~\cite{kitti} is captured by cameras and depth sensors mounted on a driving car and consists of $61$ outdoor video scenes. We follow the train/test split as Eigen~\textit{et al.}~\cite{silog,ST-CLSTM,fmnet,MM21} with $32$ videos for training and 697 samples from the rest $29$ videos for testing.


%

\noindent \textbf{Evaluation Metrics.}
We evaluate both the depth accuracy and temporal consistency. For the temporal consistency metric, we adopt the optical flow based warping metric ($OPW$) following FMNet~\cite{fmnet}, which can be computed as:
\begin{equation}
    OPW = \frac{1}{N-1}\sum_{n=2}^{N}\mathcal{L}_t(n,n-1)\,.
\end{equation}
We report the average $OPW$ of all the videos in the test sets. As for the depth metrics, we adopt the commonly-applied $Rel$ and $\delta_i (i=1,2,3)$.



\begin{figure*}[!t]
\begin{center}
   \includegraphics[width=0.97\textwidth,trim=0 0 0 0,clip]{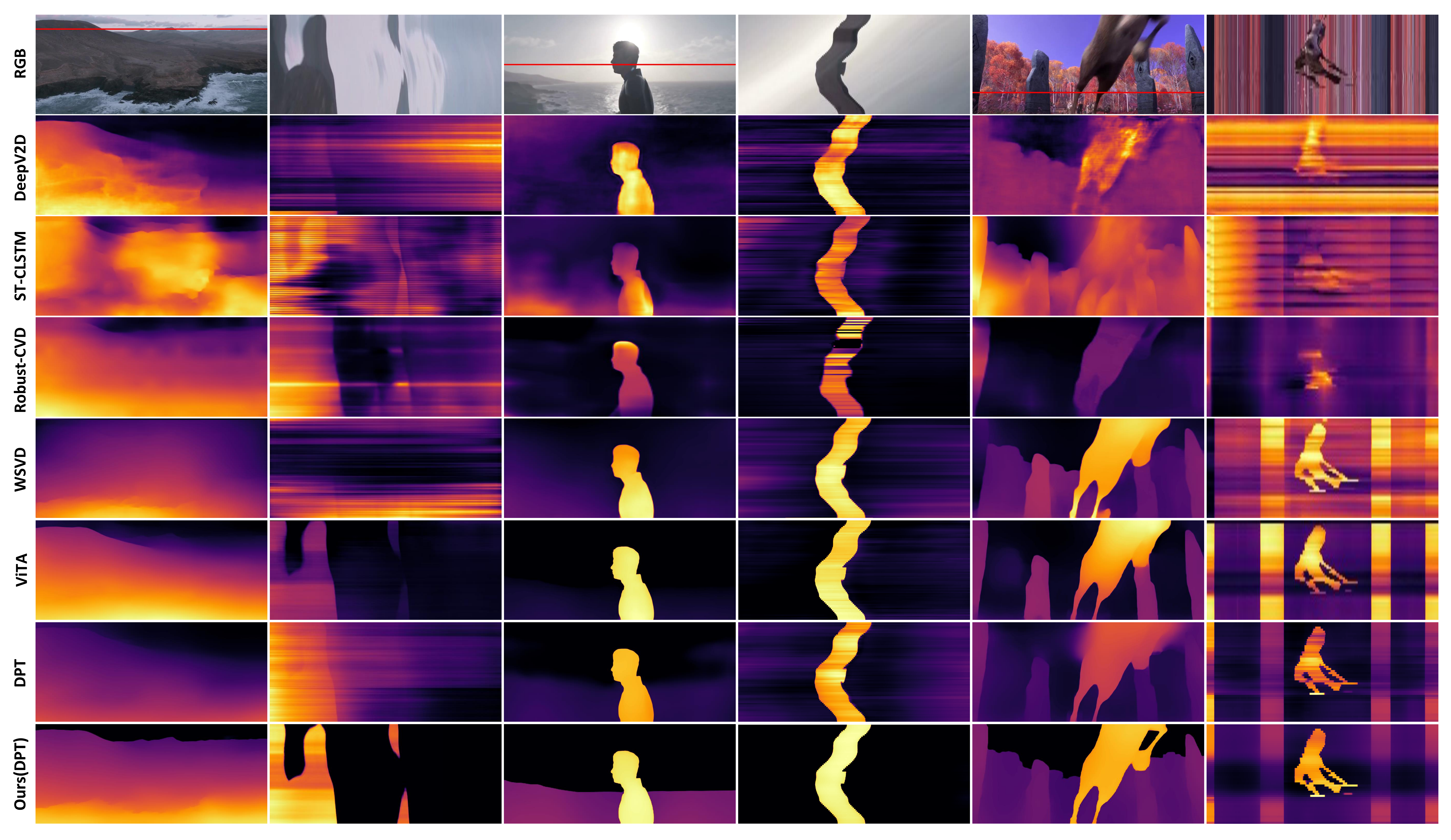}
\end{center}
\vspace{-10pt}
   \caption{
   \textbf{Qualitative comparisons.}
   DeepV2D~\cite{deepv2d} and Robust-CVD~\cite{rcvd} show obvious artifacts in those videos. We draw the scanline slice over time; fewer zigzagging pattern means better consistency. Compared with the other video depth methods, our \sx{} is more robust on natural scenes and achieves better spatial accuracy and temporal consistency.}
\label{fig:qpqp}
\vspace{-7pt}
\end{figure*}

\subsection{Comparisons with Other Video Depth Methods}

\noindent \textbf{Comparisons with the test-time training methods.}
First focus on the test-time training approaches~\cite{CVD,rcvd,dycvd}. As shown in \reftab{}~\ref{tab:bigone}, our learning-based framework outperforms these approaches by large margins in terms of inference speed, accuracy and consistency. Our \sx{} shows at least $6.6\%$ and $7.6\%$ improvements for $\delta_1$ and $OPW$ than Robust-CVD~\cite{rcvd} on \data{}, Sintel~\cite{sintel}, NYUDV2~\cite{nyu}, and KITTI~\cite{kitti}. Our learning-based approach is over one hundred times faster than Robust-CVD~\cite{rcvd}. Our strong performance demonstrates that learning-based frameworks are capable of attaining great performance with much higher efficiency than test-time-training-based ones~\cite{CVD,rcvd,dycvd}.

It is also worth-noticing that test-time-training-based approaches are not robust for natural scenes. CVD~\cite{CVD} and Zhang \textit{et al.}~\cite{dycvd} fail on some videos on \data{} and Sintel~\cite{sintel} due to erroneous pose estimation results. Hence, some of their results are not reported in \reftab{}~\ref{tab:bigone}. Refer to the supplement for more details. Although Robust-CVD~\cite{rcvd} can produce results for all testing videos by jointly optimizing the camera poses and depth, it is still not robust for many videos and produces obvious artifacts as shown in \reffig{}~\ref{fig:qpqp}.

\noindent \textbf{Comparisons with the learning-based methods.} The proposed \sx{} also attains better accuracy and consistency than previous learning-based approaches~\cite{ST-CLSTM,fmnet,MM21,wsvd,icmr23,deepv2d} on all the four datasets, including natural scenes and closed domain. As shown in \reftab{}~\ref{tab:bigone}, on our \data{} and Sintel with natural scenes, the proposed \sx{} shows obvious advantages: improving $\delta_1$ and $OPW$ by over $9\%$ and $18.6\%$ compared with previous learning-based methods. Note that, our \sx{} can benefit from stronger single-image models and obtain better performance, which will be discussed and proved in \reftab{}~\ref{tab:blab}.  

To better compare \sx{} with previous learning-based methods, we only use NYUDV2~\cite{nyu} or KITTI~\cite{kitti} as training and evaluation data for comparisons. As shown in \reftab{}~\ref{tab:nyu}, the \sx{} improves the FMNet~\cite{fmnet} by $9.9\%$ and $3.8\%$ in terms of $\delta_1$ and $OPW$ on NYUDV2~\cite{nyu}. We also achieve better performance than DeepV2D~\cite{deepv2d}, which is the previous \sota{} structure-from-motion-based method~\cite{deepv2d,sfm21,lighted,colmapsfm} but can only deal with completely static scenes. The results demonstrate that using our architecture alone can also obtain better performance.

\begin{table}
\caption{
\textbf{Comparisons of learning-based approaches on NYUDV2~\cite{nyu} and KITTI~\cite{kitti}}.
All the compared methods use NYUDV2~\cite{nyu} or KITTI~\cite{kitti} as the training and evaluation data, following the train/test split as Eigen~\textit{et al.}~\cite{silog}. Our \sx{} trained from scratch also achieves better performance than all the other methods.}
\vspace{-7pt}
\label{tab:nyu}
\begin{center}
\resizebox{\columnwidth}{!}{
\begin{tabular}{lcccccc}
\toprule
Method & $\delta_1\uparrow$ & $\delta_2\uparrow$ & $\delta_3\uparrow$ & $Rel\downarrow$ &$OPW\downarrow$ \\
\midrule
\multicolumn{6}{c}{NYUDV2} \\
\midrule
SC-DepthV1~\cite{scd1} & $0.813$ & $0.952$ & $0.987$ & $0.143$ &$0.465$ \\
SC-DepthV2~\cite{scd2} & $0.820$ & $0.956$ & $0.989$ & $0.138$ &$0.474$ \\
SC-DepthV3~\cite{scd3} & $0.848$ & $0.963$ & $0.991$ & $0.123$ &$0.441$ \\
ST-CLSTM~\cite{ST-CLSTM} & $0.833$ & $0.965$ & $0.991$ & $0.131$ &$0.645$ \\
Cao \textit{et al.}~\cite{MM21} & $0.835$ & $0.965$ & $0.990$ & $0.131$ &$-$ \\
FMNet~\cite{fmnet} & $0.832$ & $0.968$ & $0.992$ & $0.134$ &$0.387$ \\
DeepV2D~\cite{deepv2d} & $0.924$ & $0.982$ & $0.994$ & $0.082$ &$0.402$ \\
Ours-Large-scratch(DPT-Large) & $\textbf{0.931}$ & $\textbf{0.989}$ & $\textbf{0.996}$ & $\textbf{0.081}$ &$\textbf{0.345}$ \\
\midrule
\multicolumn{6}{c}{KITTI} \\
\midrule
SC-DepthV1~\cite{scd1} & $0.860$ & $0.956$ & $0.981$ & $0.118$ &$0.402$ \\
SC-DepthV2~\cite{scd2} & $0.866$ & $0.958$ & $0.981$ & $0.118$ &$0.389$ \\
SC-DepthV3~\cite{scd3} & $0.864$ & $0.960$ & $0.984$ & $0.118$ &$0.397$ \\
ST-CLSTM~\cite{ST-CLSTM} & $0.890$ & $0.970$ & $0.989$ & $0.101$ &$0.413$ \\ 
Cao \textit{et al.}~\cite{MM21} & $0.872$ & $0.962$ & $0.986$ & $0.109$ &$-$ \\
FMNet~\cite{fmnet} & $0.886$ & $0.968$ & $0.989$ & $0.099$ &$0.375$ \\
TC-Depth~\cite{tcdepth} & $0.921$ & $-$ & $0.997$ & $0.082$ & $-$ \\
DeepV2D~\cite{deepv2d} & $0.972$ & $0.991$ & $0.996$ & $0.051$ &$0.428$ \\
Ours-Large-scratch(DPT-Large) & $\textbf{0.978}$ & $\textbf{0.998}$ & $\textbf{0.999}$ & $\textbf{0.046}$ &$\textbf{0.239}$ \\
\bottomrule
\end{tabular}
}
\end{center}
\vspace{-15pt}
\end{table}

\noindent \textbf{Qualitative Comparisons.}
We show some qualitative comparisons on natural-scene videos in \reffig{}~\ref{fig:qpqp}. We draw the scanline slice over time. Fewer zigzagging pattern means better consistency. The initial estimation of DPT~\cite{dpt} in the seventh row contains flickers and blurs, which are eliminated with the proposed \sx{}, as shown in the last row. Although the test-time-training-based Robust-CVD~\cite{rcvd} shows competitive performances on the indoor NYUDV2~\cite{nyu} dataset, it is not robust on the natural scenes. As can be observed in the fourth row, Robust-CVD produces obvious artifacts due to the erroneous pose estimation. 

\begin{figure*}
\begin{center}
   \includegraphics[width=0.97\textwidth,trim=0 0 0 0,clip]{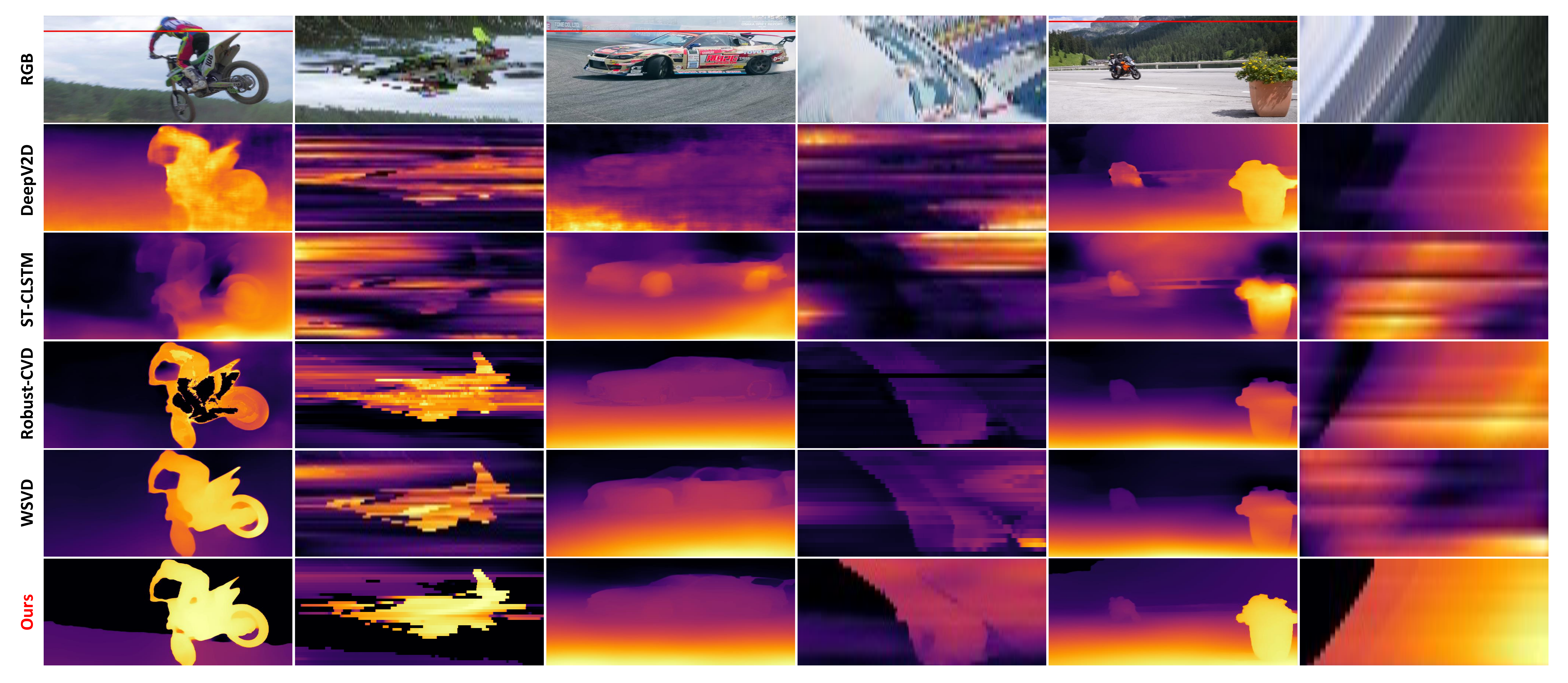}
\end{center}
\vspace{-12pt}
   \caption{\textbf{Qualitative results on the DAVIS~\cite{davis} dataset.} Our method can predict robust and consistent results on these challenging videos with large camera and object motions such as the motorcycle stunt flying over the hill and the drifting racing car, while other compared methods produce poor results or even failure cases on these difficult scenes.}
\label{fig:ksh-davis}
\vspace{-5pt}
\end{figure*}

Besides, we also showcase some visual comparisons on the in-the-wild videos from the DAVIS~\cite{davis} dataset. As shown in \reffig{}~\ref{fig:ksh-davis}, we present some challenging videos
with large camera and object motions (\textit{e.g.}, the motorcycle stunt flying over the hill and the drifting racing car). \sx{} can predict both robust and consistent disparity results, while other compared methods produce obvious artifacts and failure cases on these difficult scenes.

In both \reffig{}~\ref{fig:qpqp} and~\ref{fig:ksh-davis}, one can observe that we produce much sharper estimation at edges, especially on skylines, which can be down to our rigorous annotation pipeline for \data{} dataset, \textit{e.g.}, the ensemble strategy for sky segmentation. Please refer to our supplementary document and demo video for more visual comparisons.

\begin{table}
\caption{
    \textbf{Influence of different training data}. (a) Training with different datasets. We conduct zero-shot evaluations on Sintel~\cite{sintel} with different training data for our \sx{} model. (b) Pretrain and finetune. Pretraining on our \data{} can further improve the results on the closed-domain NYUDV2~\cite{nyu}, compared with training from scratch, even with weaker single-image depth predictors MiDaS-v2.1-Large~\cite{MiDaS} than DPT-Large~\cite{dpt}.}
    \label{tab:sjjyx}
    
    \centering
    \resizebox{0.45\columnwidth}{!}{
    \begin{subtable}[t]{0.45\linewidth}
        \addtolength{\tabcolsep}{-4pt}
        \begin{tabular}{lcc}
            \toprule
            Dataset & $\delta_1\uparrow$ & $OPW\downarrow$ \\
            \midrule
            NYUDV2 & $0.527$ &$0.435$ \\
            IRS+TartanAir & $0.542$ &$0.489$ \\
            \data{}(Ours) & $\textbf{0.591}$ & $\textbf{0.424}$ \\
            \bottomrule
        \end{tabular}
        \vspace{+6.2pt}\caption{Different Training Data}
    \end{subtable}
    }
    \resizebox{0.45\columnwidth}{!}{
    \begin{subtable}[t]{0.45\linewidth}
         \addtolength{\tabcolsep}{-4pt}
        \begin{tabular}{lcc}
       
            \toprule
            Setting & $\delta_1\uparrow$ & $OPW\downarrow$ \\
            \midrule
          Scratch(DPT) & $0.931$ &$0.345$ \\
          Pretrain(MiDaS) & $0.941$ & $0.347$ \\
          Pretrain(DPT) & $\textbf{0.950}$ & $\textbf{0.339}$ \\
            \bottomrule
        \end{tabular}
         \vspace{+1pt}\caption{Pretrain and Finetune}
    \end{subtable}
    }
\vspace{-15pt}
\end{table}

\noindent \textbf{Influence of Training Data.}
The quality and diversity of data can greatly influence the learning-based video depth models. Our \data{} dataset offers hundreds of times more data and scenes compared to previous works, which can be used to train robust learning-based models in the wild. To better show the difference, we compare our dataset with the existing datasets under zero-shot cross-dataset setting. As shown in \reftab{}~\ref{tab:sjjyx} (a), we train our \sx{} with existing video depth datasets~\cite{nyu,irs,tata} and evaluate the model on Sintel~\cite{sintel} dataset. With both quantity and diversity, using \data{} as the training data yields the best accuracy and consistency. Our VDW
dataset is far more diverse for training robust video depth models, compared with large closed-domain dataset NYUDV2~\cite{nyu} or synthetic natural-scene dataset like IRS~\cite{irs} and TartanAir~\cite{tata}.

Moreover, although the proposed \data{} is designed for natural scenes, it can also boost the performance on closed domains by serving as pretraining data. As in \reftab{}~\ref{tab:sjjyx} (b), the \data{}-pretrained model outperforms the model that is trained from scratch, even with weaker single-image model (MiDaS-v2.1-Large~\cite{MiDaS}). These results suggest that \data{} can also benefit some closed-domain scenarios. This conclusion is also proved on the KITTI dataset~\cite{kitti} by comparing the quantitative results in the last row of \reftab{}~\ref{tab:bigone} (Ours-Large) and \reftab{}~\ref{tab:nyu} (Ours-Large-scratch).

\begin{table}
\caption{\textbf{Efficacy of the bidirectional inference strategy with \flow{}.} The simple bidirectional averaging can enlarge temporal receptive fields and improve the consistency beyond the forward or backward outcomes produced by the previous or post sliding window. Besides, temporal consistency can be further enhanced by our adaptive flow-guided fusion. We report the results on the VDW test set with DPT-Large~\cite{dpt} and MiDaS-v2.1-Large~\cite{MiDaS} as different depth predictors.} 
    \label{tab:sxsx}
    \centering
    \resizebox{0.45\columnwidth}{!}{
    \begin{subtable}[t]{0.45\linewidth}
    \addtolength{\tabcolsep}{-4pt}
        \begin{tabular}{lcc}
            \toprule
            Method & $\delta_1\uparrow$ & $OPW\downarrow$ \\
            \midrule
            DPT-Large~\cite{dpt} & $0.730$ &$0.470$ \\
            Pre-window & $0.741$ &$0.165$ \\
            Post-window & $0.741$ &$0.174$ \\
            Averaging & $\underline{0.742}$ &$\underline{0.147}$ \\
            \midrule
            Flow-Guided & $\textbf{0.742}$ &$\textbf{0.129}$ \\
            \bottomrule
        \end{tabular}
        \vspace{+6pt}\caption{DPT Initialization}
    \end{subtable}
    }
    \resizebox{0.45\columnwidth}{!}{
    \begin{subtable}[t]{0.45\linewidth}
    \addtolength{\tabcolsep}{-4pt}
        \begin{tabular}{lcc}
            \toprule
            Method & $\delta_1\uparrow$ & $OPW\downarrow$ \\
            \midrule
            MiDaS-Large~\cite{MiDaS} & $0.651$ &$0.676$ \\
            Pre-window & $0.700$ &$0.207$ \\
            Post-window & $0.699$ &$0.218$ \\
            Averaging & $\underline{0.700}$ &$\underline{0.180}$ \\
            \midrule
            Flow-Guided & $\textbf{0.701}$ &$\textbf{0.148}$ \\
            \bottomrule
        \end{tabular}
        \vspace{+1pt}\caption{MiDaS Initialization}
    \end{subtable}
    }
\vspace{-10pt}
\end{table}

\subsection{\FLOW{}}
\label{sec:flowexpexp}
Here, we conduct experiments to expound on the effectiveness of our bidirectional inference strategy with \flow{}. 
As shown in the first four rows of \reftab{}~\ref{tab:sxsx}, whether using DPT~\cite{dpt} or MiDaS~\cite{MiDaS} as the single-image depth predictor, our \sx{} can already enforce the temporal consistency only with the previous or post sliding window of the target frame. Compared with the single-direction results, the bidirectional inference with simple averaging can improve the consistency by over $10.9\%$ with larger temporal receptive fields.


Besides, we propose the \flow{} paradigm to adaptively fuse the bidirectional disparity results of reference and target frames. With the flow-guided relevance maps and the adaptive fusion, the temporal consistency can be further enhanced. As shown in the last two rows of \reftab{}~\ref{tab:sxsx}, compared with the simple averaging, our \flow{} can further improve the temporal consistency by $17.7\%$.

\subsection{Efficiency Comparisons and Lightweight Model}
\label{sec:spd}

\begin{table}[!t]
\caption{\textbf{Comparisons of FLOPs and model parameters.} We evaluate the efficiency of our \ssmall{}, \llarge{}, and different depth predictors including DPT-Large~\cite{dpt}, NeWCRFs~\cite{newcrfs}, and MiDaS-v2.1-Large~\cite{MiDaS}. The FLOPs are evaluated on a $384 \times 384$ video sequence with four frames. 
The model parameters and FLOPs for \ssmall{} and \llarge{} throughout the manuscript are in addition to the single-image depth predictors.
Our lightweight \ssmall{} model has $7$ times fewer parameters and $17$ times fewer FLOPs than \llarge{}.}
\vspace{-10pt}
\label{tab:flopszw}
\addtolength{\tabcolsep}{-4pt}
\begin{center}
\resizebox{\columnwidth}{!}{
\begin{tabular}{lcccccc}
\toprule
  & DPT~\cite{dpt} & NeWCRFs~\cite{newcrfs}&MiDaS~\cite{MiDaS} & Ours-Large & Ours-Small \\
\midrule
FLOPs ($G$) & $1011.32$ & $550.47$& $415.24$& $254.53$ & $\textbf{35.51}$\\
Params ($M$) & $341.26$ & $270.33$ &$104.18$ & $88.31$ & $\textbf{5.04}$ \\
\bottomrule
\end{tabular}
}
\end{center}
\vspace{-10pt}
\end{table}

\noindent\textbf{Evaluations of Efficiency}. For \llarge{}, we compare the inference time on a $640\times480$ video with eight frames. The inference is conducted on one NVIDIA RTX A6000 GPU. As shown in \reftab{}~\ref{tab:bigone}, the proposed \llarge{} reduces the inference time by hundreds of times compared to the test-time-training-based CVD~\cite{CVD}, Robust-CVD~\cite{rcvd}, and Zhang \textit{et al.}~\cite{dycvd}. The learning-based method DeepV2D~\cite{deepv2d} alternately estimates depth and camera poses, which is time-consuming. WSVD~\cite{wsvd} is also slower than \llarge{}. 

We also compare the computational costs of the proposed \ssmall{}, \llarge{}, and different depth predictors~\cite{MiDaS,dpt,newcrfs}. Model parameters and FLOPs are reported in \reftab{}~\ref{tab:flopszw}. The FLOPs are evaluated on a $384 \times 384$ video sequence with four frames. Our \llarge{} only incurs limited computation overhead compared with the depth predictors~\cite{MiDaS,dpt,newcrfs}. Besides, compared with \llarge{} model, our \ssmall{} has $7$ times fewer parameters and $17$ times fewer FLOPs. As indicated in \reftab{}~\ref{tab:smallpf}, our \ssmall{} can achieve inference speeds of over $30$ fps, surpassing the previous real-time video depth model ST-CLSTM~\cite{ST-CLSTM}.

\begin{figure}[!t]
\begin{center}
   \includegraphics[width=0.48\textwidth,trim=10 30 10 30,clip]{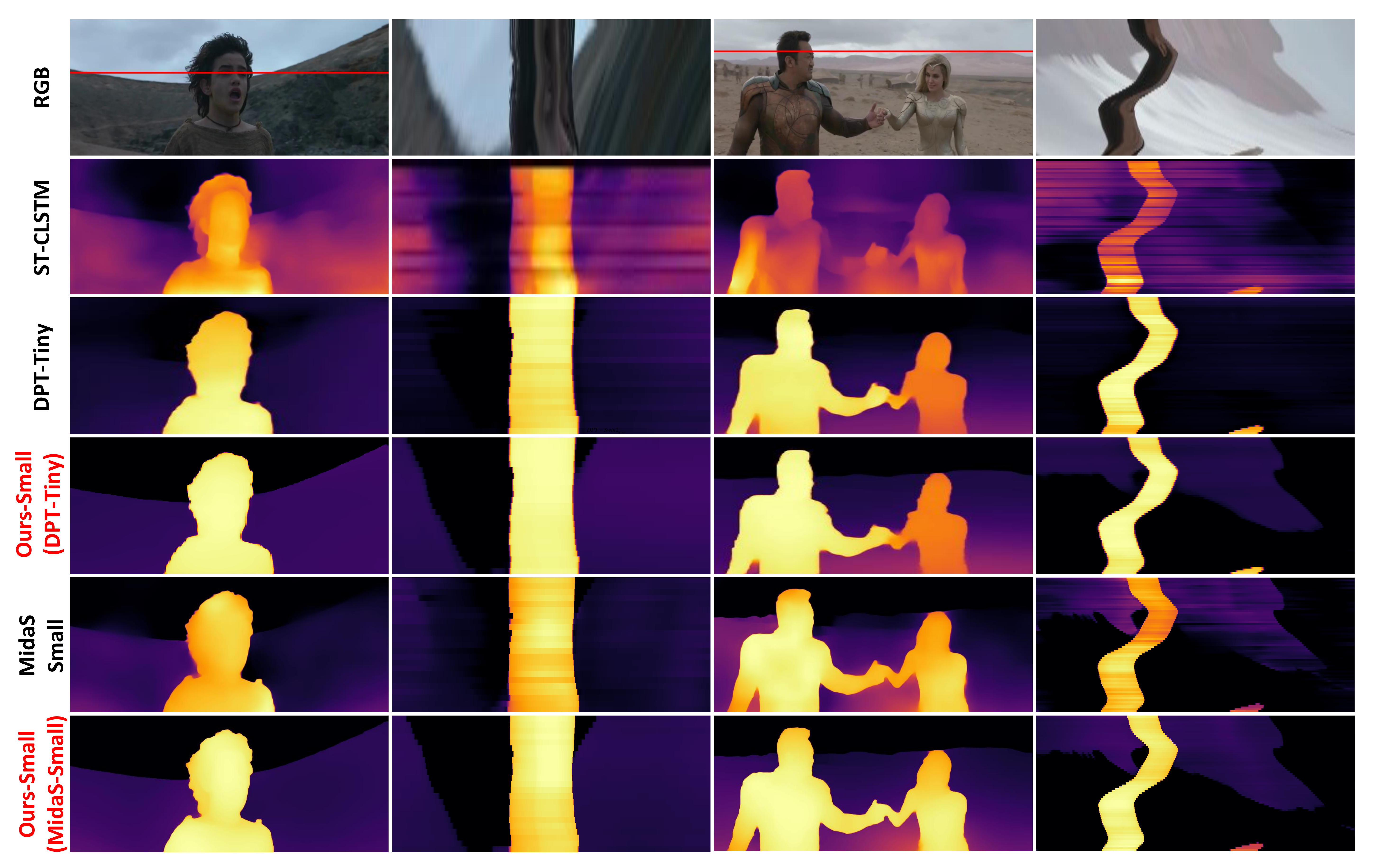}
\end{center}
\vspace{-3pt}
\caption{\textbf{Visual results of \ssmall{} model.} We compare \ssmall{} with lightweight depth predictors DPT-Swin2-Tiny~\cite{MiDaSV31} and MiDaS-v2.1-Small~\cite{MiDaS}, along with the previous real-time video depth model ST-CLSTM~\cite{ST-CLSTM}. Best view zoomed in on-screen for details.}
\label{fig:smallv1}
\end{figure}

\begin{table}
\addtolength{\tabcolsep}{-5pt}
\begin{center}
\caption{
\textbf{Quantitative Comparisons of lightweight single-image depth predictors and video depth models on VDW.} We also report model parameters and the average FPS of processing all the testing videos with the input resolution of $896\times384$. The FPS numbers of our model include the initial depth predictors. Our \ssmall{} shows both strong performance and high efficiency with different depth predictors.}
\label{tab:smallpf}
\resizebox{\columnwidth}{!}{
\begin{tabular}{lccccccccc}
\toprule
Method & $\delta_1\uparrow$ & $\delta_2\uparrow$ & $\delta_3\uparrow$ & $Rel\downarrow$ &$OPW\downarrow$ & $FPS\uparrow$ & Params ($M$)$\downarrow$ \\
\midrule
MiDaS-v2.1-Small~\cite{MiDaS} & $0.564$ & $0.794$ & $0.891$ & $0.401$ &$0.652$ & $\textbf{45.06}$ & $21.48$\\
DPT-Swin2-Tiny~\cite{MiDaSV31} & $0.672$ & $0.862$ & $0.931$ & $0.264$ &$0.619$ & $41.55$ & $42.73$\\
\midrule
ST-CLSTM~\cite{ST-CLSTM} & $0.477$ & $0.732$ & $0.857$ & $0.521$ & $0.448$ & $26.82$ & $15.39$ \\
Ours-Small(MiDaS-v2.1-Small) & $0.622$ & $0.832$ & $0.913$ & $0.347$ & $\textbf{0.236}$ & $35.17$ & $\textbf{5.04}$\\
Ours-Small(DPT-Swin2-Tiny) & $\textbf{0.704}$ & $\textbf{0.881}$ & $\textbf{0.944}$ & $\textbf{0.251}$ &$0.259$ & $32.75$ & $\textbf{5.04}$\\

\bottomrule
\end{tabular}
}
\end{center}
\end{table}

\noindent \textbf{Quantitative and Visual Results of \ssmall{} model.} We assess and compare the spatial accuracy and temporal consistency of lightweight single-image depth predictors and video depth models in \reftab{}~\ref{tab:smallpf}. Collaborating with different lightweight depth predictors MiDaS-v2.1-Small~\cite{MiDaS} and DPT-Swin2-Tiny~\cite{MiDaSV31}, our \ssmall{} can improve the $OPW$ by over $58.1\%$ and $\delta_1$ by over $3.2\%$. Meanwhile, \ssmall{} maintains real-time processing of $35.17$ fps and compact model structure only with $5.04 M$ parameters. 

Visual comparisons are shown in \reffig{}~\ref{fig:smallv1}. \ssmall{} achieves significant improvements over lightweight depth predictors~\cite{MiDaS,MiDaSV31} and previous real-time ST-CLSTM~\cite{ST-CLSTM} in both the spatial accuracy and temporal consistency. 


\subsection{Extension to Video Semantic Segmentation}
\label{sec:yyfgsy}

In this section, we delve into the dataset, evaluation metrics, and experimental results of video semantic segmentation.

\subsubsection{Dataset and Evaluation Metrics}

\noindent \textbf{\city{} Dataset.} We follow previous single-image segmenters~\cite{segformer,oneformer} and video semantic segmentation approaches~\cite{etc,ifr,tmanet,mrcfa,cffm,SSLTM} to conduct experiments and evaluations on \city{}~\cite{cityscapes} dataset. \city{}~\cite{cityscapes} is a widely-used standard benchmark for semantic segmentation. It contains videos with 30 frames and 17 fps of urban street scenes. The $20^{th}$ frame of each video has high-quality annotations. A total of $5@000$ frames are finely annotated. In the training stage, we use the official training set with $2@975$ annotated frames as target frames.

\begin{table}
\caption{
\textbf{Comparisons of video semantic segmentation approaches on \city{}~\cite{cityscapes} validation set.} Following SSLTM~\cite{SSLTM}, all methods adopt ResNet-50~\cite{resnet} as backbone.}
\vspace{-7pt}
\label{tab:segcompare}
\addtolength{\tabcolsep}{-1pt}
\begin{center}
\resizebox{1.0\columnwidth}{!}{
\begin{tabular}{lccc}
\toprule
Method & Params ($M$)$\downarrow$ & mIoU$\uparrow$ & TC$\uparrow$ \\
\midrule
ETC~\cite{etc} (ECCV'20) & $39.1$ & $77.91$ & $71.29$ \\
TMANet~\cite{tmanet} (ICIP'21) & $32.1$ & $78.50$ & $-$ \\
CFFM~\cite{cffm} (CVPR'22) & $39.6$ & $78.14$ & $72.53$ \\
MRCFA~\cite{mrcfa} (ECCV'22) & $39.2$ & $78.12$ & $72.21$ \\
IFR~\cite{ifr} (CVPR'22) & $46.7$ & $78.42$ & $71.94$ \\
SSLTM~\cite{SSLTM} (CVPR'23) & $43.1$ & $79.69$ & $-$ \\
\midrule
Ours & $42.7$ & $\textbf{80.84}$ & $\textbf{73.07}$ \\
\bottomrule
\end{tabular}
}
\end{center}
\vspace{-5pt}
\end{table}

\begin{table}
    \caption{
\textbf{Different semantic segmenters on \city{}~\cite{cityscapes} validation set.} The results demonstrate the efficacy of our plug-and-play manner in video semantic segmentation.}
\vspace{-7pt}
\label{tab:blabseg}
    \renewcommand\arraystretch{0.5}
    \begin{center}
    \resizebox{\columnwidth}{!}{
    \begin{tabular}{lcccccc}
    \toprule
    \multirow{2}{*}{} & 
    \multicolumn{2}{c}{Initial} &
    \multicolumn{3}{c}{$\;\;\;\;\;$Ours} \\
    \cmidrule{2-3} \cmidrule{5-6}
        & mIoU$\uparrow$ & TC$\uparrow$ & & 
        mIoU$\uparrow$ & TC$\uparrow$ \\
    \midrule
    SegFormer-B1~\cite{segformer}    &$77.87$& $70.32$ & &$79.91$ & $72.94$\\
    SegFormer-B3~\cite{segformer}    &$80.45$& $71.24$ & &$80.84$ & $73.07$\\
    OneFormer~\cite{oneformer}    &$\textbf{83.02}$& $\textbf{71.38}$ & &$\textbf{83.09}$ & $\textbf{73.12}$\\
    
    \bottomrule
    \end{tabular}
    }
\end{center}
\end{table}

\noindent \textbf{Evaluation Metrics.} To evaluate the accuracy of semantic segmentation on \city{}~\cite{cityscapes} dataset, we adopt the mean Intersection over Union (mIoU) following prior arts~\cite{etc,ifr,tmanet,mrcfa,cffm,SSLTM}.

To evaluate the consistency of video results, previous methods~\cite{etc,mrcfa,cffm,SSLTM} mainly adopt the mean Video Consistency (mVC)~\cite{vspw} or the temporal consistency (TC)~\cite{etc} metrics. However, the calculation of mVC requires annotations of all video frames, while \city{}~\cite{cityscapes} only annotates one frame per video in both training and validation sets. Thus, we utilize the TC metric~\cite{etc} for comparisons, which does not rely on segmentation ground truth. We denote the segmentation result of the $n^{th}$ frame as $\mathcal{Q}_{n}$. $\hat{\mathcal{Q}}_{n-1}$ represents the warped segmentation map from frame $n-1$ to frame $n$ by optical flow~\cite{gmflow}. TC~\cite{etc} calculates mIoU between $\mathcal{Q}_{n}$ and $\hat{\mathcal{Q}}_{n-1}$ to measure the consistency:

\begin{equation}
    TC = \frac{1}{N-1}\sum_{n=2}^{N}\frac{\mathcal{Q}_{n}\bigcap \hat{\mathcal{Q}}_{n-1}}{\mathcal{Q}_{n}\bigcup \hat{\mathcal{Q}}_{n-1}}\,.
\end{equation}
Therefore, contrary to $OPW$ for video depth, larger TC represents better consistency in the semantic segmentation maps. We report the average TC of all testing videos.

\begin{figure*}[!t]
\begin{center}
   \includegraphics[width=0.92\textwidth,trim=0 0 0 0,clip]{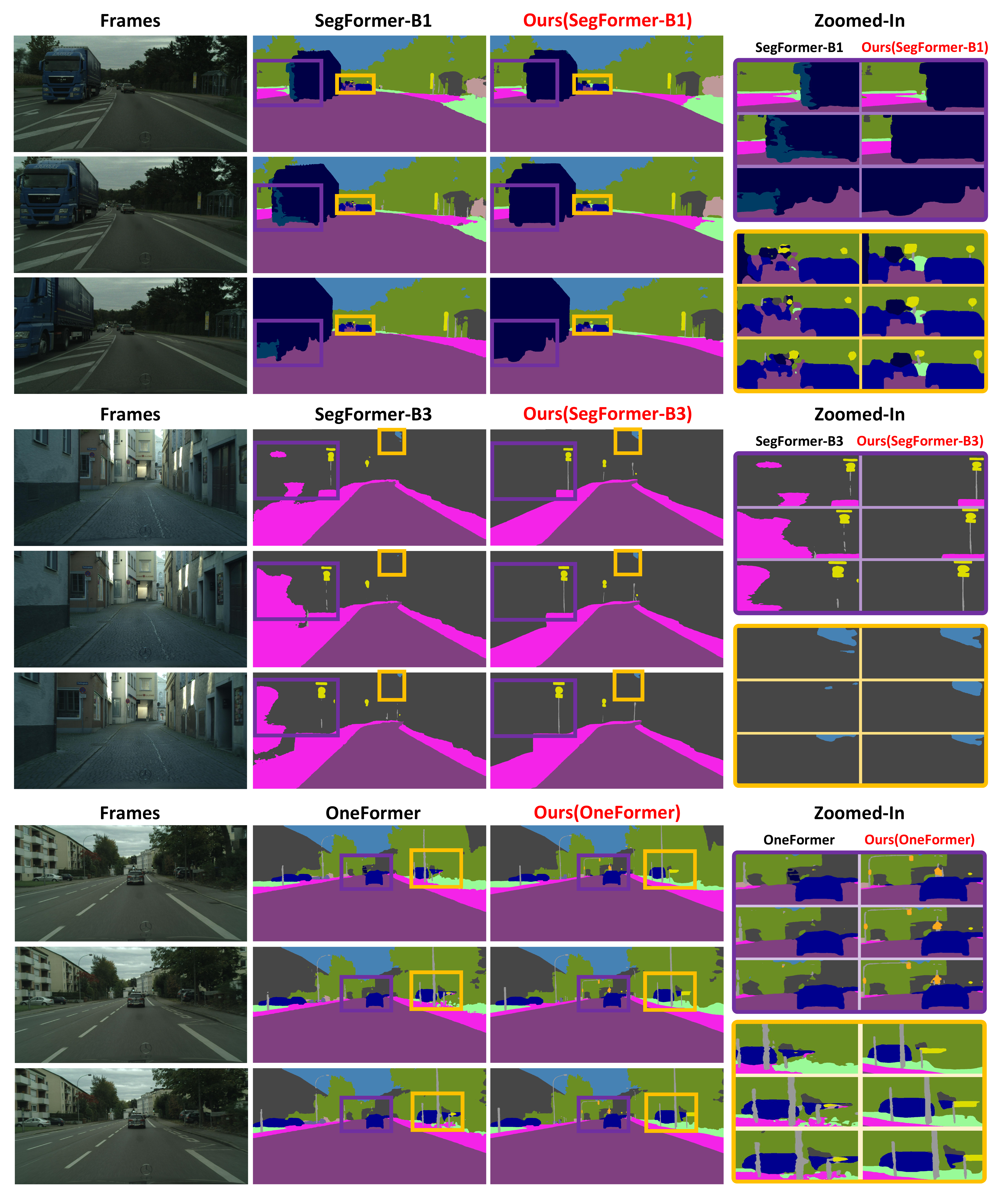}
\end{center}
   \vspace{-10pt}
   \caption{\textbf{Visual results of video semantic segmentation on \city{}~\cite{cityscapes} dataset.} We compare \sx{} with three different semantic segmenters SegFormer-B1, SegFormer-B3~\cite{segformer}, and OneFormer~\cite{oneformer} to demonstrate the efficacy of our plug-and-play paradigm. Our \sx{} can stabilize the temporal flickers in the initial segmentation maps. We highlight and zoom-in the regions with obvious difference in rectangular boxes. Best view zoomed in on-screen for details.}
\label{fig:segksh}
\vspace{-5pt}
\end{figure*}

\subsubsection{Comparisons with Prior Arts}

Firstly, we compare our \sx{} with the previous \sota{} video semantic segmentation approaches. The experimental results are shown in \reftab{}~\ref{tab:segcompare}. We follow SSLTM~\cite{SSLTM} to use ResNet-50~\cite{resnet} as the backbone for fair comparisons. With similar model parameters to prior arts~\cite{etc,tmanet,SSLTM,cffm,mrcfa,ifr}, our approach achieves \sota{} performance on both the segmentation accuracy and video consistency (mIoU and TC) for video semantic segmentation. The results prove the applicability and generality of our \sx{} framework in video dense prediction.

\begin{figure*}[!t]
\begin{center}
\includegraphics[width=0.92\textwidth,trim=0 0 0 0,clip]{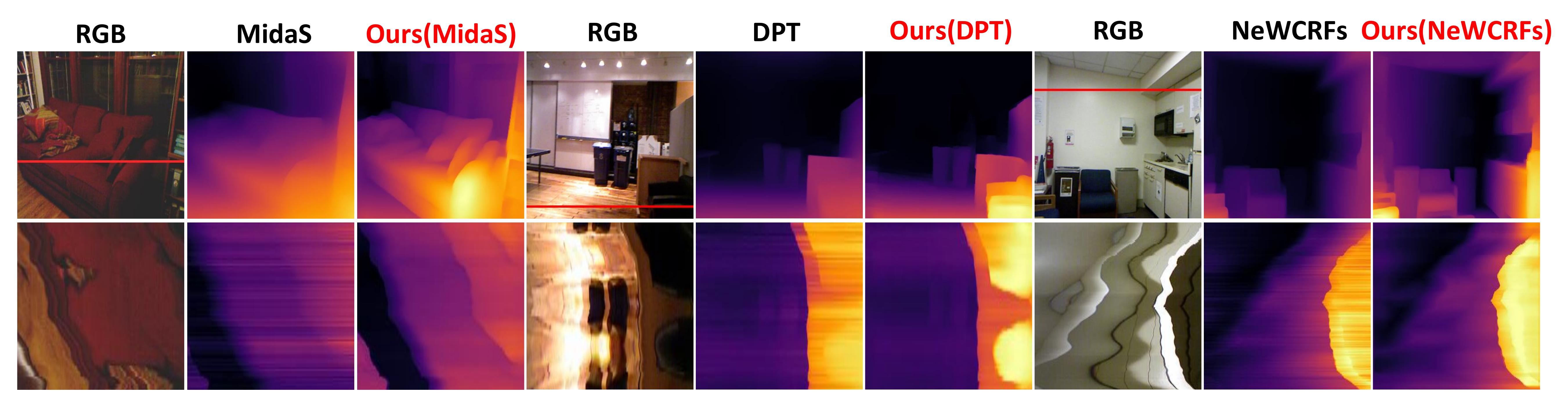}
\end{center}
\vspace{-10pt}
   \caption{\textbf{Visual results on the NYUDV2~\cite{nyu} dataset.} We compare \llarge{} with three different depth predictors, including the MiDaS-v2.1-Large~\cite{MiDaS}, DPT-Large~\cite{dpt}, and NeWCRFs\cite{newcrfs}.}
\label{fig:nyudp}
\vspace{-10pt}
\end{figure*}

Besides, we conduct evaluations to prove the effectiveness of our plug-and-play framework in video semantic segmentation. The quantitative results are presented in \reftab{}~\ref{tab:blabseg}. We utilize three different initial semantic segmenters including SegFormer-B1, SegFormer-B3~\cite{segformer}, and OneFormer~\cite{oneformer}. Our \sx{} can enforce the temporal consistency over those different segmenters. Visual results are shown in \reffig{}~\ref{fig:segksh}. Regions with obvious improvements are highlighted and zoomed in by rectangular boxes. Our \sx{} framework and pluggable paradigm showcase strong effectiveness in both video depth estimation and semantic segmentation tasks, proving the versatility and generality of our method.

\begin{table}
\caption{
\textbf{Comparisons of different depth predictors on the NYUDV2~\cite{nyu} dataset.} Our \llarge{} is compatible with different depth predictors in a plug-and-play manner.}
\vspace{-7pt}
\label{tab:blab}
    \begin{center}
    \resizebox{\columnwidth}{!}{
    \begin{tabular}{lccccccc}
    \toprule
    \multirow{2}{*}{} & 
    \multicolumn{3}{c}{Initial} &
    \multicolumn{4}{c}{Ours} \\
    \cmidrule{2-4} \cmidrule{6-8}
        & $\delta_1\uparrow$ & $Rel\downarrow$ &
        $OPW\downarrow$ & & 
        $\delta_1\uparrow$ & $Rel\downarrow$ & $OPW\downarrow$ \\
    \midrule
    MiDaS-v2.1-Large~\cite{MiDaS}    &$0.910$& $0.095$ &$0.862$ &&$0.941$& $0.076$ & $0.373$ \\
    DPT-Large~\cite{dpt}        &$0.928$&$0.084$ &$0.811$ &&$0.950$&$0.072$ & $0.364$\\
    NeWCRFs~\cite{newcrfs} &$\textbf{0.937}$& $\textbf{0.072}$&$\textbf{0.645}$ &&$\textbf{0.957}$& $\textbf{0.068}$& $\textbf{0.326}$\\
    \bottomrule
    \end{tabular}
    }
\end{center}
\vspace{-7pt}
\end{table}

\begin{table}[!t]
\caption{\textbf{Temporal loss and inter-frame intervals} $\textbf{l}$. We randomly split $100$ videos for training and $10$ videos for testing from our \data{} dataset in these two experiments.}
    \label{tab:clipop}
    \centering
    \resizebox{0.47\columnwidth}{!}{
    \begin{subtable}[t]{0.45\linewidth}
    \addtolength{\tabcolsep}{-4pt}
        \begin{tabular}{lcc}
            \toprule
            Method & $\delta_1\uparrow$ & $OPW\downarrow$ \\
            \midrule
            DPT-Large~\cite{dpt} & $0.621$ &$0.492$ \\
            $w/o\;\mathcal{L}^t$ & $\textbf{0.627}$ &$0.303$ \\
            $w/\;\mathcal{L}^t$ & $0.625$ &$\textbf{0.216}$ \\
            \bottomrule
        \end{tabular}
        \vspace{+6pt}\caption{Temporal Loss}
    \end{subtable}
    }
    \resizebox{0.47\columnwidth}{!}{
    \begin{subtable}[t]{0.45\linewidth}
        \begin{tabular}{lcc}
            \toprule
            Method & $\delta_1\uparrow$ & $OPW\downarrow$ \\
            \midrule
           $l=1$ & $\textbf{0.625}$ &$\textbf{0.216}$ \\
           $l=3$ & $0.618$ &$0.219$ \\
            $l=5$ & $0.621$ &$0.246$ \\
            \bottomrule
        \end{tabular}
        \vspace{+1.5pt}\caption{Inter-frame Intervals}
    \end{subtable}
    }
\vspace{-10pt}   
\end{table}

\subsection{Ablation Studies}
\label{sec:abl}
Here we verify the effectiveness of the proposed method by ablation studies. We first ablate our plug-and-play paradigm with different depth predictors or semantic segmenters. Besides, We also discuss 
the temporal loss, the reference frames, and the baselines without the stabilization network. 

\noindent \textbf{Plug-and-play Manner.} As shown in \reftab{}~\ref{tab:blab}, we directly adapt our \llarge{} model to three different single-image depth models DPT-Large~\cite{dpt}, MiDaS-v2.1-Large~\cite{MiDaS}, and NeWCRFs~\cite{newcrfs}. For NeWCRFs~\cite{newcrfs}, we adopt their official checkpoint on NYUDV2~\cite{nyu}. By post-processing their initial flickering disparity maps, our \sx{} achieves better temporal consistency and spatial accuracy. With higher initial depth accuracy, the spatial performance of our \sx{} is also improved. The experiment demonstrates the effectiveness of our plug-and-play manner. Visual comparisons with those three depth predictors are shown in \reffig{}~\ref{fig:nyudp}. Depth maps and scanline slice prove our accuracy and consistency.

On the other hand, in pursuit of real-time processing, our \ssmall{} can cooperate with lightweight depth predictors in the pluggable paradigm, \textit{e.g.}, MiDaS-v2.1-Small~\cite{MiDaS} and DPT-Swin2-Tiny~\cite{MiDaSV31}. As proved in \reftab{}~\ref{tab:smallpf}, only with $5.04 M$ model parameters, our \ssmall{} can improve the consistency, accuracy, and achieve processing speeds of $35.17$ fps. Qualitative comparisons with those
lightweight depth predictors can be found in \reffig{}~\ref{fig:smallv1}.

The plug-and-play manner can be effectively extended to video semantic segmentation. As shown in \reftab{}~\ref{tab:blabseg} and \reffig{}~\ref{fig:segksh}, with SegFormer-B1, SegFormer-B3~\cite{segformer}, and OneFormer~\cite{oneformer} as segmenters, the temporal consistency and segmentation accuracy are both improved by \sx{}. With higher initial segmentation accuracy from OneFormer~\cite{oneformer}, the \sx{} can also achieve better performance.


\begin{table}[!t]
\caption{\textbf{Baselines without \sx{} stabilization network and reference frames numbers} $\textbf{n}$. The experiment is conducted on the same VDW subset as \reftab{}~\ref{tab:clipop}.}
    \label{tab:nostab}
    \centering
    
    \hspace{+2pt}
    \resizebox{0.46\columnwidth}{!}{
    \begin{subtable}[t]{0.5\linewidth}
        \centering
        \addtolength{\tabcolsep}{-5.3pt}
        \begin{tabular}{lccc}
            \toprule
           DPT w/ & Single-frame & Multi-frame & \textbf{Ours} \\
           \midrule
           $\delta_1\uparrow$ & $0.615$ & $0.608$ & $\textbf{0.625}$ \\
           $OPW\downarrow$ &$0.488$ &$0.471$ &$\textbf{0.216}$\\
           \bottomrule
        \end{tabular}
        \vspace{+1pt}\caption{$\;$W/o Stabilization Network}
    \end{subtable}
    }
    \hspace{+12pt}
    \resizebox{0.46\columnwidth}{!}{
    \begin{subtable}[t]{0.5\linewidth}
         \addtolength{\tabcolsep}{-4pt}
         \begin{tabular}{lcccc}
            \toprule
            & n=1 & n=2 & \textbf{n=3} & n=4 \\
            \midrule
            $\delta_1 \uparrow$ & $0.618$ &$0.622$ & $\textbf{0.625}$ & $\textbf{0.625}$\\
            $OPW\downarrow$ & $0.272$ & $0.233$ & $\textbf{0.216}$ & $0.224$ \\
           \bottomrule
           
        \end{tabular}
        \vspace{+1pt}\caption{Reference Frames}
        
    \end{subtable}
    }
    
\vspace{-7pt} 
\end{table}

\begin{table}[t]
    
    \centering
    \caption{\textbf{Forward training and backward inference.} The domain gap is modest and our model can handle it with minor impacts on the performance. Compared with training solely in the forward direction, bidirectional training only yields subtle improvements on the VDW dataset.}
    \hspace{-20pt}
    \resizebox{0.47\columnwidth}{!}{
    \begin{subtable}[t]{0.45\linewidth}
    \addtolength{\tabcolsep}{-4pt}
        \begin{tabular}{lcc}
            \toprule
            Inference & $\delta_1\uparrow$ & $OPW\downarrow$ \\
            \midrule
            Forward & $0.741$ &$0.165$ \\
            Backward & $0.741$ &$0.174$ \\
            Bidirectional & $\textbf{0.742}$ &$\textbf{0.129}$ \\
            \bottomrule
        \end{tabular}
        \vspace{+7pt}
        \hspace{-15pt}\caption{Forward Training\;\;\;}
    \end{subtable}
    }
    \hspace{-12pt}
    \resizebox{0.47\columnwidth}{!}{
    \begin{subtable}[t]{0.45\linewidth}
        \begin{tabular}{lcc}
            \toprule
            Inference & $\delta_1\uparrow$ & $OPW\downarrow$ \\
            \midrule
           Forward & $0.741$ &$0.164$ \\
           Backward & $\textbf{0.742}$ &$0.167$ \\
           Bidirectional & $\textbf{0.742}$ &$\textbf{0.123}$ \\
            \bottomrule
        \end{tabular}
        \vspace{+2.5pt}
        \caption{Bidirectional Training}
    \end{subtable}
    }
    
    \label{tab:forback}
\vspace{-10pt}
\end{table}

\noindent \textbf{Temporal Loss.} As in \reftab{}~\ref{tab:clipop} (a), without the temporal loss as explicit supervision, our \sbn{} can enforce temporal consistency. Adding the temporal loss can further remove flickers and improve temporal consistency.

\noindent \textbf{Reference Frame Intervals.} We denote the inter-frame intervals as $l$. As shown in \reftab{}~\ref{tab:clipop} (b), $l=1$ attains the best performance in our experiments.

\begin{figure}[!t]
\begin{center}
   \includegraphics[width=0.495\textwidth,trim=0 0 0 0,clip]{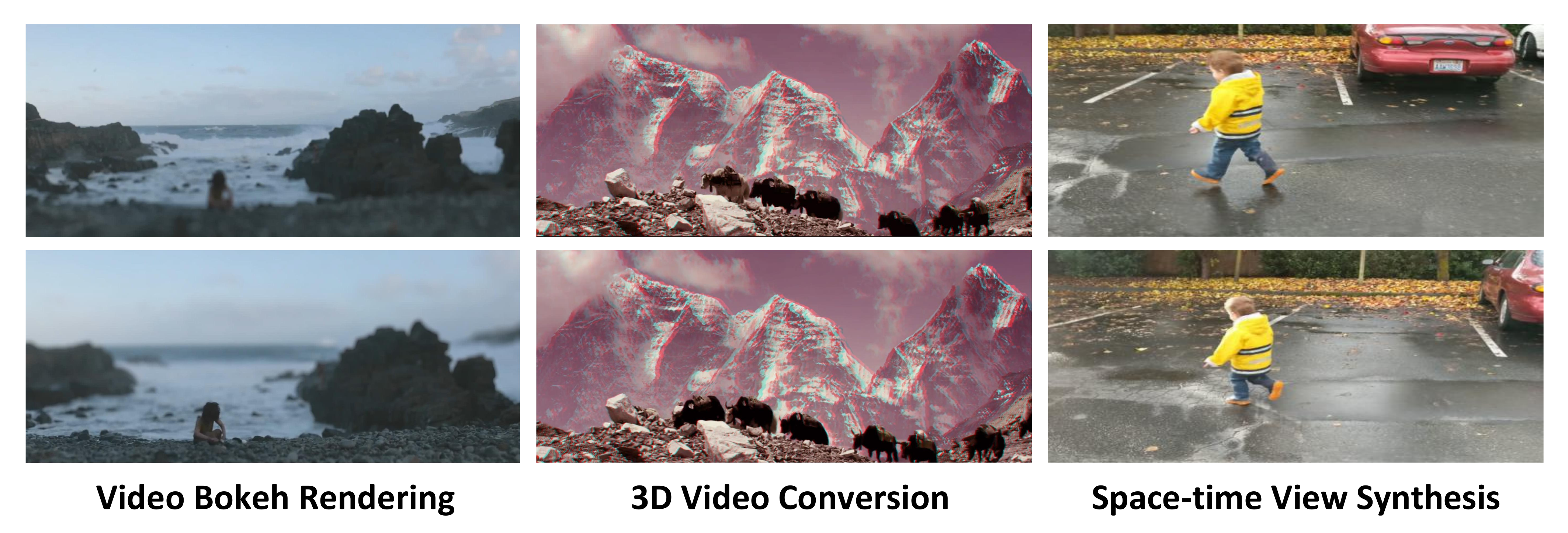}
\end{center}
\vspace{-7pt}
   \caption{\textbf{Depth-based applications.} The consistent and accurate results from \sx{} can be directly applied to various downstream video applications, \textit{e.g.}, 3D video conversion~\cite{n1}, video bokeh rendering~\cite{bokehme,videobokeh}, and space-time view synthesis~\cite{nsff,dynibar}. Best view zoomed in on-screen.}
\label{fig:effect}
\end{figure}

\begin{figure}[!t]
\begin{center}
   \includegraphics[width=0.480\textwidth,trim=0 0 0 0,clip]{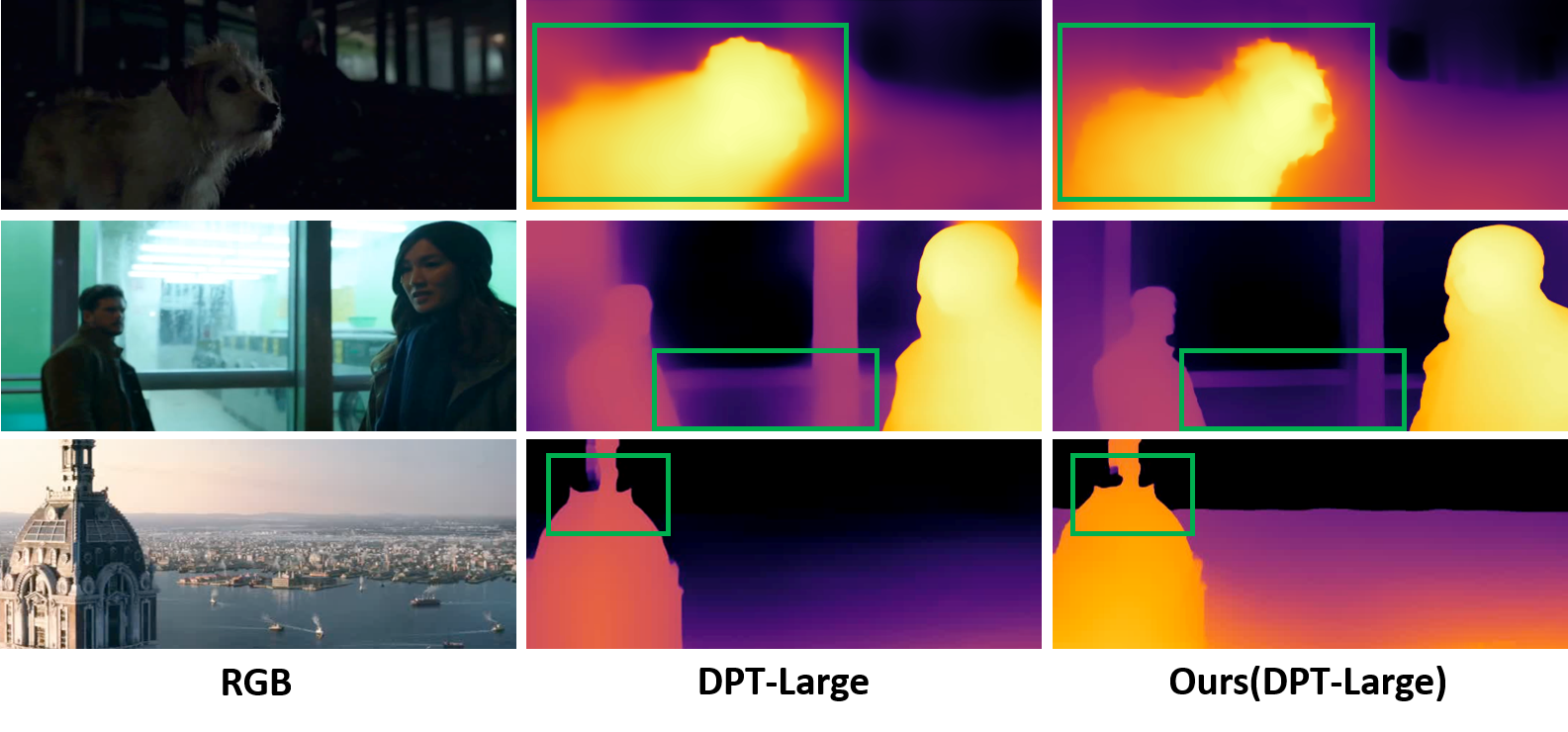}
\end{center}
\vspace{-7pt}
   \caption{\textbf{Failure cases.} The analysis of failure cases could inspire future research avenues. In future work, more advanced techniques could be explored to enhance the robustness of video depth models against adverse lighting conditions, transparent surfaces, and specular reflections.}
\label{fig:failc}
\vspace{-7pt}
\end{figure}

\noindent \textbf{Reference Frame Numbers.} As shown in \reftab{}~\ref{tab:nostab} (b), using three reference frames ($n$=3) for a target frame 
achieves the best results. 
More reference frames ($n$=4) increase computational costs but bring no improvement, which can be caused by the temporal redundancy of videos. Overall, we adopt three reference frames with the inter-frame interval $l=1$ for other experiments. The default setting works well for most videos. However, no single setting can be optimal for all videos. Different scenes, frame rates, objects, and motions in videos could all have an impact. For instance, we further discuss varied frame rates in the supplementary document.

\noindent \textbf{Baselines without Stabilization Network.} With DPT-Large~\cite{dpt} as the depth predictor, we train and evaluate two baselines without the stabilization network on the same subset as \reftab{}~\ref{tab:clipop}. We use the same temporal window as \sx{}. The first baseline (Single-frame) can only process each frame independently. Temporal window and loss $\mathcal{L}_t$ are used for consistency. The second baseline (Multi-frame) uses neighboring frames concatenated by channels to predict disparity of the target frame.
Training and inference strategies are both kept the same as \sx{}. As shown in the \reftab{}~\ref{tab:nostab} (a), temporal flickers cannot be solved by simply adding temporal windows and training loss on baselines. Proper designs are needed for inter-frame correlations. Our stabilization network improves consistency significantly.

\noindent \textbf{Forward Training and Backward Inference.} There exists a modest domain gap between forward training and backward inference. Certain scenarios that occur naturally in the forward direction, such as water spilling from a cup, cannot be realistically reversed to depict water returning to the cup. Nevertheless, given that a reversed video is intrinsically still a video, the domain gap has minor impacts on the model. As presented in \reftab{}~\ref{tab:forback}, we compare the models trained on forward and bidirectional sequences. The subtle change in performance indicates that the model can handle this small domain gap. Bidirectional training would further escalate the training costs. Thus, considering the trade-off between performance and training burdens, we implement the main training process of all our models in the forward direction.

\begin{figure}[!t]
\begin{center}
   \includegraphics[width=0.490\textwidth,trim=10 0 0 0,clip]{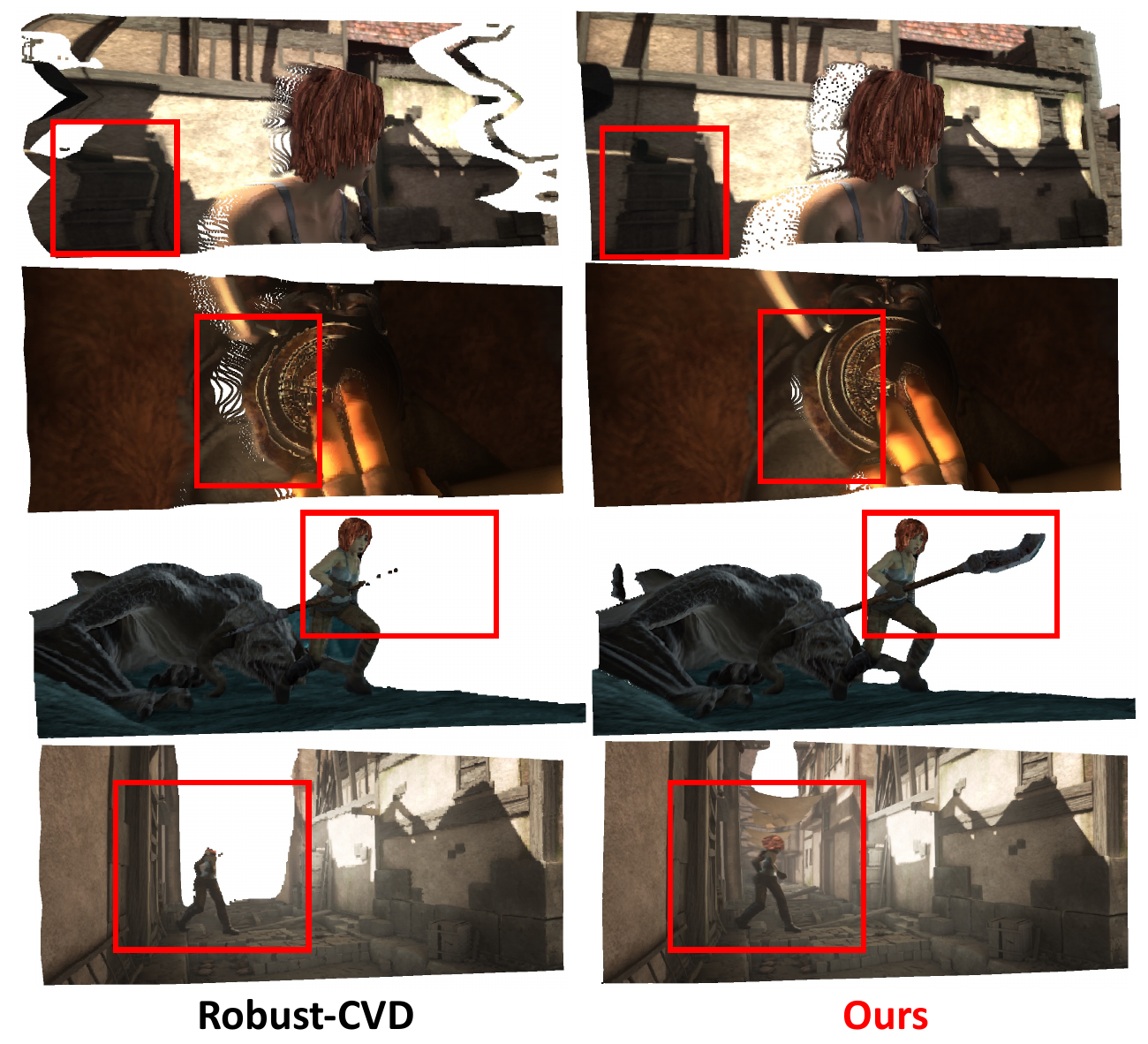}
\end{center}
\vspace{-7pt}
   \caption{\textbf{Point cloud reconstructions.} Based on the videos of the Sintel dataset~\cite{sintel}, we render the point clouds through Open3D~\cite{open3d} from a novel view point. We compare the reconstruction results of Robust-CVD~\cite{rcvd} and our \sx{}. Due to the depth errors and artifacts produced by Robust-CVD~\cite{rcvd}, their reconstructions exhibit noticeable distortion, deformation, and structural incompleteness. In contrast, our \sx{} framework can produce point clouds with better spatial geometry and structural integrity.}
\label{fig:pointkshksh}
\vspace{-7pt}
\end{figure}

\subsection{Depth-based Applications}
\label{sec:dbapp}
With any in-the-wild monocular video and the disparity results predicted by our \sx{}, a multitude of downstream applications can be implemented. As shown in \reffig{}~\ref{fig:effect}, we apply our disparity results to 3D video conversion~\cite{n1}, video bokeh rendering~\cite{bokehme,videobokeh}, and space-time view synthesis~\cite{nsff,dynibar}. The consistent and robust predictions from \sx{} can boost various depth-based applications.

Furthermore, depth maps serve as one of the bridges linking 2D and 3D spaces. Thus, merely verifying the performance of depth prediction is insufficient. It is also essential to carry out 3D reconstructions with integral and accurate spatial geometric structures, which can truly demonstrate the applicability of video depth models in 3D applications. As shown in \reffig{}~\ref{fig:pointkshksh}, we utilize depth maps to reconstruct point clouds in 3D space. We compare the reconstruction results of Robust-CVD~\cite{rcvd} and our \sx{}. 
The reconstructions from Robust-CVD~\cite{rcvd} exhibit noticeable distortion, deformation, and structural incompleteness due to their depth errors and artifacts. In contrast, our
\sx{} framework can produce point clouds with better spatial geometry and structural integrity. The results demonstrate the applicability of \sx{} in the field of 3D vision.

\subsection{Failure Cases}
Our \sx{} can seamlessly stabilize different depth predictors, predicting both consistent and accurate depth results. However, some drawbacks still exist. 
As shown in \reffig{}~\ref{fig:failc}, several subtle failures occur in regions with adverse lighting conditions, such as the blurred contour of the dog at night, the depth error of the transparent window, and the incomplete tower tip under light overexposure. 

These failures could be attributed to two factors. Firstly, depth errors from depth predictors affect \sx{}. Both DPT-Large~\cite{dpt} and \sx{} produce unsatisfactory results in these areas. Additionally, the suboptimal predictions may result from inaccurate ground truth. The disparity annotated by optical flow~\cite{gmflow,flownet2} is not entirely reliable for objects with specular reflections or transparency. 

Analyzing these failure cases could inspire future research avenues to enhance the robustness of depth models against adverse lighting conditions, challenging weather, and transparent surfaces. Some recent attempts~\cite{ww1,ww2,ww3} have started to explore these directions. Overall, we aim to develop consistent video depth models with strong robustness and generality in various in-the-wild scenarios.


\section{Conclusion}
In this paper, we propose a \sx{} framework and a large-scale natural-scene \data{} dataset for video depth estimation. Different from previous learning-based video depth models that function as stand-alone models, our \sx{} learns to stabilize the flickering results from the estimations of single-image depth models. In this way, \sx{} can focus on the learning of temporal consistency, while inheriting the depth accuracy from the depth predictors without further tuning. We also elaborate on the \data{} dataset to alleviate the data shortage. To our best knowledge, it is currently the largest video depth dataset in the wild. Besides, we also propose a bidirectional inference strategy with \flow{} to further improve temporal consistency by adaptively fusing forward and backward predictions. We instantiate a comprehensive model family with
variants ranging from small to large scales to balance efficiency and performance. To further prove the versatility of our \sx{} framework in video dense prediction and other downstream applications, we extend \sx{} to video semantic segmentation and applications like bokeh rendering, novel view synthesis, and 3D reconstruction. We hope our work can serve as a solid baseline and provide a data foundation for the learning-based video dense prediction.



\bibliographystyle{IEEEtran}
\bibliography{main}

\vfill

\newpage
\begin{IEEEbiography}[{\includegraphics[width=1in,height=1.25in,clip,keepaspectratio]{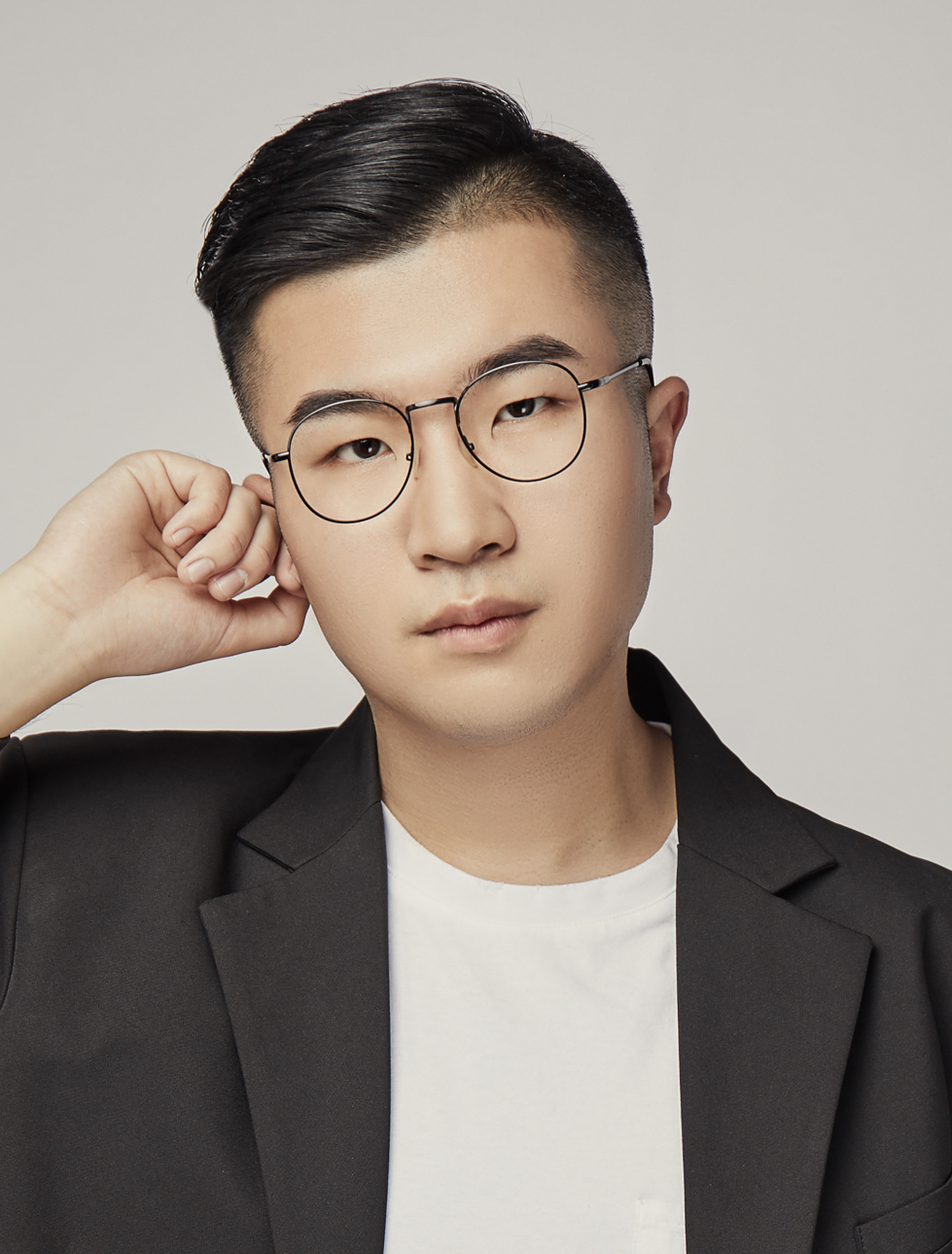}}]{Yiran Wang} received the B.S. degree from Huazhong University of Science and Technology, Wuhan, China, in 2021. He is currently pursuing the Ph.D. degree with the School of Artificial Intelligence and Automation, Huazhong University of Science and Technology, Wuhan, China. His research interests include computer vision and pattern recognition, with particular emphasis on depth estimation, video consistency, dense prediction, and 3D vision.
    
\end{IEEEbiography}

\vskip 0pt plus -1fil
\begin{IEEEbiography}[{\includegraphics[width=1in,height=1.25in,clip,keepaspectratio]{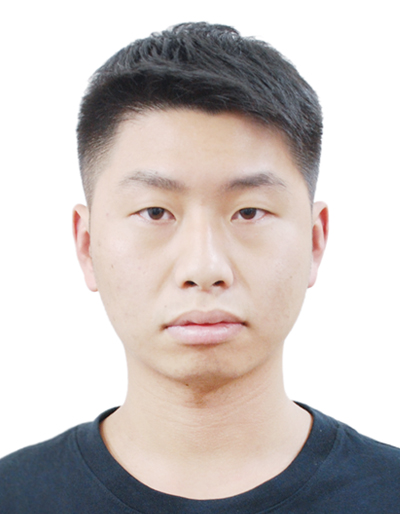}}]{Min Shi} received the B.S. degree and M.S. degree from the School of Artificial Intelligence and Automation, Huazhong University of Science and Technology, China. He is interested in computer vision and deep learning, few-shot learning, multi-modal models, and 3D vision.
    
\end{IEEEbiography}

\vskip 0pt plus -1fil
\begin{IEEEbiography}[{\includegraphics[width=1in,height=1.25in,clip,keepaspectratio]{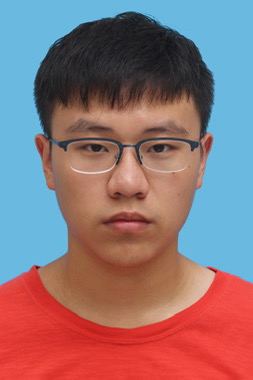}}]{Jiaqi Li} received the B.S. degree from Huazhong University of Science and Technology, Wuhan, China, in 2023. He is currently pursuing the M.S. degree with the School of Artificial Intelligence and Automation, Huazhong University of Science and Technology, Wuhan, China. His research interests lie in 3D Vision, with particular emphasis on monocular and multi-view depth estimation.
\end{IEEEbiography}

\vskip 0pt plus -1fil
\begin{IEEEbiography}[{\includegraphics[width=1in,height=1.25in,clip,keepaspectratio]{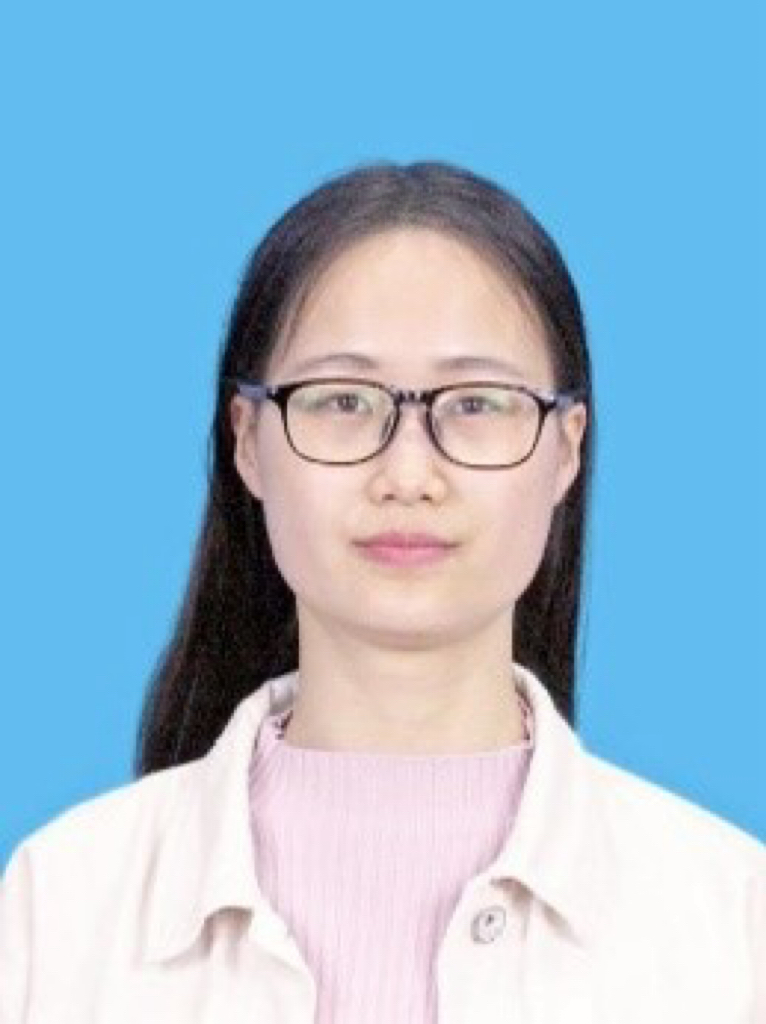}}]{Chaoyi Hong}
		received the B.S. degree from Huazhong University of Science and Technology, Wuhan, China, in 2019. She is pursuing a Ph.D. degree with the School of Artificial Intelligence and Automation, Huazhong University of Science and Technology. Her research interests include computer vision and deep learning, focusing on dense prediction and aesthetics-related tasks, particularly on image matting and image cropping.
\end{IEEEbiography}

\vspace{+15pt}
\vskip 0pt plus -1fil
\begin{IEEEbiography}[{\includegraphics[width=1in,height=1.25in,clip,keepaspectratio]{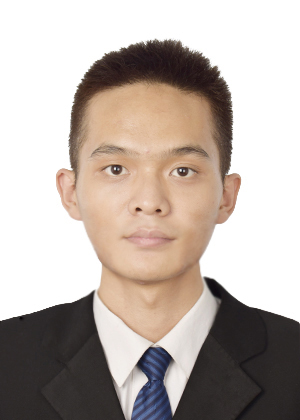}}]{Zihao Huang}
		received the B.S. degree from Huazhong University of Science and Technology, Wuhan, China, in 2023. He is currently pursuing the M.S. degree with the School of Artificial Intelligence and Automation, Huazhong University of Science and Technology, Wuhan, China. His research interests mainly include 3D avatar creation, focusing on fast and high-quality 3D human reconstructions.
\end{IEEEbiography}

\vskip 0pt plus -1fil
\begin{IEEEbiography}[{\includegraphics[width=1in,height=1.25in,clip,keepaspectratio]{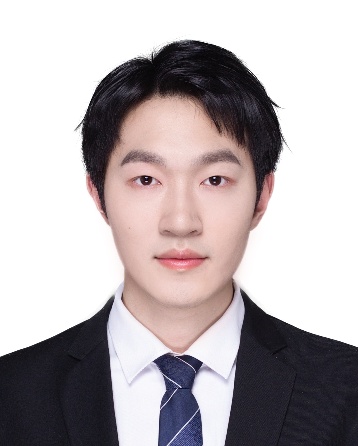}}]{Juewen Peng}
		received the B.S. and M.S. degrees from the School of Artificial Intelligence and Automation, Huazhong University of Science and Technology. He is currently pursuing the Ph.D. degree in the College of Computing and Data Science, Nanyang Technological University, Singapore. His research interests include computational photography, bokeh rendering, image deblurring, and 3D vision.
\end{IEEEbiography}

\vskip 0pt plus -1fil
\begin{IEEEbiography}[{\includegraphics[width=1in,height=1.25in,clip,keepaspectratio]{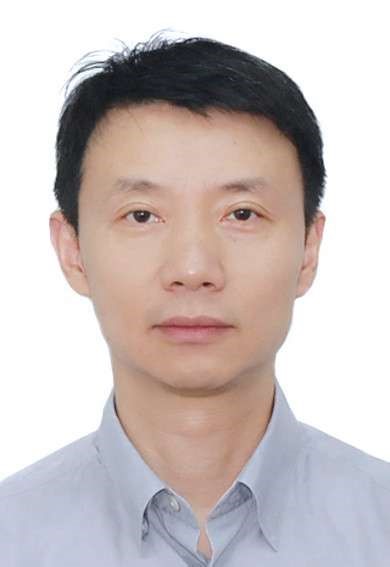}}]{Zhiguo Cao}
		(Member, IEEE) is currently a Professor with the School of Artificial Intelligence and Automation, Huazhong University of Science and Technology. His research interests spread across computational photography, monocular depth estimation, 3d video processing,
motion detection, and human action analysis. He has published dozens of papers in international journals and prominent conferences.
\end{IEEEbiography}

\vskip 0pt plus -1fil
\begin{IEEEbiography}[{\includegraphics[width=1in,height=1.25in,clip,keepaspectratio]{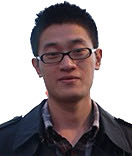}}]{Jianming Zhang}
		received the B.S. and M.S. degrees in mathematics from Tsinghua University, Beijing, China, in 2008 and 2011, respectively, and the Ph.D. degree in computer science from Boston University, Boston, Massachusetts, in 2016. He is a principal research scientist at Adobe. His research interests include visual saliency, image segmentation, 3D understanding, and generative models.
\end{IEEEbiography}

\vskip 0pt plus -1fil
\begin{IEEEbiography}[{\includegraphics[width=1in,height=1.25in,clip,keepaspectratio]{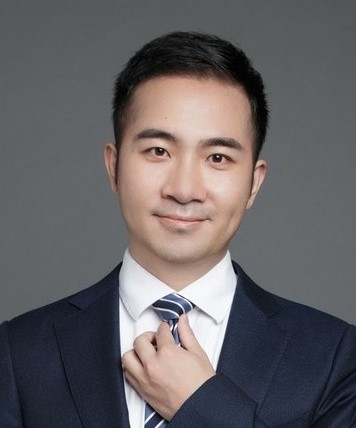}}]{Ke Xian}
		received the B.S. and Ph.D. degrees from Huazhong University of Science and Technology (HUST), China. From August 2016 to September 2017, he worked as a joint training Ph.D. student with the University of Adelaide, Australia. From October 2021 to November 2023, he worked as a Research Fellow with the S-Lab, Nanyang Technological University (NTU), Singapore. Now, he is a lecturer with the School of Electronic Information and Communications, Huazhong University of Science and Technology. His research interests include robust 3D vision, 2D scene understanding, and computational photography.
\end{IEEEbiography}

\vskip 0pt plus -1fil
\begin{IEEEbiography}[{\includegraphics[width=1in,height=1.25in,clip,keepaspectratio]{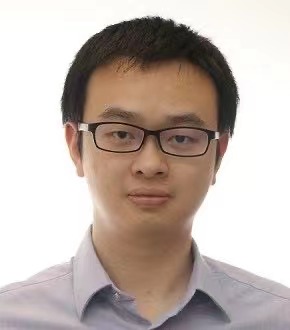}}]{Guosheng Lin}
		is an Associate Professor at the College of Computing and Data Science, Nanyang Technological University, Singapore. He received his Ph.D. degree from The University of Adelaide in 2014. His research interests are in computer vision and machine learning including scene understanding, 3D vision and generative learning.
\end{IEEEbiography}

\ifCLASSOPTIONcaptionsoff
  \newpage
\fi

\newpage
 
\appendices


\section{More details on the \data{} Dataset}
\subsection{Releasing of the VDW Dataset.} 
We have released the VDW dataset under strict conditions. We must ensure that the release won’t violate any copyright requirements. To this end, we will not release any video frames or the derived data in public. Instead, we provide metadata and detailed toolkits, which can be used to reproduce VDW or generate your own data. All the metadata and toolkits are licensed under CC BY-NC-SA 4.0~\cite{cccc}, which can only be used for academic and research purposes. Refer to the VDW website \url{https://raymondwang987.github.io/VDW/} for more information.


\subsection{Dataset Construction}
\noindent\textbf{Data Acquisition and Pre-processing.} Here we add more details on data acquisition and pre-processing (Sec.~4, page 7, main paper). Having obtained the raw videos, we use FFmpeg~\cite{ffmpeg} and PySceneDetect~\cite{pyscene} to split all the videos into $104@582$ sequences. We manually check and remove the duplicated, chaotic, and blur scenes. Videos that are wrongly split by the scene detect tools are also removed. Finally, we reserve $32@405$ videos with more than six million frames for disparity annotation.

\begin{figure}[!t]
\begin{center}
   \includegraphics[width=0.47\textwidth,trim=0 0 0 0,clip]{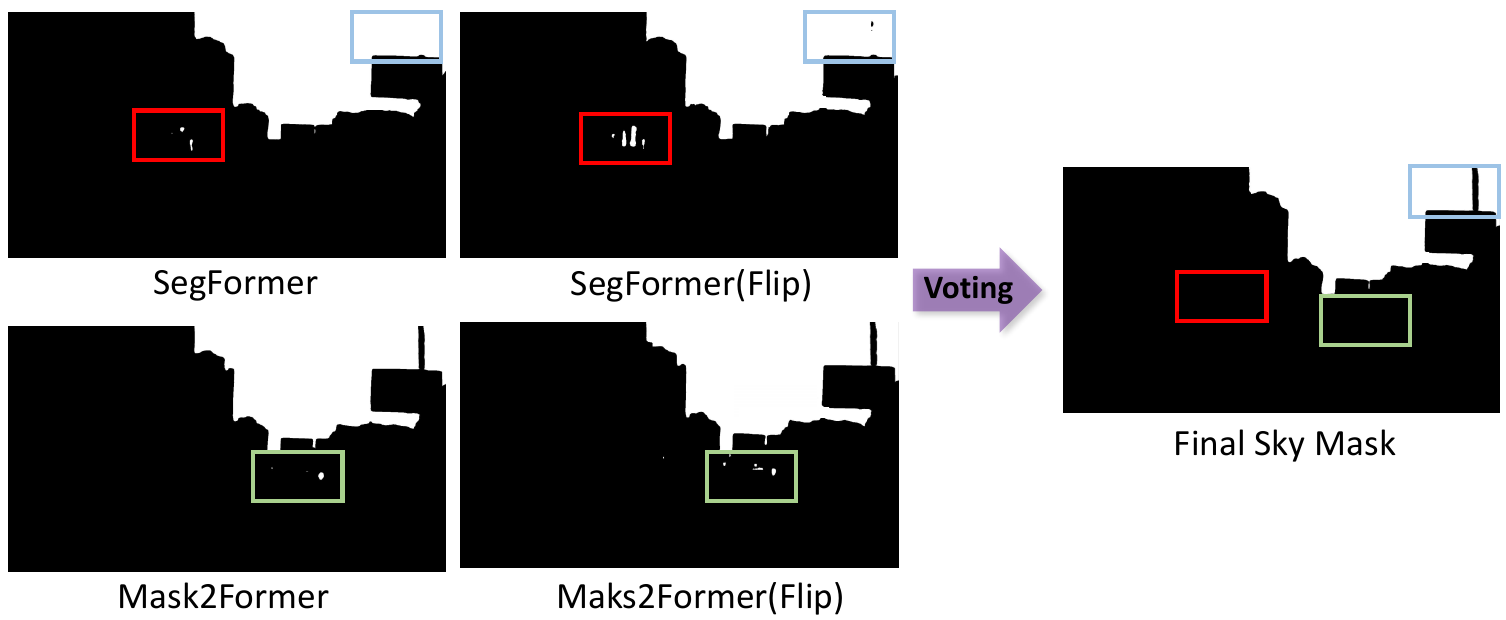}
\end{center}
\vspace{-5pt}
   \caption{
   \textbf{Model ensemble strategy for sky segmentation on \data{} dataset.} White area represents sky regions. Errors and noises in the rectangles are removed by model ensemble and voting, which improves the quality of the ground truth.}
\label{fig:skyvote}
\end{figure}

\noindent\textbf{Disparity Annotation.} In Sec.~4 
of the main paper, we mentioned that the disparity ground truth is obtained via sky segmentation and optical flow estimation. Here we specify the details. Compared with common practice~\cite{kexian2018,MiDaS}, we introduce a few engineering improvements to make the disparity maps more accurate. As the sky is considered to be infinitely far, pixels in the sky regions should be segmented and set to the minimum value in the disparity maps. We find that using a single segmentation model~\cite{MiDaSseg,FFM1} like prior arts~\cite{MiDaS,kexian2018} causes errors and noises in the sky regions. Hence, we generate the sky masks in a model ensemble manner. Each frame along with its horizontally flipped copy are fed into two \sota{} semantic segmentation models SegFormer~\cite{segformer} and Mask2Former~\cite{mask2former}, which yields four sky masks in total. A pixel is considered as the sky when it is positive in more than two predicted sky masks. Besides, we also fill the connected regions with less than $50$ pixels to further remove the noisy holes in the sky masks. Such ensemble strategy can improve the quality of the ground truth as shown in \reffig{}~\ref{fig:skyvote}, and consequently improves the performance of the trained models, especially on skylines as shown in \reffig{}~\ref{fig:ksh1}.

Following the practice of previous single-image depth datasets~\cite{kexian2018,MiDaS}, we adopt a \sota{} optical flow model GMFlow~\cite{gmflow} to generate the ground truth disparity of the left- and right-eye views. The estimated optical flow is bidirectional. We perform a consistency check between the optical flow pairs to obtain the valid masks for training. We adopt the adaptive consistency threshold for each pixel as~\cite{unflow}. The ground truth of each video is normalized by its minimum and maximum disparity. Then, the disparity value is discretized into $65@535$ intervals. \reffig{}~\ref{fig:datagtmore} shows more examples of our \data{} dataset.


\begin{figure*}[!t]
\begin{center}
   \includegraphics[width=\textwidth,trim=360 0 360 220,clip]{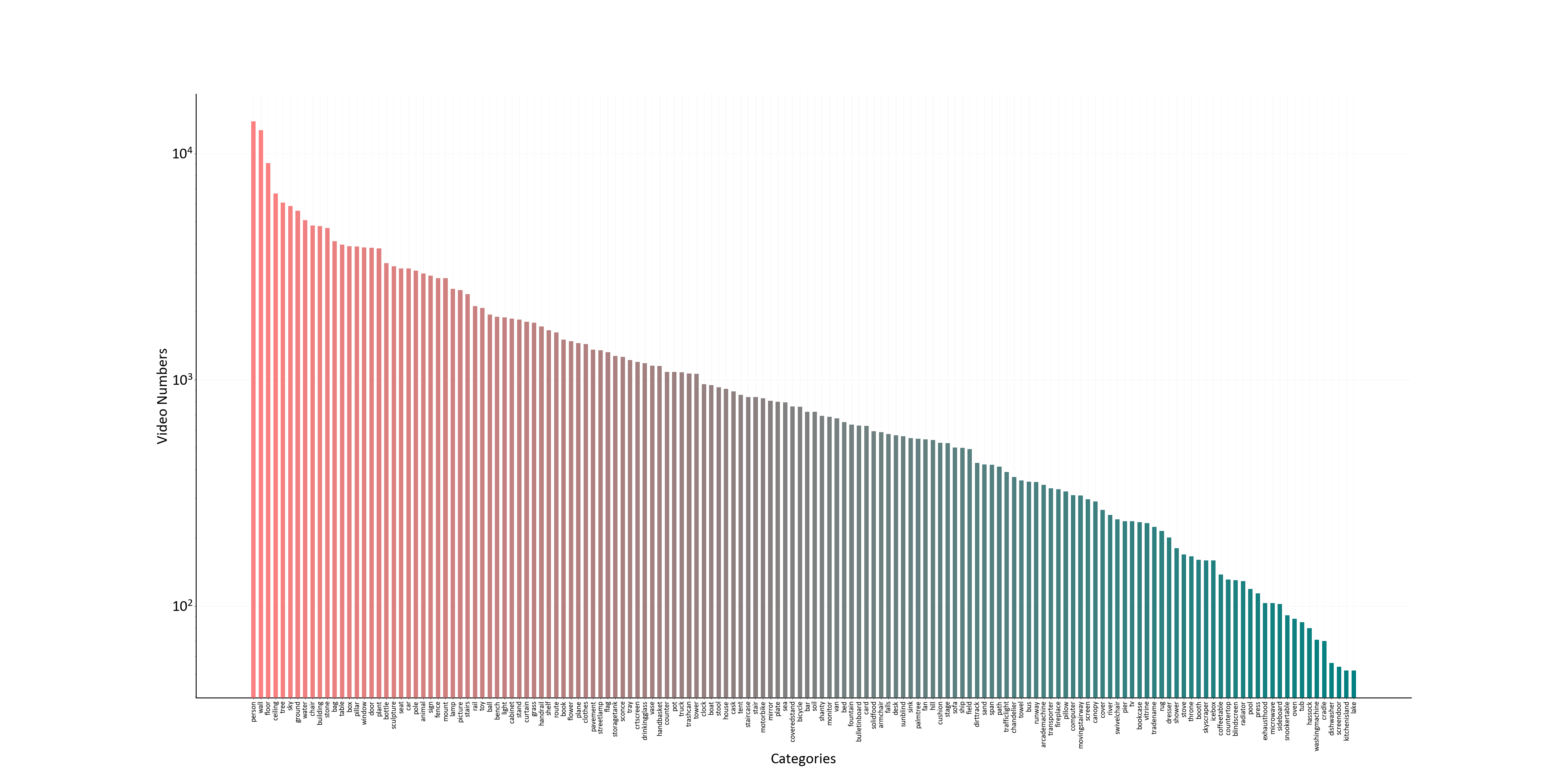}
\end{center}
\vspace{-10pt}
   \caption{
   The statistics of the $150$ semantic categories in \data{} dataset.}
\label{fig:150}
\end{figure*}

\begin{table}[!t]
\caption{\textbf{Video and frame numbers statistics of VDW training set.} Our \data{} dataset contains $14@203$ videos from movies, animations, documentaries, and web videos.}
\label{tab:dsc}
\begin{center}
\resizebox{0.975\columnwidth}{!}{
\begin{tabular}{llcc}
\toprule
Sources & Titles & Videos & Frames \\
\midrule
\multirow{4}{*}{Documentaries} & Deepsea Challenge & $210$ & $38@078$ \\
& Kingdom of Plants & $253$ & $95@742$ \\
& Little Monsters & $242$ & $50@420$ \\
& Jerusalem & $37$ & $21@574$ \\
\midrule
\multirow{2}{*}{Animations} & Coco & $1@079$ & $146@002$ \\
& Kung Fu Panda 3 & $959$ & $68@405$ \\
\midrule
\multirow{18}{*}{Movies} & Exodus: Gods and Kings & $1@339$ & $99@146$ \\
& Geostorm & $857$ & $52@028$ \\
& Hugo & $301$ & $25@091$ \\
& Mission: Impossible-Fallout & $664$ & $46@344$ \\
& Noah & $1@160$ & $85@161$ \\
& Pompeii & $158$ & $10@112$ \\
& Spider-Man: No Way Home & $914$ & $75@077$ \\
& The Legend of Tarzan & $735$ & $64@840$ \\
& The Three Musketeers & $253$ & $18@180$ \\
& Gravity & $191$ & $38@332$ \\
& Silent Hill 2 & $72$ & $5@076$ \\
& Transformers: Age of Extinction & $1@323$ & $84@619$ \\
& Doctor Strange & $299$ & $23@779$ \\
& Battle of the Year & $454$ & $19@613$ \\
& Justice League & $428$ & $37@202$ \\
& The Hobbit 2 & $644$ & $53@391$ \\
& The Great Gatsby & $729$ & $49@079$ \\
& Billy Lynn's Long Halftime Walk & $242$ & $29@137$ \\
\midrule
\multirow{2}{*}{Web Videos} & YouTube & $514$ & $40@897$ \\
& bilibili & $146$ & $17@243$ \\
\midrule
All & $-$ & $14@203$ & $2@237@320$ \\
\bottomrule
\end{tabular}
}
\end{center}

\vspace{-14pt}

\end{table}

\noindent\textbf{Invalid Sample Filtering.} Having obtained the annotations, we further filter the videos that are not qualified for our dataset. According to optical flow and valid masks, samples with the following three conditions are removed: 1) more than $30\%$ of pixels in the consistency masks are invalid; 2) more than $10\%$ of pixels have vertical disparity larger than two pixels; 3) the average range of horizontal disparity is less than $15$ pixels. Then, we manually check all the videos along with their corresponding ground truth, and remove the samples with obvious errors. Finally, we retain $14@203$ videos with $2@237@320$ frames in \data{} dataset.
\begin{figure*}
\begin{center}
   \includegraphics[width=0.97\textwidth,trim=0 0 0 0,clip]{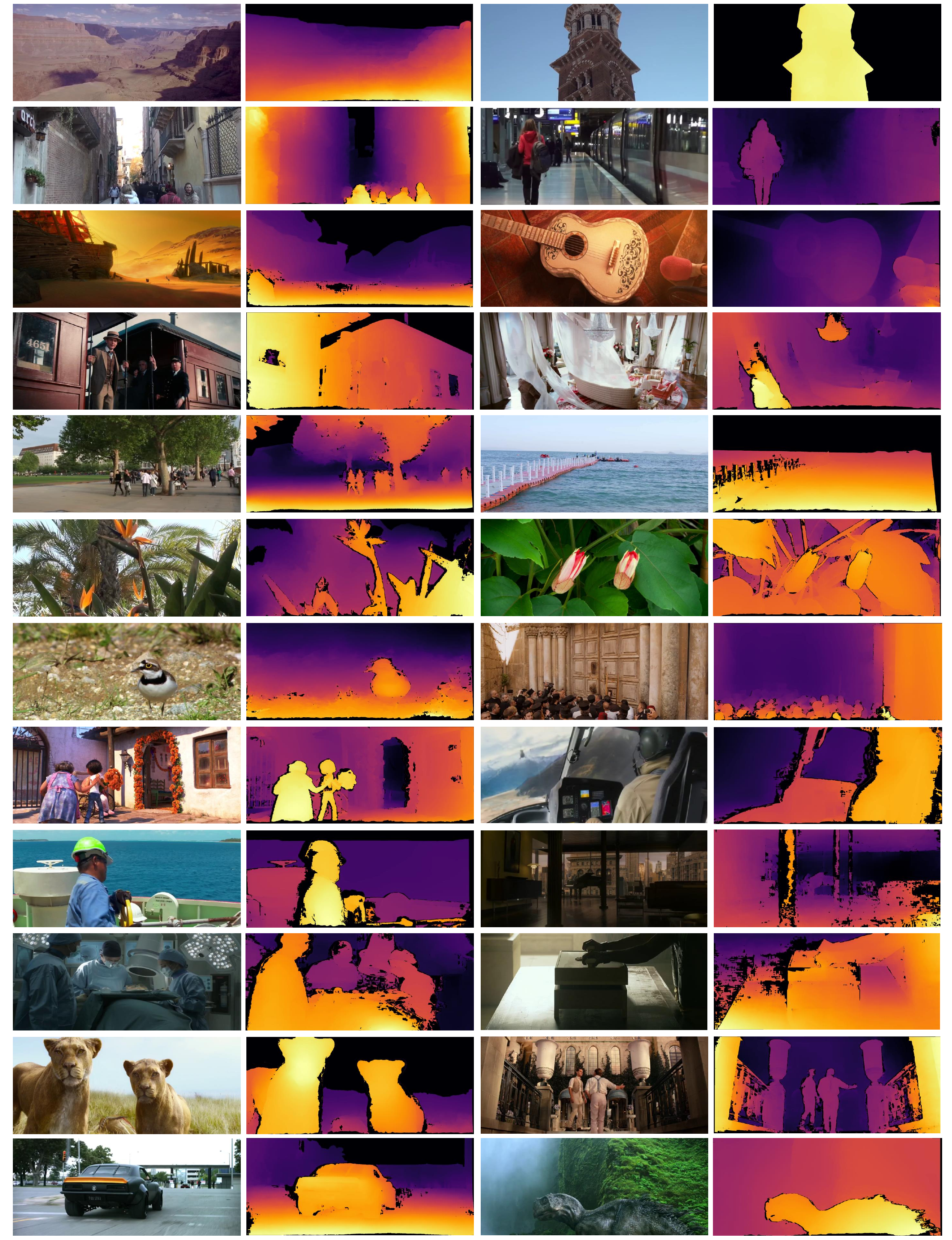}
\end{center}
\vspace{-12pt}
   \caption{
   \textbf{More examples of our \data{} dataset.} Sky regions and invalid pixels are masked out.}
\label{fig:datagtmore}
\end{figure*}

\begin{table}[!t]
\caption{\textbf{Video and frame numbers statistics of VDW test set.} VDW test set adopts different data sources from training data, \textit{i.e.}, different movies, web videos, or animations.}
\label{tab:test}
\begin{center}
\resizebox{0.975\columnwidth}{!}{
\begin{tabular}{llcc}
\toprule
Sources & Titles & Videos & Frames \\
\midrule
\multirow{3}{*}{Movies} & Eternals & $39$ & $4@802$ \\
& Everest & $17$ & $2@922$ \\
& Fantastic Beasts and Where to Find Them & $17$ & $27@27$ \\
\midrule
\multirow{1}{*}{Animation} & Frozen 2 & $10$ & $1@098$ \\
\midrule
Web Videos & bilibili & $7$ & $1@073$ \\
\midrule
All & $-$ & $90$ & $12@622$ \\
\bottomrule
\end{tabular}
}
\end{center}
\vspace{-15pt}

\end{table}

\subsection{Data Statistics}
\label{sec:ddtj}
\noindent \textbf{Data Sources.} Taking over $6$ months to process, \data{} training set contains $14@203$ videos with $2@237@320$ frames. The detailed data sources of training set and test set are listed in \reftab{}~\ref{tab:dsc} and \reftab{}~\ref{tab:test} respectively. 

\noindent \textbf{Frame Rates and Frame Numbers.} For all our sequences, the lowest frame rate is $12$ fps, the highest frame rate is $60$ fps, and the average frame rate is $28.92$ fps. Even some special videos, such as fast-forward or slow-motion sequences, are included in the VDW dataset. The minimum frame number is $18$ while the maximum is $8@005$.

\noindent \textbf{Objects Presented in the VDW Dataset.} To verify the diversity of objects in our videos. We conduct semantic segmentation with Mask2Former~\cite{mask2former} trained on ADE20K~\cite{ade20k}. All the $150$ categories are covered in our dataset. The five categories that present most frequently are person $(97.2\%)$, wall $(89.1\%)$, floor $(63.5\%)$, ceiling $(46.5\%)$, and tree $(42.3\%)$. Each category can be found in at least $50$ videos. \reffig{}~\ref{fig:150} shows the detailed statistics of all the $150$ categories.

\subsection{Discussions of Data Characteristics} 
\noindent \textbf{Animations.} To enhance the diversity and generality of our VDW dataset, we include a small portion of animations (around 9\% of frames), while the majority (91\%) consists of real-world videos. The combination allows models to generalize well in natural scenes and produce robust predictions for animated videos. This could benefit various tasks that involve stylized videos, such as 3D video conversion, virtual reality, and video editing. Users can decide which parts to use, depending on their tasks and data requirements.

\noindent \textbf{Disparity of Stereo Films.} We mainly pursue data scale, diversity, and generality. Using other methods (\textit{e.g.}, LiDAR, or Kinect) to annotate over 2 million frames in diverse scenes would incur much higher costs or may even be impractical. Thus, we adopt the disparity of stereo videos, which are more accessible and effective. However, the downside is that the disparity in stereo movies is not always trustworthy, as it could be adjusted for viewing comfort. We have implemented several measures to alleviate this problem, including removing overly unrealistic sequences and conducting rigorous data checking. Users can also combine the VDW with other datasets for training.



\begin{figure}[!t]
\begin{center}
   \includegraphics[width=0.40\textwidth,trim=0 0 0 0,clip]{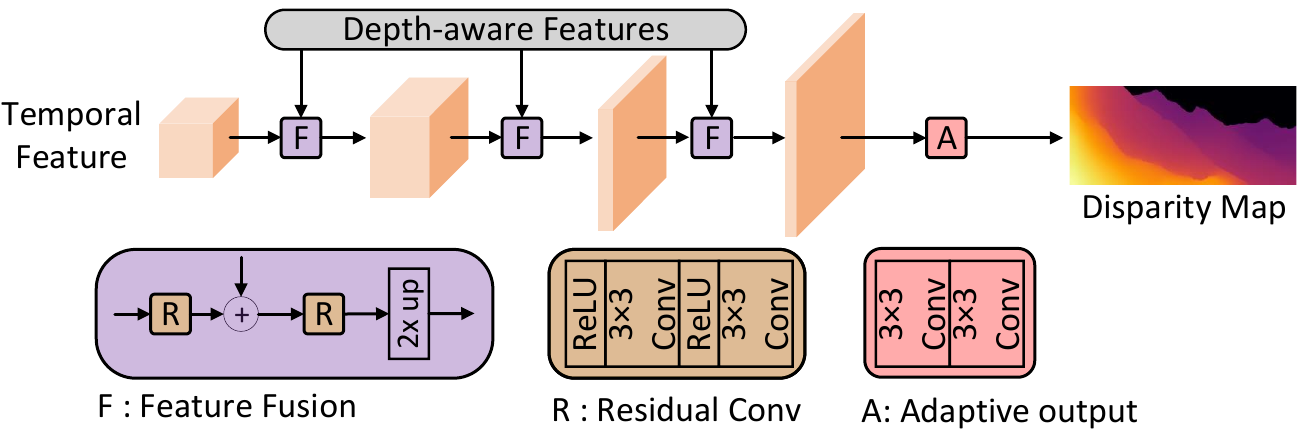}
\end{center}
\vspace{-5pt}
   \caption{\textbf{The decoder architecture for depth estimation.}}
\label{fig:decoder}
\end{figure}

\section{More Implementation Details for \sx{}}
\subsection{Decoder Architecture}
Here we specify the decoder architecture for video depth estimation. The decoder architecture is illustrated in \reffig{}~\ref{fig:decoder}. To fuse the depth-aware features from the backbone~\cite{segformer} and temporal features from the cross-attention module, feature fusion modules (FFM)~\cite{FFM1,FFM2} and skip connections are adopted. Resolutions are gradually increased while channel numbers are decreased. At last, we use an adaptive output module to adjust the channel and restore the disparity maps.

As for the decoder of video semantic segmentation, we apply the simple and common architecture as prior arts~\cite{cffm,segformer,mrcfa,SSLTM}. 

\subsection{Feature Encoder}
Feature Encoders~\cite{VIT,segformer,resnet,resnext,mvit} possess strong scene understanding and feature encoding capabilities, because of their comprehensive structural designs and model pre-training. Compared to the details in RGB images, feature encoders~\cite{VIT,segformer,resnet,resnext,mvit} extract high-level scene and semantic information with large receptive fields. Therefore, encoders with different structures, \textit{e.g.}, convolutional~\cite{resnest,resnet,resnext} or attention-based~\cite{VIT,mvit,segformer} backbones, have been widely used in dense prediction tasks such as depth estimation and semantic segmentation. Our \sx{} also employs feature encoders~\cite{segformer,resnet} to extract features.

To be specific, for video depth estimation, we adopt the Mit-b5~\cite{segformer} in our \llarge{} model to encode depth-aware features, considering its strong performance and capacity. For the lightweight \ssmall{} model, we utilize Mit-b0~\cite{segformer} to achieve real-time processing. Besides, in our experiments of video semantic segmentation, we follow SSLTM~\cite{SSLTM} to leverage ResNet-50~\cite{resnet} as the backbone, conducting fair comparisons with similar amounts of model parameters.

\subsection{Loss Function}
As mentioned in Sec.~3.2 in the main paper, the training loss for depth estimation consists of a spatial loss and a temporal loss. Here we specify the computation process. 

For the spatial loss, we adopt the widely-used affinity invariant loss and gradient matching loss~\cite{MiDaS,dpt} as $\mathcal{L}_s$. For the affinity invariant loss, let $D$ and $D^*$ denote the predicted disparity  and ground truth respectively, we first calculate the scale and shift:
\begin{equation}
    t(D) = median(D), 
    s(D)=\frac{1}{M}\sum_{i=1}^M|D_i-t(D_i)|\,,
\end{equation}
where $M$ denotes the number of valid pixels. The prediction and the ground truth are aligned to zero translation and unit scale as follows:
\begin{equation}
    \tilde{D} = \frac{D-t(D)}{s(D)},\tilde{D}^* = \frac{D^*-t(D^*)}{s(D^*)}.
\end{equation}
Then the affinity invariant loss can be formulated as:
\begin{equation}
    \mathcal{L}_{af} = \frac{1}{M}\sum_{i=1}^M|\tilde{D} - \tilde{D}^*|.
\end{equation}
Besides, we also adopt the multi-scale gradient matching loss~\cite{MiDaS}, which can improve smoothness of homogeneous regions and sharpness of discontinuities in the disparity maps. The gradient matching loss is formulated as:
\begin{equation}
    \mathcal{L}_{grad}=\frac{1}{M}\sum_{k=1}^K\sum_{i=1}^M(|\nabla_x R_i^k|+|\nabla_y R_i^k|),
\end{equation}
where $R_i = \tilde{D}_i - \tilde{D}^*_i$, and $R^k$ denotes the difference between the disparity maps at scale $k=1,2,3,\cdots,K$ (the resolution is halved at each level). Following DPT~\cite{dpt}, we set $K=4$ and set the weight $\mu$ of $\mathcal{L}_{grad}$ to $0.5$. The spatial loss can be expressed as:
\begin{equation}
    \mathcal{L}_s = \mathcal{L}_{af} + \mu \mathcal{L}_{grad},
\end{equation}

As for the spatial loss of semantic segmentation, we adopt the widely-used cross-entropy loss for supervision.

\noindent \textbf{Temporal Loss.}
In Sec.~3.2 of main paper, we mentioned that the temporal loss is masked with a visibility mask $O_{n \Rightarrow n-1}$ calculated from the warping discrepancy between frame $F_{n}$ and the warped frame $\hat{F}_{n-1}$. This mask is obtained by:
\begin{equation}
    O_{n \Rightarrow n-1} = \exp(-\gamma||F_{n}-\hat{F}_{n-1}||_2^2)\,.
\end{equation}
We set $\gamma = 50$ and use bilinear sampling layer for warping.

\subsection{Depth and Disparity}
Here, we illustrate the reasons for using disparity in our implementations. Firstly, our VDW dataset is annotated with the disparity from optical flow~\cite{gmflow}, making it straightforward for us to work with disparity. Secondly, we utilize different versions of MiDaS and DPT~\cite{dpt,MiDaS,MiDaSV31} as the initial predictors, which produce relative disparity maps. Keeping the input and output settings of \sx{} similar to those of MiDaS and DPT~\cite{dpt,MiDaS,MiDaSV31}, with disparity for training and inference, is convenient for the experiments. For other initial predictors that produce depth maps, their initial depth can be converted to disparity for input. 

Besides, we also discuss the advantages and disadvantages of disparity and depth. Disparity is more sensitive to objects at close distances and can better distinguish between foreground objects and the background, which is beneficial for downstream tasks such as bokeh rendering~\cite{bokehme,videobokeh}, 3D video conversion~\cite{n1}, and shallow depth of field effect~\cite{shallow}. On the other hand, depth maps can better differentiate distant objects, making them more suitable for autonomous driving tasks. Therefore, considering the applications in~\refsec{}~5.7 of the main paper, using disparity could be more convenient for our experiments.

\section{More Experimental Results}




\subsection{Depth Metrics}
Here we specify the evaluation metrics for depth accuracy. we adopt commonly-applied depth evaluation metrics: Mean relative error (Rel) and accuracy with threshold $t$. 

\noindent\textbf{Mean relative error (Rel):} $\frac{1}{M}\sum_{i=1}^M\frac{||D_i - D_i^*||_1}{D_i^*};$

\noindent\textbf{Accuracy with threshold $\textbf{t}$}: Percentage of $D_i$ such that $\\max(\frac{D_i}{D_i^*},\frac{D_i^*}{D_i}) = \delta<t\in\left[1.25, 1.25^2, 1.25^3\right]$, where $M$ denotes pixel numbers, $D_i$ and $D_i^*$ are prediction and ground truth of pixel $i$.


\begin{table}
\caption{\textbf{Comparisons on \data{} dataset.} The first $2$ rows show the results of different single-image depth predictors. The next $5$ rows contain video depth approaches. The last $2$ rows consist of the results of our \sx{}. Best performance is in boldface. Second best is underlined.}
\label{tab:vdw}
\begin{center}
\resizebox{\columnwidth}{!}{
\begin{tabular}{lcccccc}
\toprule
Method & $\delta_1\uparrow$ & $\delta_2\uparrow$ & $\delta_3\uparrow$ & $Rel\downarrow$ &$OPW\downarrow$ \\
\midrule

MiDaS-v2.1-Large~\cite{MiDaS} & $0.651$ & $0.857$ & $0.935$ & $0.288$ &$0.676$ \\
DPT-Large~\cite{dpt} & $\underline{0.730}$ & $\underline{0.894}$ & $\underline{0.952}$ & $\underline{0.215}$ &$0.470$ \\
\midrule
ST-CLSTM~\cite{ST-CLSTM} & $0.477$ & $0.709$ & $0.838$ & $0.521$ &$0.448$ \\
FMNet~\cite{fmnet} & $0.472$ & $0.716$ & $0.837$ & $0.514$ &$0.402$ \\
DeepV2D~\cite{deepv2d} & $0.546$ & $0.722$ & $0.835$ & $0.528$ &$0.427$ \\
WSVD~\cite{wsvd} & $0.637$ & $0.831$ & $0.914$ & $0.314$ &$0.462$ \\
Robust-CVD~\cite{rcvd} & $0.676$ & $0.855$ & $0.928$ & $0.261$ &$0.279$ \\

\midrule
Ours-Large(MiDaS-v2.1-Large) & $0.701$ & $0.885$ & $0.947$ & $0.239$ &$\underline{0.148}$ \\
Ours-Large(DPT-Large) & $\textbf{0.742}$ & $\textbf{0.897}$ & $\textbf{0.957}$ & $\textbf{0.208}$ & $\textbf{0.129}$ \\
\bottomrule
\end{tabular}
}
\end{center}
\vspace{-5pt}

\end{table}

\begin{figure}[!t]
\begin{center}
   \includegraphics[width=0.495\textwidth,trim=0 0 0 10,clip]{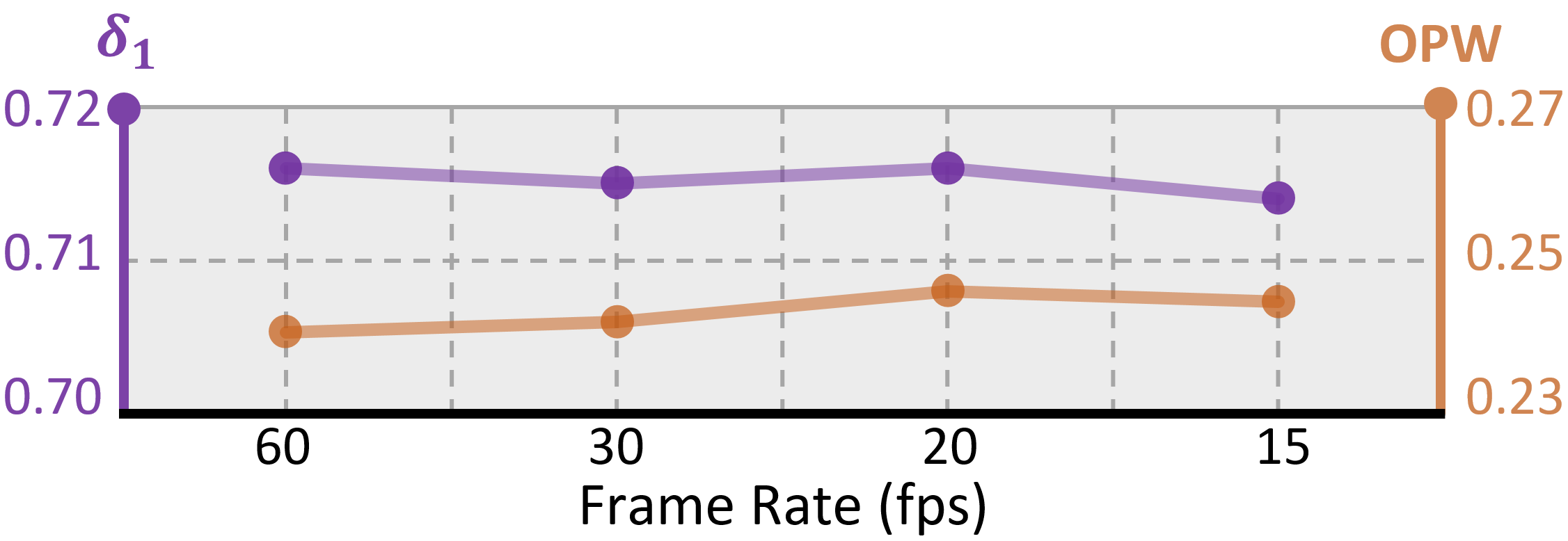}
\end{center}
\vspace{-7pt}
   \caption{\textbf{Robustness across different frame rates.} The temporal window contains three reference frames. For a certain video, we conduct evaluations under frame rates from $15$ to $60$ fps. \sx{} only exhibits minimal performance fluctuations, proving our robustness on various frame rates.}
\label{fig:fpsexp}
\end{figure}

\begin{figure*}
\begin{center}
   \includegraphics[width=0.97\textwidth,trim=0 0 0 0,clip]{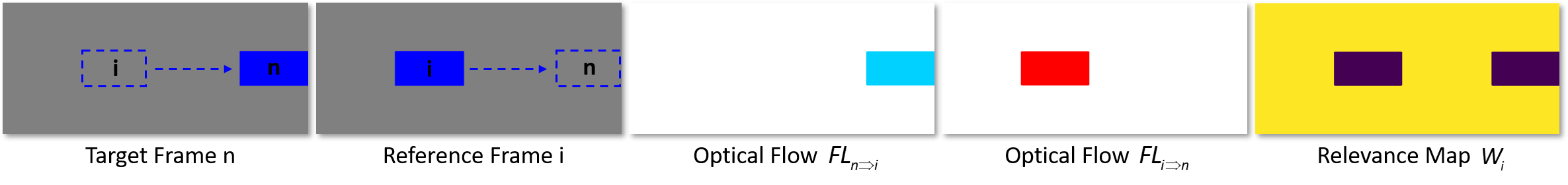}
\end{center}
\vspace{-12pt}
   \caption{
   \textbf{The principle for \flow{}.} We conduct a toy experiment to illustrate the principle, proving that the \flow{} strategy does not introduce systematic errors in the presence of motion. For the relevance map $W_i$, brighter colors indicate higher values, while darker colors indicate lower fusion weights.}
\label{fig:toy}
\end{figure*}

\begin{figure*}
\begin{center}
   \includegraphics[width=0.97\textwidth,trim=0 0 0 0,clip]{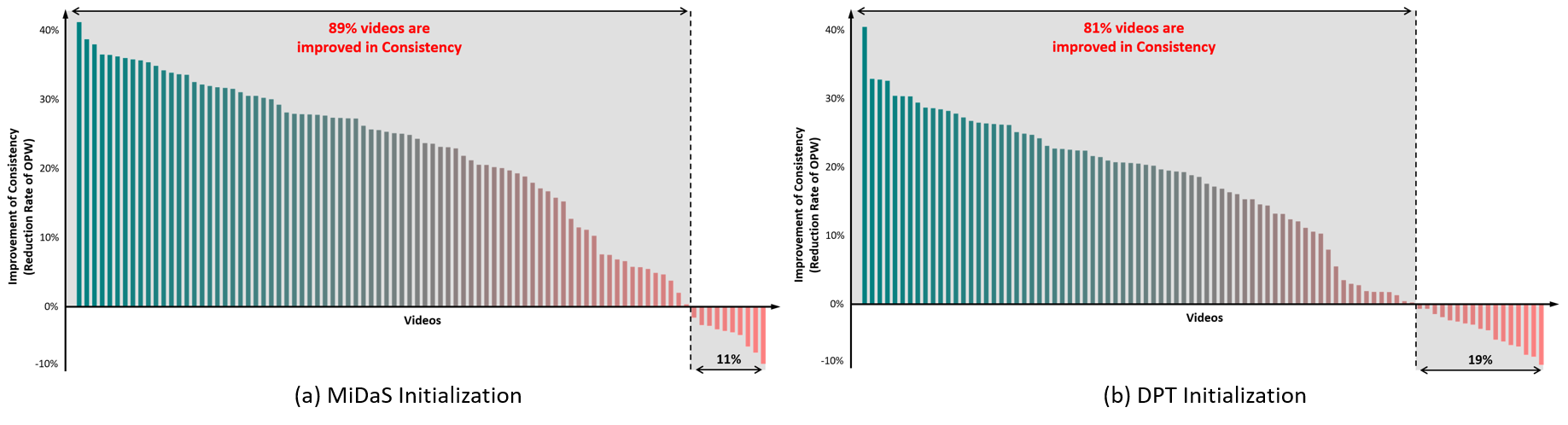}
\end{center}
\vspace{-12pt}
   \caption{
   \textbf{The effectiveness of \flow{}}. We showcase the detailed statistics of consistency improvements brought by flow-guided consistency fusion over the bidirectional inference with averaging. The reduction rates of $OPW$ are calculated for the 90 videos in the VDW test set, using MiDaS~\cite{MiDaS} or DPT~\cite{dpt} as varied depth predictors.}
\label{fig:rrrate}
\end{figure*}

\subsection{Robustness across Various Frame Rates}
\noindent \textbf{The Impacts of Frame Rates.} Similar to image resolution in the spatial dimension, we consider frame rates as the temporal resolution of videos. Videos with high frame rates represent small sampling intervals between consecutive frames. The inter-frame motions of moving objects and the camera could be smooth and coherent, providing sufficient temporal information. Thus, it could be easier for video depth models to predict consistent depth results. In contrast, lower frame rates represent larger sampling intervals and reduced inter-frame continuity. With lower resolution and less information in the temporal dimension, it becomes more challenging to stabilize the flickers in the predictions.

\noindent \textbf{Robustness on Different Frame Rates.} As illustrated in \refsec{}~\ref{sec:ddtj}, the proposed VDW dataset includes source videos with diverse frame rates. Therefore, our \sx{} can acquire strong robustness across various frame rates. As shown in \reffig{}~\ref{fig:fpsexp}, we conduct evaluations under varied frame rates from $15$ to $60$ fps for a certain video. The temporal window is fixed with three reference frames. Our model only exhibits minimal performance fluctuations. The results prove that our setting of the temporal window is appropriate and sufficient, showing robustness against the variations of fps.

We utilize a sequence of $60$ fps from the VDW dataset for the experiment in \reffig{}~\ref{fig:fpsexp}. Directly sampling the original video will result in varied frames for evaluation, making the metrics incomparable. Instead, we reduce the frame rates by increasing the inter-frame intervals of reference frames. For example, for the target frame $n$, using reference frames $i\in\{n\pm 3,n\pm 2,n\pm 1\}$ represents the original frame rate of 60 fps, while adopting $i\in\{n\pm 6,n\pm 4,n\pm 2\}$ represents 30 fps. In this way, we can still obtain the predictions for all original frames and compare the performance.

\noindent \textbf{The Ideal Setting of the Temporal Window.} No single setting can be optimal for all the videos. Different scenes, frame rates, objects, and motions in videos could all have an impact. For example, we cannot guarantee that three reference frames work best for all the videos, such as some special videos with extremely high frame rates, \textit{e.g.}, over $120$ fps. But this could be simply solved by adjusting the input inter-frame intervals without the need to retrain the model. 

Based on Table 11 (b) and Table 12 (b) in the main paper, we utilize three reference frames with the inter-frame interval $l = 1$ as the standard setting for all our experiments on different datasets~\cite{nyu,kitti,nvds,sintel,davis}. \reffig{}~\ref{fig:fpsexp} proves our strong robustness across various frame rates. Thus, users can simply follow the default setting for most videos. They can also adjust these settings, \textit{e.g.}, the inter-frame interval, according to their specific videos and applications.

\begin{table}[!t]
\caption{\textbf{Comparisons on the Sintel dataset.} We only report CVD~\cite{CVD} and Zhang \textit{et al.}~\cite{dycvd} on the $12$ videos with valid outputs, while other methods are on the $23$ videos.}
\label{tab:sintel23}
\begin{center}
\resizebox{\columnwidth}{!}{
\begin{tabular}{lcccccc}
\toprule
Method & $\delta_1\uparrow$ & $\delta_2\uparrow$ & $\delta_3\uparrow$ & $Rel\downarrow$ &$OPW\downarrow$ \\
\midrule
MiDaS-v2.1-Large~\cite{MiDaS} & $0.485$ & $0.693$ & $0.787$ & $0.410$ &$0.843$ \\
DPT-Large~\cite{dpt} & $\textbf{0.597}$ & $\underline{0.768}$ & $\underline{0.846}$ & $\underline{0.339}$ &$0.612$ \\
\midrule
ST-CLSTM~\cite{ST-CLSTM} & $0.351$ & $0.571$ & $0.706$ & $0.517$ &$0.585$ \\
FMNet~\cite{fmnet} & $0.357$ & $0.579$ & $0.712$ & $0.513$ &$0.521$ \\
DeepV2D~\cite{deepv2d} & $0.486$ & $0.674$ & $0.760$ & $0.526$ &$0.534$ \\
WSVD~\cite{wsvd} & $0.501$ & $0.709$ & $0.804$ & $0.439$ &$0.577$ \\
CVD~\cite{CVD} & $0.518$ & $0.741$ & $0.832$ & $0.406$ &$0.497$ \\
Robust-CVD~\cite{rcvd} & $0.521$ & $0.727$ & $0.833$ & $0.422$ &$0.475$ \\
Zhang \textit{et al.}~\cite{dycvd} & $0.522$ & $0.727$ & $0.831$ & $0.342$ &$0.481$ \\
\midrule

Ours-Large(MiDaS-v2.1-Large) & $0.532$ & $0.731$ & $0.833$ & $0.372$ &$\underline{0.447}$ \\
Ours-Large(DPT-Large) & $\underline{0.591}$ & $\textbf{0.770}$ & $\textbf{0.849}$ & $\textbf{0.335}$ & $\textbf{0.403}$ \\
\bottomrule
\end{tabular}
}
\end{center}
\vspace{-5pt}

\end{table}

\subsection{\FLOW{}}
We provide a toy experiment to illustrate the principle of \flow{}, showing that the strategy does not introduce systematic errors in the presence of motion, which is presented in \reffig{}~\ref{fig:toy}. From reference frame $i$ to target frame $n$, we assume that a deep blue rectangle (as the foreground object) is moving horizontally, while the gray areas represent the static background (e.g., a wall), as shown in the first and second columns of \reffig{}~\ref{fig:toy}. The bidirectional optical flow $FL_{n\Rightarrow i}$ and ${FL}_{i\Rightarrow n}$ can be calculated and visualized in the third and fourth columns. The white areas represent the static background where the optical flow is zero. The sky blue and the red areas showcase pixels with motion (\textit{i.e.}, large flow magnitude), representing the moving object in frame $n$ and frame $i$. For the relevance map $W_i$, we add the magnitude of $FL_{n\Rightarrow i}$ and ${FL}_{i\Rightarrow n}$ to perform the negative exponential transformation as Eq.~8 of the main paper. In this way, as shown in the last column of \reffig{}~\ref{fig:toy}, values of the relevance map $W_i$ are very low at the positions of the moving object in both frame $n$ and frame $i$. This prevents the fusion of the reference frame $i$ and preserves the depth $D_n^{bi}$ of the target frame $n$, as Eq.~9 of the main paper. The result is correct because, for the moving rectangle, foreground and background pixels are misaligned, and the reference frames should not be fused. 

Consequently, \flow{} does not introduce systematic errors in the presence of motion. For moving objects and regions, the depth of reference frames tends not to be fused due to misalignment. The original bidirectional depth $D_n^{bi}$ of the target frame will be preserved.

To further prove the effectiveness, we showcase the detailed statistics of consistency improvements brought by flow-guided consistency fusion over the bidirectional inference with averaging. As shown in \reffig{}~\ref{fig:rrrate}, the reduction rates of $OPW$ are calculated for the 90 videos in the VDW test set. With MiDaS~\cite{MiDaS} or DPT~\cite{dpt} as the depth predictor, $89\%$ and $81\%$ of all the videos achieve better consistency (above the X-axis) respectively. The depth accuracy is also maintained as proved by Table~5 of the main paper. Overall, bidirectional inference and \flow{} are simple but effective methods for improving consistency without introducing systematic errors, because of the adaptive fusion based on bidirectional optical flow, motion amplitude, and relevance maps. The experiments demonstrate that our approach only requires optical flow and works well for most testing videos.

On the other hand, the relations among camera motion, object motion, and depth variations could be complex in different scenarios. Our assumptions and methods could not fully cover all corner cases due to the diversity of real-world videos. We also try to explore some mechanisms that could be more comprehensive theoretically. However, these techniques introduce new problems in practice. For instance, camera motion compensation~\cite{camocomp} can be adopted to decouple the camera and object motion. But their reliance on camera parameters (\textit{e.g.}, the FOV) is impractical for in-the-wild videos, leading to failure cases and artifacts. Therefore, we use bidirectional optical flow to perform motion compensation in the flow-guided consistency fusion.

\subsection{Zero-shot Evaluations and Model Finetuning}
As presented in \reftab{}~\ref{tab:zerofine}, we report the results of \llarge{} with zero-shot evaluations (\textit{i.e.}, only trained on the VDW dataset)
and model finetuning on the NYUDV2~\cite{nyu} dataset. For zero-shot evaluations, our model improves both the temporal consistency and depth accuracy over the depth predictors~\cite{MiDaS,dpt}, showing the generalization ability of our method. Besides, the finetuning can further improve the depth accuracy for closed-domain applications, \textit{e.g.}, the static indoor scenes of the NYUDV2~\cite{nyu} dataset.

\subsection{Runtime Analysis}
For the \ssmall{} model, we report the runtime of each component in milliseconds ($ms$), including the depth predictors~\cite{MiDaS,MiDaSV31}, the feature encoder, the cross-attention module, and the decoder. The stabilization network achieves faster inference speed than lightweight depth predictors~\cite{MiDaS,MiDaSV31}. Combining all components, \ssmall{} can still achieve real-time processing of over $30$ fps.

\subsection{More Quantitative Comparisons}
In the main paper, only $\delta_1$, $Rel$, and $OPW$ are reported. The additional results on the VDW and the Sintel~\cite{sintel} dataset are shown in \reftab{}~\ref{tab:vdw} and \reftab{}~\ref{tab:sintel23}. Besides, as CVD~\cite{CVD} and Zhang~\textit{et al.}~\cite{dycvd} cannot produce results on $11$ of $23$ videos in Sintel~\cite{sintel} dataset, we additionally report the results on the other $12$ videos in \reftab{}~\ref{tab:sintel12}. 


\begin{table}
    \centering

    \caption{\textbf{Zero-shot evaluations and model finetuning.} DPT-Large~\cite{dpt} and MiDaS-v2.1-Large~\cite{MiDaS} are adopted as different depth predictors. We report the results of \llarge{} with zero-shot evaluations (\textit{i.e.}, only trained on the VDW dataset) and model finetuning on the NYUDV2~\cite{nyu} dataset.} 
    \label{tab:zerofine}
    
    \hspace{-15pt}\resizebox{0.45\columnwidth}{!}{
    \begin{subtable}[t]{0.45\linewidth}
    \addtolength{\tabcolsep}{-4pt}
        \begin{tabular}{lcc}
            \toprule
            Method & $\delta_1\uparrow$ & $OPW\downarrow$ \\
            \midrule
            DPT~\cite{dpt} & $0.928$ &$0.811$ \\
            Zero-Shot(DPT) & $\underline{0.930}$ &$\underline{0.351}$ \\
            Finetune(DPT) & $\textbf{0.950}$ &$\textbf{0.339}$ \\
            \bottomrule
        \end{tabular}
        \vspace{+1pt}\caption{DPT Initialization}
    \end{subtable}
    }
    \hspace{+5pt}\resizebox{0.45\columnwidth}{!}{
    \begin{subtable}[t]{0.45\linewidth}
    \addtolength{\tabcolsep}{-4pt}
        \begin{tabular}{lcc}
            \toprule
            Method & $\delta_1\uparrow$ & $OPW\downarrow$ \\
            \midrule
            MiDaS~\cite{MiDaS} & $0.910$ &$0.862$ \\
            Zero-Shot(MiDaS) & $\underline{0.919}$ &$\textbf{0.332}$ \\
            Finetune(MiDaS) & $\textbf{0.941}$ &$\underline{0.347}$ \\
            \bottomrule
        \end{tabular}
        \vspace{+1pt}\caption{MiDaS Initialization}
    \end{subtable}
    }
    
\end{table}

\begin{table}[!t]
    \centering
    \caption{\textbf{Runtime of the lightweight \ssmall{} model.} We report the runtime of each component to predict one $896\times 384$ frame on an NVIDIA RTX A6000 GPU. The \ssmall{} model shows high efficiency for real-time applications.}
    \label{tab:mstime}
    
    \addtolength{\tabcolsep}{-3pt}
    
    \resizebox{\columnwidth}{!}{
    \begin{tabular}{lccc}
        \toprule
        Module & Component & Runtime ($ms$) & Overall ($ms$) \\
        \midrule
        \multirow{2}{*}{Depth Predictor} & DPT-Swin2-Tiny~\cite{MiDaSV31} & $24.18$ & $24.18$ \\
        & MiDaS-v2.1-Small~\cite{MiDaS} & $22.35$ & $22.35$\\
        \midrule
        \multirow{2}{*}{Stabilization}& Feature Encoder& $1.34$ & \multirow{3}{*}{$6.29$}\\
        \multirow{2}{*}{Network} & Cross-attention & $0.89$ & \\
         & Decoder & $4.06$ & \\
        \bottomrule
    \end{tabular}
    }
\end{table}

\begin{table}
\caption{\textbf{Comparisons on the $12$ videos of Sintel~\cite{sintel} dataset.} We test the $12$ videos that CVD~\cite{CVD} and Zhang \textit{et al.}~\cite{dycvd} can produce results for comparisons.}
\vspace{-5pt}
\label{tab:sintel12}
\begin{center}
\resizebox{\columnwidth}{!}{
\begin{tabular}{lcccccc}
\toprule
Method & $\delta_1\uparrow$ & $\delta_2\uparrow$ & $\delta_3\uparrow$ & $Rel\downarrow$ &$OPW\downarrow$ \\
\midrule
MiDaS-v2.1-Large~\cite{MiDaS} & $0.670$ & $0.853$ & $0.902$ & $0.246$ &$0.712$ \\
DPT-Large~\cite{dpt} & $\textbf{0.747}$ & $\underline{0.874}$ & $0.917$ & $\textbf{0.196}$ &$0.671$ \\
\midrule
ST-CLSTM~\cite{ST-CLSTM} & $0.477$ & $0.711$ & $0.827$ & $0.366$ &$0.547$ \\
FMNet~\cite{fmnet} & $0.492$ & $0.728$ & $0.825$ & $0.363$ &$0.516$ \\
DeepV2D~\cite{deepv2d} & $0.509$ & $0.735$ & $0.827$ & $0.384$ &$0.575$ \\
CVD~\cite{CVD} & $0.518$ & $0.741$ & $0.832$ & $0.406$ &$0.497$ \\
Zhang \textit{et al.}~\cite{dycvd} & $0.522$ & $0.727$ & $0.831$ & $0.342$ &$0.481$ \\
WSVD~\cite{wsvd} & $0.621$ & $0.822$ & $0.891$ & $0.305$ &$0.581$ \\
Robust-CVD~\cite{rcvd} & $0.673$ & $0.848$ & $0.888$ & $0.284$ &$0.447$ \\

\midrule

Ours-Large(MiDaS-v2.1-Large) & $0.701$ & $0.867$ & $\underline{0.918}$ & $0.215$ &$\underline{0.403}$ \\
Ours-Large(DPT-Large) & $\underline{0.741}$ & $\textbf{0.878}$ & $\textbf{0.925}$ & $\underline{0.201}$ & $\textbf{0.392}$ \\
\bottomrule
\end{tabular}
}
\end{center}
\vspace{-5pt}

\end{table}


\subsection{More Qualitative Results.}
We show more visual comparisons in \reffig{}~\ref{fig:ksh1} and \ref{fig:ksh-sin}. We draw the scanline slice over time. Fewer zigzagging pattern means better consistency. Please refer to our demo video and project page for more video results and comparisons.

\begin{figure*}
\begin{center}
   \includegraphics[width=0.965\textwidth,trim=0 0 0 0,clip]{suppfig/qiepian-supp1.pdf}
\end{center}
   \caption{
   \textbf{More qualitative results on natural scenes.} The first image in each pair is the RGB frame, while the second is the scanline slice over time. Fewer zigzagging pattern means better consistency.}
\label{fig:ksh1}
\end{figure*}

\begin{figure*}
\begin{center}
   \includegraphics[width=0.97\textwidth,trim=0 0 0 0,clip]{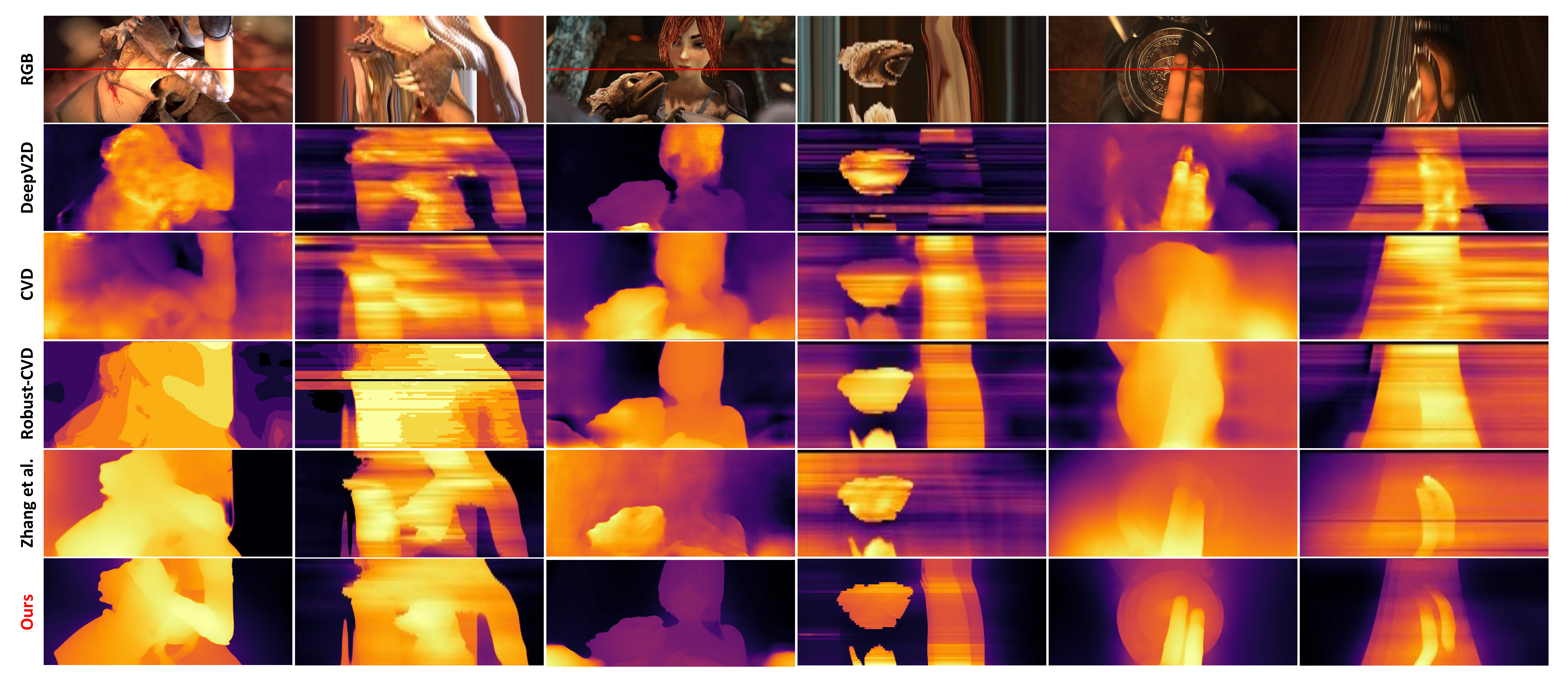}
\end{center}
\vspace{-12pt}
   \caption{
   \textbf{Qualitative results on Sintel~\cite{sintel} dataset.} We compare the results of DeepV2D~\cite{deepv2d}, CVD~\cite{CVD}, Robust-CVD~\cite{rcvd}, and Zhang \textit{et al.}~\cite{dycvd}. Without relying on test-time training~\cite{CVD,rcvd,dycvd}, we conduct zero-shot evaluations on Sintel~\cite{sintel} and achieve significantly better performance than those test-time-training-based methods~\cite{CVD,rcvd,dycvd}.}
\label{fig:ksh-sin}
\vspace{-5pt}
\end{figure*}

\section{Image Attribution}
We properly attribute the sources of all images and figures throughout our main manuscript and supplementary document, as presented in \reftab{}~\ref{tab:attribute}. We also specify the images from movies, documentaries, and animations with their IMDB movie numbers. Web videos and public datasets do not have IMDB numbers, with $-$ as representation.

\begin{table}[!h]
    \centering
    \caption{\textbf{Attribution of the images in our main manuscript and supplement.} We report the image attribution of all figures throughout our paper.}
    \label{tab:attribute}
    
    \addtolength{\tabcolsep}{-6pt}
    
    \resizebox{0.922\columnwidth}{!}{
    \begin{tabular}{lccc}
        \toprule
        Figures & Types & Attribution & IMDB Numbers\\
        \midrule
        \multicolumn{4}{c}{Main Manuscript} \\
        \midrule
        \reffig{}~1 & Movie & Everest & 2719848 \\
        \cmidrule{2-4}
        \reffig{}~2 & Movie & Everest & 2719848 \\
        \cmidrule{2-4}
        \multirow{2}{*}{\reffig{}~4} &  Animation & Frozen 2 & 4520988 \\
        & Movie & Eternals & 9032400 \\
        \cmidrule{2-4}
        \multirow{6}{*}{\reffig{}~5} &  Web Video & YouTube & $-$ \\
        & Documentary & Jerusalem & 2385006 \\
        & Documentary & Kingdom of Plants & 2117380 \\
        &  Animation & Kung Fu Panda 3 & 2267968\\
        & Movie & The Hobbit 2 & 1170358\\
        & Movie & The Great Gatsby & 1343092\\
        \cmidrule{2-4}
        \multirow{2}{*}{\reffig{}~7} & Movie & Eternals & 9032400 \\
        &  Animation & Frozen 2 & 4520988 \\
        \cmidrule{2-4}
        \reffig{}~8 &  Public Dataset & DAVIS~\cite{davis} & $-$ \\
        \cmidrule{2-4}
        \reffig{}~9 & Movie & Eternals & 9032400 \\
        \cmidrule{2-4}
        \reffig{}~10 &  Public Dataset & CityScapes~\cite{cityscapes} & $-$ \\
        \cmidrule{2-4}
        \reffig{}~11 &  Public Dataset & NYUDV2~\cite{nyu} & $-$ \\
        \cmidrule{2-4}
        \multirow{3}{*}{\reffig{}~12} & Movie & Eternals & 9032400 \\
        & Movie & Everest & 2719848 \\
        & Web Video & NSFF~\cite{nsff} Demo & $-$ \\
        \cmidrule{2-4}
        \multirow{2}{*}{\reffig{}~13} & Movie & Eternals & 9032400 \\
        & Movie & Fantastic Beasts and Where to Find Them & 3183660 \\
        \cmidrule{2-4}
        \reffig{}~14 &  Public Dataset & Sintel~\cite{sintel} & $-$ \\

        \midrule
        \multicolumn{4}{c}{Supplementary Document} \\
        \midrule
        \multirow{14}{*}{\reffig{}~17} & Web Video & YouTube & $-$ \\
        & Web Video & bilibili & $-$ \\
        & Documentary & Jerusalem & 2385006 \\
        & Documentary & Kingdom of Plants & 2117380 \\
        & Documentary & Little Monsters & 11019830 \\
        & Documentary & Deepsea Challenge & 2332883 \\
        &  Animation & Kung Fu Panda 3 & 2267968\\
        &  Animation & Coco & 2380307\\
        & Movie & The Great Gatsby & 1343092\\
        & Movie & Mission: Impossible-Fallout & 4912910 \\
        & Movie & Doctor Strange & 1211837 \\
        & Movie & Transformers: Age of Extinction  & 2109248 \\
        & Movie & The Legend of Tarzan & 0918940 \\
        & Movie & Exodus: Gods and Kings & 1528100 \\
        \cmidrule{2-4}
        \multirow{4}{*}{\reffig{}~22} & Web Video & YouTube & $-$ \\
        & Web Video & bilibili & $-$ \\
        & Movie & Eternals & 9032400 \\
        & Movie & Fantastic Beasts and Where to Find Them & 3183660 \\
        \cmidrule{2-4}
        \reffig{}~23 &  Public Dataset & Sintel~\cite{sintel} & $-$ \\

        \bottomrule
    \end{tabular}
    }
\end{table}

\ifCLASSOPTIONcaptionsoff
  \newpage
\fi

\end{document}